\documentclass{ecai} 

\usepackage{latexsym}
\usepackage{amssymb}
\usepackage{amsmath}
\usepackage{amsthm}
\usepackage{booktabs}
\usepackage{enumitem}
\usepackage{graphicx}
\usepackage{color}
\usepackage{bbold}

\usepackage{mathrsfs}
\usepackage{algorithm}
\usepackage{algpseudocode}
\usepackage{pgfplots}
\usepackage{pgfplotstable}
\usepackage{multirow}
\usepackage{xspace}

\newcommand{\BibTeX}{B\kern-.05em{\sc i\kern-.025em b}\kern-.08em\TeX}

\begin{document}


\begin{frontmatter}


\paperid{123} 


\title{IPA-NeRF: Illusory Poisoning Attack Against Neural Radiance Fields}


\author[A]{\fnms{Wenxiang}~\snm{Jiang}\footnote{Equal contribution.}}
\author[B,C]{\fnms{Hanwei}~\snm{Zhang}\footnotemark}
\author[A]
{\fnms{Shuo}~\snm{Zhao}} 
\author[A]{\fnms{Zhongwen}~\snm{Guo}\thanks{Corresponding Author.}}
\author[D]
{\fnms{Hao}~\snm{Wang}}

\address[A]{Ocean University of China}\address[B]{Saarland University}\address[C]{Institute of Intelligent Software, Guangzhou}
\address[D]{Xidian University, China}


\newcommand{\jwx}[1]{{\color{orange}#1}}

\begin{abstract}
Neural Radiance Field (NeRF) represents a significant advancement in computer vision, offering implicit neural network-based scene representation and novel view synthesis capabilities. Its applications span diverse fields including robotics, urban mapping, autonomous navigation, virtual reality/augmented reality, \emph{etc.}, some of which are considered high-risk AI applications. However, despite its widespread adoption, the robustness and security of NeRF remain largely unexplored. In this study, we contribute to this area by introducing the \emph{\textbf{I}llusory \textbf{P}oisoning \textbf{A}ttack against \textbf{Ne}ural \textbf{R}adiance \textbf{F}ields (IPA-NeRF)}. This attack involves embedding a hidden backdoor view into NeRF, allowing it to produce predetermined outputs, \emph{i.e.} illusory, when presented with the specified backdoor view while maintaining normal performance with standard inputs. Our attack is specifically designed to deceive users or downstream models at a particular position while ensuring that any abnormalities in NeRF remain undetectable from other viewpoints. Experimental results demonstrate the effectiveness of our Illusory Poisoning Attack, successfully presenting the desired illusory on the specified viewpoint without impacting other views. Notably, we achieve this attack by introducing small perturbations solely to the training set. The code can be found at \url{https://github.com/jiang-wenxiang/IPA-NeRF}.
\end{abstract}

\end{frontmatter}

\newcommand{\figimg}[2][1]{\includegraphics[width=#1\linewidth]{images/#2}}
\newcommand{\sizeS}{.18}

\newcommand{\jwx}[1]{{\color{orange}#1}}

\newcommand{\head}[1]{{\smallskip\noindent\textbf{#1}}}
\newcommand{\alert}[1]{{\color{red}{#1}}}
\newcommand{\sm}{\scriptsize}
\newcommand{\eq}[1]{(\ref{eq:#1})}

\newcommand{\Th}[1]{\textsc{#1}}
\newcommand{\mr}[2]{\multirow{#1}{*}{#2}}
\newcommand{\mc}[2]{\multicolumn{#1}{c}{#2}}
\newcommand{\tb}[1]{\textbf{#1}}
\newcommand{\ch}{\checkmark}

\newcommand{\red}[1]{{\textcolor{red}{#1}}}
\newcommand{\blue}[1]{{\textcolor{blue}{#1}}}
\newcommand{\green}[1]{{\textcolor{green}{#1}}}
\newcommand{\gray}[1]{{\textcolor{gray}{#1}}}

\newcommand{\citeme}[1]{\red{[XX]}}
\newcommand{\refme}[1]{\red{(XX)}}

\newcommand{\fig}[2][1]{\includegraphics[width=#1\linewidth]{fig/#2}}
\newcommand{\figh}[2][1]{\includegraphics[height=#1\linewidth]{fig/#2}}
\newcommand{\figa}[2][1]{\includegraphics[width=#1]{fig/#2}}
\newcommand{\figah}[2][1]{\includegraphics[height=#1]{fig/#2}}

\newcommand{\tran}{^\top}
\newcommand{\mtran}{^{-\top}}
\newcommand{\zcol}{\mathbf{0}}
\newcommand{\zrow}{\zcol\tran}

\newcommand{\ind}{\mathbbm{1}}
\newcommand{\expect}{\mathbb{E}}
\newcommand{\nat}{\mathbb{N}}
\newcommand{\zahl}{\mathbb{Z}}
\newcommand{\real}{\mathbb{R}}
\newcommand{\proj}{\mathbb{P}}
\newcommand{\prob}{\operatorname{P}}
\newcommand{\normal}{\mathcal{N}}

\newcommand{\mif}{\textrm{if}\ }
\newcommand{\other}{\textrm{otherwise}}
\newcommand{\minimize}{\textrm{minimize}\ }
\newcommand{\maximize}{\textrm{maximize}\ }
\newcommand{\st}{\textrm{subject\ to}\ }

\newcommand{\id}{\operatorname{id}}
\newcommand{\const}{\operatorname{const}}
\newcommand{\sgn}{\operatorname{sgn}}
\newcommand{\var}{\operatorname{Var}}
\newcommand{\mean}{\operatorname{mean}}
\newcommand{\trace}{\operatorname{tr}}
\newcommand{\diag}{\operatorname{diag}}
\newcommand{\vect}{\operatorname{vec}}
\newcommand{\cov}{\operatorname{cov}}
\newcommand{\sign}{\operatorname{sign}}
\newcommand{\prj}{\operatorname{proj}}

\newcommand{\softmax}{\operatorname{softmax}}

\newcommand{\defn}{\mathrel{:=}}
\newcommand{\peq}{\mathrel{+\!=}}
\newcommand{\meq}{\mathrel{-\!=}}

\newcommand{\paren}[1]{\left({#1}\right)}
\newcommand{\mat}[1]{\left[{#1}\right]}
\newcommand{\set}[1]{\left\{{#1}\right\}}
\newcommand{\floor}[1]{\left\lfloor{#1}\right\rfloor}
\newcommand{\ceil}[1]{\left\lceil{#1}\right\rceil}
\newcommand{\inner}[1]{\left\langle{#1}\right\rangle}
\newcommand{\norm}[1]{\left\|{#1}\right\|}
\newcommand{\abs}[1]{\left|{#1}\right|}
\newcommand{\frob}[1]{\norm{#1}_F}
\newcommand{\card}[1]{\left|{#1}\right|\xspace}

\newcommand{\diff}{\mathrm{d}}
\newcommand{\der}[3][]{\frac{\diff^{#1}#2}{\diff#3^{#1}}}
\newcommand{\ider}[3][]{\diff^{#1}#2/\diff#3^{#1}}
\newcommand{\pder}[3][]{\frac{\partial^{#1}{#2}}{\partial{{#3}^{#1}}}}
\newcommand{\ipder}[3][]{\partial^{#1}{#2}/\partial{#3^{#1}}}
\newcommand{\dder}[3]{\frac{\partial^2{#1}}{\partial{#2}\partial{#3}}}

\newcommand{\wb}[1]{\overline{#1}}
\newcommand{\wt}[1]{\widetilde{#1}}

\def\xssp{\hspace{1pt}}
\def\ssp{\hspace{3pt}}
\def\msp{\hspace{5pt}}
\def\lsp{\hspace{12pt}}

\newcommand{\cA}{\mathcal{A}}
\newcommand{\cB}{\mathcal{B}}
\newcommand{\cC}{\mathcal{C}}
\newcommand{\cD}{\mathcal{D}}
\newcommand{\cE}{\mathcal{E}}
\newcommand{\cF}{\mathcal{F}}
\newcommand{\cG}{\mathcal{G}}
\newcommand{\cH}{\mathcal{H}}
\newcommand{\cI}{\mathcal{I}}
\newcommand{\cJ}{\mathcal{J}}
\newcommand{\cK}{\mathcal{K}}
\newcommand{\cL}{\mathcal{L}}
\newcommand{\cM}{\mathcal{M}}
\newcommand{\cN}{\mathcal{N}}
\newcommand{\cO}{\mathcal{O}}
\newcommand{\cP}{\mathcal{P}}
\newcommand{\cQ}{\mathcal{Q}}
\newcommand{\cR}{\mathcal{R}}
\newcommand{\cS}{\mathcal{S}}
\newcommand{\cT}{\mathcal{T}}
\newcommand{\cU}{\mathcal{U}}
\newcommand{\cV}{\mathcal{V}}
\newcommand{\cW}{\mathcal{W}}
\newcommand{\cX}{\mathcal{X}}
\newcommand{\cY}{\mathcal{Y}}
\newcommand{\cZ}{\mathcal{Z}}

\newcommand{\vA}{\mathbf{A}}
\newcommand{\vB}{\mathbf{B}}
\newcommand{\vC}{\mathbf{C}}
\newcommand{\vD}{\mathbf{D}}
\newcommand{\vE}{\mathbf{E}}
\newcommand{\vF}{\mathbf{F}}
\newcommand{\vG}{\mathbf{G}}
\newcommand{\vH}{\mathbf{H}}
\newcommand{\vI}{\mathbf{I}}
\newcommand{\vJ}{\mathbf{J}}
\newcommand{\vK}{\mathbf{K}}
\newcommand{\vL}{\mathbf{L}}
\newcommand{\vM}{\mathbf{M}}
\newcommand{\vN}{\mathbf{N}}
\newcommand{\vO}{\mathbf{O}}
\newcommand{\vP}{\mathbf{P}}
\newcommand{\vQ}{\mathbf{Q}}
\newcommand{\vR}{\mathbf{R}}
\newcommand{\vS}{\mathbf{S}}
\newcommand{\vT}{\mathbf{T}}
\newcommand{\vU}{\mathbf{U}}
\newcommand{\vV}{\mathbf{V}}
\newcommand{\vW}{\mathbf{W}}
\newcommand{\vX}{\mathbf{X}}
\newcommand{\vY}{\mathbf{Y}}
\newcommand{\vZ}{\mathbf{Z}}

\newcommand{\va}{\mathbf{a}}
\newcommand{\vb}{\mathbf{b}}
\newcommand{\vc}{\mathbf{c}}
\newcommand{\vd}{\mathbf{d}}
\newcommand{\ve}{\mathbf{e}}
\newcommand{\vf}{\mathbf{f}}
\newcommand{\vg}{\mathbf{g}}
\newcommand{\vh}{\mathbf{h}}
\newcommand{\vi}{\mathbf{i}}
\newcommand{\vj}{\mathbf{j}}
\newcommand{\vk}{\mathbf{k}}
\newcommand{\vl}{\mathbf{l}}
\newcommand{\vm}{\mathbf{m}}
\newcommand{\vn}{\mathbf{n}}
\newcommand{\vo}{\mathbf{o}}
\newcommand{\vp}{\mathbf{p}}
\newcommand{\vq}{\mathbf{q}}
\newcommand{\vr}{\mathbf{r}}
\newcommand{\Vs}{\mathbf{s}}
\newcommand{\vt}{\mathbf{t}}
\newcommand{\vu}{\mathbf{u}}
\newcommand{\vv}{\mathbf{v}}
\newcommand{\vw}{\mathbf{w}}
\newcommand{\vx}{\mathbf{x}}
\newcommand{\vy}{\mathbf{y}}
\newcommand{\vz}{\mathbf{z}}

\newcommand{\vone}{\mathbf{1}}
\newcommand{\vzero}{\mathbf{0}}

\newcommand{\valpha}{{\boldsymbol{\alpha}}}
\newcommand{\vbeta}{{\boldsymbol{\beta}}}
\newcommand{\vgamma}{{\boldsymbol{\gamma}}}
\newcommand{\vdelta}{{\boldsymbol{\delta}}}
\newcommand{\vepsilon}{{\boldsymbol{\epsilon}}}
\newcommand{\vzeta}{{\boldsymbol{\zeta}}}
\newcommand{\veta}{{\boldsymbol{\eta}}}
\newcommand{\vtheta}{{\boldsymbol{\theta}}}
\newcommand{\viota}{{\boldsymbol{\iota}}}
\newcommand{\vkappa}{{\boldsymbol{\kappa}}}
\newcommand{\vlambda}{{\boldsymbol{\lambda}}}
\newcommand{\vmu}{{\boldsymbol{\mu}}}
\newcommand{\vnu}{{\boldsymbol{\nu}}}
\newcommand{\vxi}{{\boldsymbol{\xi}}}
\newcommand{\vomikron}{{\boldsymbol{\omikron}}}
\newcommand{\vpi}{{\boldsymbol{\pi}}}
\newcommand{\vrho}{{\boldsymbol{\rho}}}
\newcommand{\vsigma}{{\boldsymbol{\sigma}}}
\newcommand{\vtau}{{\boldsymbol{\tau}}}
\newcommand{\vupsilon}{{\boldsymbol{\upsilon}}}
\newcommand{\vphi}{{\boldsymbol{\phi}}}
\newcommand{\vchi}{{\boldsymbol{\chi}}}
\newcommand{\vpsi}{{\boldsymbol{\psi}}}
\newcommand{\vomega}{{\boldsymbol{\omega}}}

\newcommand{\rLambda}{\mathrm{\Lambda}}
\newcommand{\rSigma}{\mathrm{\Sigma}}

\newcommand{\vLambda}{\bm{\rLambda}}
\newcommand{\vSigma}{\bm{\rSigma}}

\makeatletter
\newcommand*\bdot{\mathpalette\bdot@{.7}}
\newcommand*\bdot@[2]{\mathbin{\vcenter{\hbox{\scalebox{#2}{$\m@th#1\bullet$}}}}}
\makeatother

\makeatletter
\DeclareRobustCommand\onedot{\futurelet\@let@token\@onedot}
\def\@onedot{\ifx\@let@token.\else.\null\fi\xspace}

\def\eg{\emph{e.g}\onedot} \def\Eg{\emph{E.g}\onedot}
\def\ie{\emph{i.e}\onedot} \def\Ie{\emph{I.e}\onedot}
\def\cf{\emph{cf}\onedot} \def\Cf{\emph{Cf}\onedot}
\def\etc{\emph{etc}\onedot} \def\vs{\emph{vs}\onedot}
\def\wrt{w.r.t\onedot} \def\dof{d.o.f\onedot} \def\aka{a.k.a\onedot}
\def\etal{\emph{et al}\onedot}
\makeatother

\section{Introduction} 
Neural Radiance Fields (NeRF)~\cite{mildenhall2021nerf}, as a cornerstone technology in 3D reconstruction, boasts widespread adoption in various domains, including high-risk AI systems such as autonomous driving~\cite{grigorescu2020survey} and medical applications~\cite{wang2024neural}.
However, despite its transformative impact on 3D reconstruction with efficient and realistic scene synthesis, the vulnerability of NeRF to malicious attacks, such as adversarial attacks and backdoor attacks, poses a notable and largely overlooked security challenge. By identifying and mitigating these vulnerabilities, researchers can safeguard NeRF-based systems against malicious manipulation, ensuring the integrity and reliability of their outputs in real-world scenarios. Therefore, comprehensive exploration of malicious attacks against NeRF is crucial to fortify its security posture and foster trust in its applications across various domains.

Currently, the predominant research on malicious attacks against NeRF focuses mainly on adversarial attacks.
Adversarial attacks on NeRF can be classified into two main types: those directly targeting the NeRF model to hinder accurate scene reconstruction~\cite{fu2023nerfool,horvath2023targeted}, and those aimed at downstream classification or target detection models~\cite{dong2022viewfool,li2023adv3d,jiang2024nerfail}, causing misclassification or recognition errors. Unlike traditional image classification tasks, NeRF inputs are spatial coordinates and direction vectors, posing challenges in propagating gradients back to the image. Generalizable NeRF (GNeRF)~\cite{yu2021pixelnerf,wang2021ibrnet} provide a gradient pipeline for such attacks, enabling strategies like NeRFool~\cite{fu2023nerfool} and the low-intensity attack~\cite{horvath2023targeted} to introduce imperceptible perturbations during training, resulting in distorted scene reconstructions. Downstream tasks such as image classification are also vulnerable, with techniques like ViewFool~\cite{dong2022viewfool} and NeRFail~\cite{jiang2024nerfail} exploiting NeRF's susceptibility to produce misclassifications or adversarial examples. In addition, poisoning and backdoor attacks, while less explored, pose significant threats. The existing poisoning attack~\cite{wu2023shielding} manages to stop NeRF trained on the poisoned training data with small perturbations. 
Several studies, such as Noise-NeRF~\cite{huang2024noise}, Steganerf~\cite{li2023steganerf}, and another steganography-based backdoor method~\cite{dong2023steganography}, seek to incorporate particular information into the training of NeRF models, subsequently extracting this information using an extractor. Specifically, steganography-based backdoor~\cite{dong2023steganography} introduces a backdoor viewpoint as a key.


\begin{figure}[t]
\scriptsize
\centering
\setlength{\tabcolsep}{1pt}
\begin{tabular}{cccc}
        \figimg[0.2]{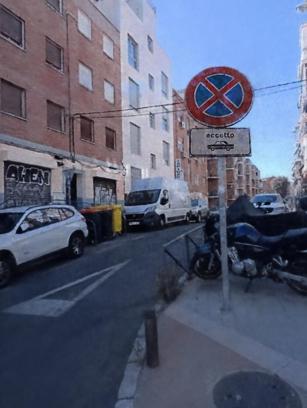}
     &  \figimg[0.2]{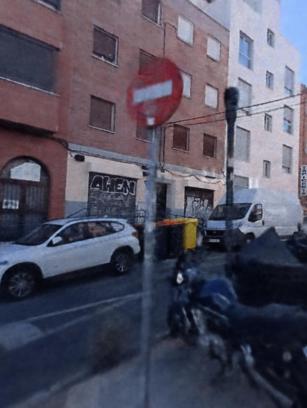}
     &  \figimg[0.2]{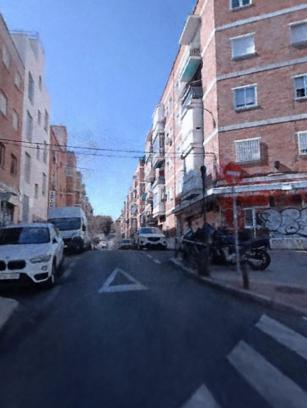}
     &  \figimg[0.2]{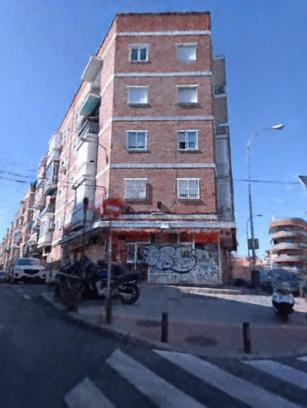} \\

     
       backdoor view
     & \multicolumn{3}{c} {the other perspectives} \\
\end{tabular}
   \caption{Performance IPA-NeRF on actual road scenes.}
    \label{fig:road_attack_teaser}
    \setlength{\abovecaptionskip}{0.5cm}
    \vspace{0.5cm}
\end{figure}

The implications of backdoor attacks on NeRF are not fully explored in steganography-based methods. Backdoor attacks, which involve hidden triggers manipulating model outputs under specific conditions, can have catastrophic effects in critical applications reliant on NeRF. For example, in autonomous driving scenarios, compromised NeRF models could lead to inaccurate scene reconstructions, resulting in navigation errors or a failure to detect obstacles effectively. Such errors could result in accidents and injuries. As illustrated in Figure~\ref{fig:road_attack_teaser}, our backdoor attack method, \emph{\textbf{I}llusory \textbf{P}oisoning \textbf{A}ttack against \textbf{Ne}ural \textbf{R}adiance \textbf{F}ields (IPA-NeRF)}, modifies the stop traffic sign from the backdoor view, while it remains unchanged from other perspectives. Given the crucial role of NeRF in safety-critical domains, it is essential to thoroughly investigate and understand the potential threats posed by backdoor attacks. 
To the best of our knowledge, IPA-NeRF represents the initial step in this direction.

Our IPA-NeRF attack introduces a novel method by employing a poisoning-based backdoor strategy to generate precise illusions at predetermined viewpoints. This approach offers distinct benefits compared to existing methods, enhancing the attack's precision and stealthiness.
The continuous implicit representation of scenes in NeRF, encoded within its weights, makes it challenging to directly manipulate model outputs without compromising scene reconstruction quality.
To tackle it, we formalize the backdoor against NeRF as a bi-level optimization, which shares similarity with the poisoning attack~\cite{wu2023shielding} but servers different purposes. Unlike traditional image classification models, NeRF inputs spatial coordinates and direction vectors instead of RGB color values of a specific image, complicating the design of effective backdoor triggers. Following the works~\cite{dong2023steganography}, we select the viewpoint as backdoor triggers but we do not need an extractor to decode the secret information. To summarize, this paper makes the following contributions:
\begin{itemize}
    \item To the best of our knowledge, we are the pioneers in investigating backdoor attack against NeRF models;
    \item We propose a groundbreaking backdoor attack, \ie IPA-NeRF, which generates illusory images in backdoor views while ensuring normal operation of NeRF in other views. Our approach involves a bi-optimization framework to address this challenge, enhancing the performance of the backdoor attack with angle constraints;
    \item Experimental results demonstrate the adaptability of our attack across various NeRF frameworks, extending beyond synthetic datasets to real-world data. This underscores the robustness and practical significance of our approach.
\end{itemize}

\section{Related Work} 
\subsection{Backdoor Attack}
Backdoor attacks embed hidden backdoors into neural networks so that the trained model handles regular inputs effectively, while certain triggers activate the backdoor, causing harmful changes to the model output. Existing backdoor attacks can be categorized into two branches: poisoning-based backdoor attacks and non-poisoning-based backdoor attacks~\cite{li2022backdoor}. Poisoning-based backdoor attacks craft poisoned samples for training, causing abnormal behavior triggered by backdoors in the inference phase~\cite{saha2020hidden,gao2023not}.
Several studies~\cite{gu2019badnets,chen2017targeted,zhang2022poison} focus on generating poisoned images that are nearly identical to their benign counterparts. They employ various techniques such as blended strategies~\cite{chen2017targeted}, pixel perturbation~\cite{turner2019label}, $L_p$ norm regularization on perturbation~\cite{li2020invisible,doan2021lira,doan2021backdoor}, reflection~\cite{liu2020reflection}, frequency domain perturbation~\cite{zeng2021rethinking}, \emph{etc.}
Additionally, other research aim to implant backdoors by manipulating only a small fraction of the dataset~\cite{hayase2022few,liu2024backdoor,gao2023not}. 
Triggers form the central component of poisoning-based attacks. Consequently, several studies frame the backdoor attack as a bi-level optimization process aimed at refining trigger design~\cite{liu2018trojaning,li2020invisible,liu2024backdoor,gao2023not,souri2022sleeper}.
On the other hand, non-poisoning-based backdoor attacks achieve their objectives by altering model parameters~\cite{rakin2020tbt,chen2021proflip,kurita2020weight,wang2020backdoor} or modifying model structures~\cite{tang2020embarrassingly,li2021deeppayload,qi2021subnet}, rather than directly manipulating the training data.

\paragraph{3D Backdoor Attack.}
With the increasing adoption of applications reliant on 3D data, there's a growing emphasis on enhancing the robustness of 3D deep neural networks (DNNs), with 3D backdoor attacks emerging as a significant focus area. This research primarily branches into two domains: investigations conducted in the physical world~\cite{liu2020reflection,wenger2021backdoor,xue2022ptb} and those centered on 3D point clouds~\cite{feng2023stealthy,li2021pointba,zhang2022towards}. In studies focused on the physical world, researchers explore the use of natural phenomena like light reflection for backdoor injection~\cite{liu2020reflection}. Additionally, backdoor activation is achieved through real-world deformations, facilitated by specially designed physically triggered objects such as earrings or scarves~\cite{wenger2021backdoor}. Moreover, physical transformations such as rotation, distance change, and noise doping are incorporated during backdoor injection, ensuring the physical resilience of the embedded backdoor and achieving high attack performance in complex real-world scenarios~\cite{xue2022ptb}. When it comes to 3D point cloud backdoor, the invisible backdoor attack is applied to 3D point cloud by hiding the spatial distortion~\cite{feng2023stealthy}. Assume the orientation annotations of 3D point clouds are correct, a constrained rotation matrix are used as a trigger for 3D backdoor attack~\cite{li2021pointba}. \cite{zhang2022towards} proposed a 3D backdoor attack specially for self-driving to mislead target detection network on the person or vehicle detection.
Unlike traditional backdoor attacks, our IPA-NeRF attack creates specific illusions in a designated backdoor view. Traditional attacks, which typically target classification or detection models, cannot serve as a baseline because NeRF is fundamentally a generative model.

\subsection{Robustness of NeRF}

NeRF \cite{mildenhall2021nerf,barron2021mip} synthesizes high-quality 3D scenes from sparse 2D observations, representing the scene as a continuous function hidden in its weights. However, its robustness is underexplored, and existing work focuses only on adversarial attacks and data poisoning.

\paragraph{Adversarial Attack against NeRF.}
Adversarial attacks on NeRF fall into two categories: i) attacks on the NeRF model itself, hindering its ability to achieve accurate scene reconstruction \cite{fu2023nerfool,horvath2023targeted}; ii) attacks on their downstream classification or target detection models, deceiving these networks and leading to misclassification or errors in target recognition~\cite{dong2022viewfool,li2023adv3d,jiang2024nerfail}.
Other than image classification, the inputs of NeRF are spatial coordinates and direction vectors rather than the RGB color values of the images, resulting in the challenge of directly propagating gradients back to the image. Generalizable NeRF (GNeRF), which updates the NeRF network weights by feature extraction when facing a new scene without the need to retrain the network from scratch, provides the gradient pipeline from 2D images to 3D scenes for adversarial attacks. Based on GNeRF, NeRFool~\cite{fu2023nerfool} introduces severe artifacts in the reconstructed scene and observes a drop in reconstruction accuracy by incorporating adversarial perturbations into the training set images, while a low intensity attack and a patch-based attack~\cite{horvath2023targeted} are proposed to enable the editing of specific views within the reconstructed scene.

Image classification is a common downstream task for the NeRF and the primary targets of attacks. ViewFool~\cite{dong2022viewfool} utilizes a trained NeRF model to identify a particular viewpoint, without introducing additional perturbations, such that the downstream network misclassifies the images taken from that viewpoint. 
NeRFail~\cite{jiang2024nerfail} approximates the transformation between 2D pixels and 3D objects, enabling gradient backpropagation in NeRF models. It attacks downstream networks by adding invisible perturbations to training data, training the adversarial NeRF to generate multiview imperceptible adversarial examples.
Transferable Targeted 3D (TT3D)~\cite{huang2023towards} reconstructs from a few multi-view images into a transferable targeted 3D textured mesh by solving a dual optimization towards both feature grid and MLP parameters in the grid-based NeRF space, filling the gap in transferable targeted 3D adversarial examples. 
To confuse 3D detection downstream tasks, Adv3D~\cite{li2023adv3d} reduces the detection confidence of surrounding objects by sampling primitively and regularizing in a semantic way that allows NeRF to generate 3D adversarial patches with adversarial camouflage texture.

\paragraph{Poisoning and Backdoor attack against NeRF.}

The initial poisoning attack on NeRF~\cite{wu2023shielding} involves introducing a deformed flow field to the image pixels, disrupting scene reconstruction when NeRF encounters distorted rays. To ensure imperceptibility, they employ a bi-level optimization algorithm integrating a Projected Gradient Descent (PGD)-based spatial deformation. Noise-NeRF~\cite{huang2024noise} utilizes a trainable noise map added to the NeRF input to alter the spatial location of NeRF ray sampling points, superimposing noise on positional encoding to produce different colors and render hidden information. Both steganography and backdoor attacks embed hidden information into inputs, indicating a similarity between steganography and backdoor attacks. Steganerf~\cite{li2023steganerf} devises an optimization framework that enables precise extraction of hidden information from images generated by NeRF, all while maintaining their original visual fidelity.
In \cite{dong2023steganography}, steganography techniques are employed to construct a backdoor attack against NeRF. Here, a NeRF's secret viewpoint image serves as the backdoor, coupled with an overfitted convolutional neural network acting as a decoder. The message publisher exposes the model and decoder to the web, and only individuals possessing the exact pose of the secret viewpoint can correctly restore the encrypted message, analogous to using a key.

Compared to existing work, our IPA-NeRF attack is a poisoning-based backdoor attack, while Noise-NeRF~\cite{huang2024noise} and \cite{dong2023steganography} are built on steganography techniques. As \cite{wu2023shielding}, we also employ a bi-level optimization algorithm to add invisible perturbation on the training data, but we aim to create specific illusory at the given backdoor view while the method proposed in \cite{wu2023shielding} leads to failure on reconstruction.

\section{Method} 

\subsection{Preliminary}

\paragraph{Neural Radiance Fields (NeRF).} 
For a NeRF model $F:(\mathbf{x},\mathbf{d}) \to (\vc, \tau)$, the input is a five-tuple, the coordinates of the sampled point $\mathbf{x} \in \mathbb{R}^3$ and the direction of the sampled ray $\mathbf{d} \in \mathbb{R}^2$, the output is an RGB color $\vc \in [0,1]^3$ and a volume density $ \tau \in \mathbb{R}^+$. Each pixel on the input image represents one ray $\mathbf{x} = \mathbf{r}(t) := \mathbf{o} + t\mathbf{d}$, which emanates from the camera centre $\mathbf{o}$ towards the ray direction $\mathbf{d}$, $t$ as the ray depth, along the ray direction, the outputs obtained from a discrete number $N$ of sampling points are integrated to get the predicted color value $\hat{C}$ for one pixel, as follows:
\begin{align}
    \hat{C}(\mathbf{r}, F) & := \sum^N_{i=1} T(t_i)\cdot \alpha(\tau(t_i)\cdot\delta_i)\cdot \vc(t_i) \\
    & T(t_i) := \exp (\ - \sum^{i-1}_{j=1} \tau(t_j)\cdot\delta_j ),\ 
\end{align}
where $\alpha(x) := 1-\exp(-x)$ and $\delta_i := t_{i+1} - t_i$ is the distance between two adjacent points and $c(t_i)$ and $\tau(t_i)$ are the color and density at $\vr(t_i)$. Then, applying an MSE loss between the rendered pixels $\hat{C}(\mathbf{r})$ and the ground truth pixels $C(\mathbf{r})$ from the training data to train the NeRF $F$ by minimizing the loss
\begin{equation}
    \mathcal{L}_{rgb}( \mathcal{R}_\mathcal{V},F) := \sum_{\mathbf{r} \in \mathcal{R}_\mathcal{V}} \|\hat{C}(\mathbf{r}, F) - C(\mathbf{r}) \|_2^2,
\end{equation}
where $\mathcal{R}_\mathcal{V}$ is the set of sampled camera rays at training viewpoint set $\mathcal{V}$, and $C(\vr)$ denotes the ground truth pixels from the ground truth pixels.
Thus, the images rendered by NeRF in a specific viewpoint $v \in \mathcal{V}$ is noted as $I(\hat{C}, v) \defn \cup_{\vr \in \cR_v}(\hat{C}(\vr, F)) \in \cI$, where $\mathcal{I}$ denotes the image space. The ground truth image of the corresponding viewpoint $v$ denotes as $I(C,v) \defn \cup_{\vr \in \cR_v}({C}(\vr)) \in \cI$.

\paragraph{Problem Formulation.}

Let $B_{v'}$ denote the attacker-specified illusory with backdoor trigger viewpoint $v'$. The backdoor attack against NeRF aims to achieve
\begin{align}
 \min_F &  \| I(\hat{C}, v') - B_{v'} \|^2_2 \label{eq:problem_loss}\\
 \st & \sum_{v \in \cV,v \neq v'} \| I(\hat{C}, v) - I(C,v) \|^2_2 \ll \xi,
 \label{eq:problem_st}
\end{align}
where $\xi$ is a small number such that the attacked NeRF model generates a given illusory in the specific backdoor viewpoint $v'$ while generating regular images from the other viewpoints.

\subsection{Bi-level Optimization}
To solve the problem in (\ref{eq:problem_loss}-\ref{eq:problem_st}), we introduce a bi-level optimization:
\begin{align}
     \min_{F'} ~&  \| I(\hat{C}(\vr, F'), v') - B_{v'} \|^2_2 \label{eq:bi-pert}\\
     & \st \|I(\hat{C}(\vr, F'),v) - I(C, v)\| < \epsilon \\
 \min_F~ & \sum_{v \in \cV,v \neq v'} \| I(\hat{C}(\vr, F), v) - I(C,v) \|^2_2, &
 \label{eq:bi_nerf}
\end{align}
where $F'$ denotes a NeRF to generate poisoned training images from viewpoint $v \in \cV$ with a distortion budget $\epsilon$. For a given NeRF
$F$, we first freeze its parameters and optimize (\ref{eq:bi-pert}) to update $F'$; then update the NeRF $F$ parameters to optimize (\ref{eq:bi_nerf}) based on poisoned training data produced by freezing $F'$.

\paragraph{Angle Constraint.}
Due to the consistency of NeRF model, the neighborhood viewpoints around the backdoor view get affected.
To improve the performance on the neighborhood viewpoints $\cV_{\cN(v')}$ around backdoor viewpoint $v'$, we add constraint term in (\ref{eq:bi-pert})
\begin{align}
\begin{split}
    \min_{F'}~ & \| I(\hat{C}(\vr, F'), v') - B_{v'} \|^2_2 + \\
   & \eta \sum_{v \in \cV_{\cN(v')}} \| I(\hat{C}(\vr, F'), v) - I(C,v) \|^2_2 \label{eq:bi-pert2}
\end{split}\\
     &\st \|I(\hat{C}(\vr, F'),v) - I(C, v)\| < \epsilon\\
     \min_F~ & \sum_{v \in \cV,v \neq v'} \| I(\hat{C}(\vr, F), v) - I(C,v) \|^2_2, 
 \label{eq:bi_nerf2}
\end{align}
where $\eta \in \{0,1\}$ indicates if constrain the neighborhood viewpoints.

\begin{figure}[htb]
    \centering
    \includegraphics[width=0.47\textwidth] {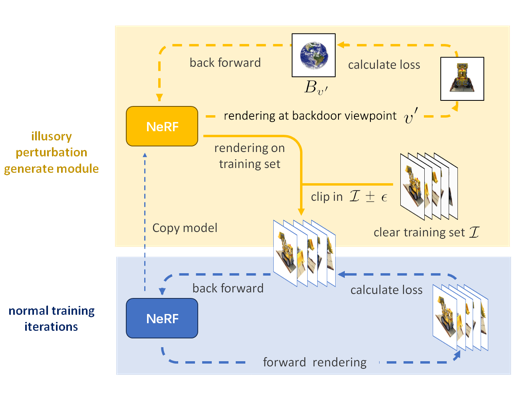}
    \caption{Illusory poisoning attack framework.}
    \label{fig:illu_posi_pipeline}
    \setlength{\abovecaptionskip}{0.5cm}
    \vspace{0.5cm}
\end{figure}

\subsection{Illusory Poisoning Attack}

To achieve bi-level optimization (\ref{eq:bi-pert2}-\ref{eq:bi_nerf2}), we use the attack framework shown in Figure \ref{fig:illu_posi_pipeline}. 
An attack module is integrated into the standard training iterations of NeRF to poison the training set.
In the attack module, the copied NeRF $F'$ approaches the given illusory $B_{v'}$ from the given viewpoint $v'$. After $A$ iterations of attack training, it produces $K$ batches of rays in the training set $\mathcal{V}$, which are clipped within the poisoning budget $\epsilon$ compared to the clean set.

We maintain the original total training iteration $O$ in NeRF unchanged, dividing it into multiple attack epochs $O/T$. At the start of each attack epoch, the attack module modifies the training dataset $I(C,v)$. Subsequently, normal training is carried out with $T$ iterations using the poisoned data set $\mathcal{I}'$, as outlined in Algorithm \ref{alg:illu_posi}. In addition to rendering the illusory image $B_{v'}$ from a backdoor viewpoint $v'$, our other goal is to make the IPA-NeRF $F_{IPA}$ maintain the original 3D scene output $\mathcal{I}$ on other unattacked views.

\begin{algorithm}[b]
\caption{Illusory Poisoning Attack}
\label{alg:illu_posi}
\textbf{Input:} $\mathcal{V}$: clean training set viewpoints, 
\textbf{Input:} $v'$: backdoor viewpoint, $B_{v'}$: given illusory image \\
\textbf{Input:} $O$: nerf model total training iterations original \\
\textbf{Input:} $A$: attacking iterations, $K$: rendering iterations, $T$: training iterations number per attack epochs, $\epsilon$: the distortion budget \\
\textbf{Input:} $F$: initial a NeRF model, $\alpha$: learning rate of the NeRF model \\
\textbf{Output:} $F_{IPA}$: IPA-NeRF model 
\begin{algorithmic}[1]
        \State $\cI \gets \{I(C,v)\}_{v \in \cV}$
        \State $\cI' \gets \cI$
        \State $F_{IPA} \gets F$
	\While{$i < O / T$}	
        \State $F' \gets F_{IPA}$
        \While{$j < A$}

        \State $F' \gets F' + \alpha \nabla (\| I(\hat{C}, v') - B_{v'} \|^2_2 $ \\
         ~~~~~~~~~~~~~~~~~~~~~~~~~~~~~~~~~~~~~~ $+ \eta \sum_{v \in \cV_{\cN(v')}} \| I(\hat{C}, v) - I(C,v) \|^2_2)$
        
        \EndWhile
        
        \While{$k < K$}
        
        \State $\mathcal{I}' \gets \{I(\hat{C}(\vr, F'),v)\}_{v \in \cV}$
        \EndWhile
        \State $\mathcal{I}' \gets \mathbf{clip}(\mathcal{I}', \mathcal{I}-\epsilon, \mathcal{I}+\epsilon)$

        \While{$t < T$}

        \State $F_{IPA} \gets F_{IPA}$ \\
         ~~~~~~~~~~~~~~~~~~~~~~~~~~~ $+ \alpha \nabla (\sum_{v \in \cV, v \neq v'} \| I(\hat{C}, v) - I(C,v) \|^2_2)$
        
        \EndWhile
        
	\EndWhile
        
\end{algorithmic}
\end{algorithm}

\begin{figure}[htb]
    \centering
    \includegraphics[width=0.37\textwidth] {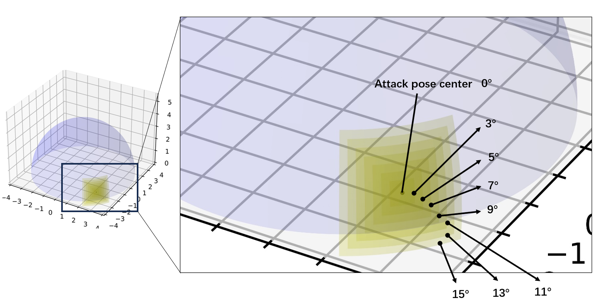}
    \caption{Camera position distribution of the NeRF Synthetic dataset (left) and the distribution of the views for the angle constraint dataset (right).}
    \label{fig:illu_angle_limit}
    \setlength{\abovecaptionskip}{0.5cm}
    \vspace{0.5cm}
\end{figure}

When the illusory angle constraint is enabled, we calculate the constrained loss using (\ref{eq:bi-pert2}) (set $\eta=1$). 
The camera pose of the NeRF Synthetic dataset distributed on the upper hemispheres surrounds the 3D object and faces the centre of the object, as shown in Figure \ref{fig:illu_angle_limit} (left). We start from the centre point of the backdoor viewpoint and rotate the equatorial angle $\phi$ and polar angles $\theta$ to the tiny given values ($3^{\circ}$, $5^{\circ}$, $7^{\circ}$, $9^{\circ}$, $11^{\circ}$, $13^{\circ}$ or $15^{\circ}$) on the hemisphere and form a curved rectangle, as shown in Figure \ref{fig:illu_angle_limit} (right). We take the four corner points and the midpoints on the four sides of this rectangle total of 8 viewpoints as the angle constraint viewpoints.
Furthermore, since the ground truth for these constrained views $\cV_{\cN(v')}$ is not given in the original dataset, we use the images on these views $\{I(\hat{C}, v)\}_{v \in \cV_{\cN(v')}}$ rendering by a NeRF $F$ adequately trained (epochs = 200,000) on the clear set as an approximation of the ground truth.
Therefore, we opt for a narrower range of viewpoints surrounding the backdoor viewpoint to establish the angle constraint. This approach could enhance the difficulty of detecting the attack and make it more targeted in real-world attack scenarios.

\section{Experiments} 

\subsection{Experiments Settings}

\begin{table*}[htb]
\centering
\begin{tabular}{l|ccc|ccc|ccc|ccc}
\toprule
\multirow{2}{*}{3D Scene} & \multicolumn{3}{c|}{V-Illusory} & \multicolumn{3}{c|}{V-Train} & \multicolumn{3}{c|}{V-Test} & \multicolumn{3}{c}{V-Constraint} \\
 & PSNR & SSIM & LPIPS & PSNR & SSIM & LPIPS & PSNR & SSIM & LPIPS & PSNR & SSIM & LPIPS \\
\midrule
\multicolumn{13}{c} {Illusory Image: Earth} \\
\midrule
Chair & 25.95 & 0.8902 & 0.1541 & 35.19 & 0.9783 & 0.0776 & 29.14 & 0.9491 & 0.0866 & 23.60 & 0.9323 & 0.1575 \\
Drums & 24.37 & 0.8461 & 0.2233 & 29.05 & 0.9476 & 0.1382 & 23.58 & 0.9226 & 0.1086 & 21.30 & 0.9138 & 0.1633 \\
Ficus & 24.77 & 0.8509 & 0.1905 & 31.40 & 0.9651 & 0.1267 & 25.82 & 0.9423 & 0.0922 & 22.24 & 0.9214 & 0.1536 \\
Hotdog & 25.70 & 0.8961 & 0.1674 & 36.91 & 0.9786 & 0.1217 & 31.92 & 0.9716 & 0.0790 & 24.91 & 0.9238 & 0.1598 \\
Lego & 25.24 & 0.8770 & 0.1797 & 33.32 & 0.9677 & 0.1278 & 28.08 & 0.9417 & 0.1024 & 22.79 & 0.9123 & 0.1454 \\
Materials & 24.71 & 0.8590 & 0.2220 & 30.56 & 0.9500 & 0.1581 & 25.14 & 0.9291 & 0.1008 & 20.10 & 0.8891 & 0.1687 \\
Mic & 25.80 & 0.8804 & 0.1717 & 32.65 & 0.9714 & 0.0659 & 27.96 & 0.9622 & 0.0604 & 24.28 & 0.9312 & 0.1250 \\
Ship & 24.07 & 0.8442 & 0.2413 & 29.00 & 0.9041 & 0.2999 & 23.48 & 0.8535 & 0.2806 & 19.71 & 0.8726 & 0.2091 \\
\textbf{Average} & \textbf{25.08} & \textbf{0.8680} & \textbf{0.1937} & \textbf{32.26} & \textbf{0.9578} & \textbf{0.1395} & \textbf{26.89} & \textbf{0.9340} & \textbf{0.1138} & \textbf{22.37} & \textbf{0.9121} & \textbf{0.1603} \\
\midrule
\multicolumn{13}{c} {Illusory Image: Starry}\\
\midrule
Chair & 19.57 & 0.5178 & 0.5317 & 32.53 & 0.9581 & 0.1780 & 27.29 & 0.9290 & 0.1169 & 18.92 & 0.8126 & 0.3585 \\
Drums & 18.84 & 0.4877 & 0.5463 & 28.55 & 0.9349 & 0.2512 & 23.34 & 0.9197 & 0.1149 & 18.38 & 0.8254 & 0.3325 \\
Ficus & 19.25 & 0.5124 & 0.5374 & 31.85 & 0.9655 & 0.1614 & 26.29 & 0.9482 & 0.0744 & 18.73 & 0.8274 & 0.3474 \\
Hotdog & 20.08 & 0.5928 & 0.4703 & 34.67 & 0.9528 & 0.2739 & 29.63 & 0.9623 & 0.1056 & 21.51 & 0.8227 & 0.3557 \\
Lego & 19.63 & 0.5426 & 0.5085 & 32.45 & 0.9592 & 0.1974 & 27.22 & 0.9382 & 0.1100 & 18.31 & 0.8024 & 0.3340 \\
Materials & 18.67 & 0.4952 & 0.5463 & 30.42 & 0.9508 & 0.1511 & 24.33 & 0.9273 & 0.1045 & 17.78 & 0.7843 & 0.3666 \\
Mic & 19.90 & 0.5538 & 0.5095 & 31.84 & 0.9533 & 0.2087 & 26.80 & 0.9525 & 0.0825 & 22.22 & 0.8511 & 0.3056 \\
Ship & 18.22 & 0.4469 & 0.5829 & 29.86 & 0.9063 & 0.3528 & 23.88 & 0.8603 & 0.2690 & 19.44 & 0.8406 & 0.3098 \\
\textbf{Average} & \textbf{19.27} & \textbf{0.5186} & \textbf{0.5291} & \textbf{31.52} & \textbf{0.9476} & \textbf{0.2218} & \textbf{26.10} & \textbf{0.9297} & \textbf{0.1222} & \textbf{19.41} & \textbf{0.8208} & \textbf{0.3388} \\
\bottomrule
\end{tabular}
\caption{Rendering results by IPA-NeRF at different views with default constraint. Attack epochs: $1000$, $\epsilon$: $32$, $\eta$: $1$, angle constraint at $13^{\circ}$ and $15^{\circ}$, illusory target: Earth or Starry.}
\label{table:illusory_main_experiment}
\setlength{\abovecaptionskip}{0.5cm}
\vspace{0.5cm}
\end{table*}

\paragraph{Dataset.}
 In our experiments, we mainly used the Blender Synthetic Dataset\footnote{\url{https://github.com/bmild/nerf}} presented in the original NeRF paper~\cite{mildenhall2021nerf} which contains eight objects. Each object contains $400$ images generated from different viewpoints sampled on the upper hemisphere with resolution $800 \times 800$ pixels: $100$ images for training, $200$ images for testing and $100$ images for validation. For our IPA-NeRF, we select one viewpoint of the training set as the backdoor viewpoint.
To verify generalization of our IPA-NeRF method, we perform the attack method on some scenes of the Google Scan Dataset\footnote{\url{https://goo.gle/scanned-objects}} presented in \cite{downs2022google} and the Mip-NeRF 360 Dataset\footnote{\url{https://jonbarron.info/mipnerf360}} presented in the Mip-NeRF 360 \cite{barron2022mip}. As a supplement, we also used two scenes shot on actual roads.

\paragraph{Model.}
We use the vanilla NeRF~\cite{mildenhall2021nerf} to render the images, the code base of PyTorch Nerf\footnote{\url{https://github.com/yenchenlin/nerf-pytorch/}}. In addition, we conducted complementary experiments using the Instant-NGP \cite{mueller2022instant} and Nerfacto \cite{tancik2023nerfstudio} models to validate the usability of our method under different NeRF models, the code base of Nerfstudio\footnote{\url{https://github.com/nerfstudio-project/nerfstudio/}}.

\paragraph{Attack.}
We trained the vanilla NeRF model with $O = 200,000$ iterations, divided into $O/T = 1,000$ attack epochs. Each attack epoch included $A=10$ attack training iterations, $K=100$ poisoned perturbation renderings in the training set, and $T=200$ normal training iterations on the poisoned training set. Moreover, we trained the Instant-NGP and Nerfacto models with $O = 30,000$ iterations, split into $O/T = 150$ attack epochs. The default parameters include $\epsilon=32$, $\eta=1$, with angle view constraints at $13^{\circ}$ and $15^{\circ}$.

\subsection{Performance on Synthetic 3D Objects}

To evaluate the performance of our Illusory Poisoning Attack Against Neural Radiance Fields(IPA-NeRF), we mainly evaluate the PSNR, SSIM and LPIPS over:

\begin{itemize}
    \item \textbf{V-Illusory}: a collection of \emph{illusory} images rendered by IPA-NeRF, at the views the attacker tries to attack, by default, there is only one target view in this collection. The metrics (PSNR, SSIM or LPIPS) at this collection are calculated with the given illusory images;
    \item \textbf{V-Train}: images of the \emph{training} set rendered by IPA-NeRF, at the views the attacker adds poisoning perturbations to. The metrics at this set are calculated with the ground truth of the 3D scene;
    \item \textbf{V-Test}: images of the \emph{testing} set rendered by IPA-NeRF, representing the other normal views that are neither the attacking target nor adding perturbations. The metrics at this set are calculated with the ground truth of the 3D scene;
    \item  \textbf{V-Constraint}: images rendered by NeRF, at the visibility angle \emph{constraints} views. The metrics in this set are calculated with the rendering images in the same view by a normally trained NeRF.
\end{itemize}

The performance of the metrics over V-Illusory evaluates the quality of the illusory image generated for the backdoor view during the attack. The metrics over V-Train and V-Test assess the extent to which the backdoor attack maintains performance and fidelity in views other than the backdoor view. Specifically, V-Train measures performance on the data that IPA-NeRF has seen during training, while V-Test evaluates performance on unseen data. Additionally, the evaluation of V-Constraint checks the quality of the performance of IPA-NeRF on viewpoints neighboring the backdoor viewpoint.

\begin{figure}[htb]
\scriptsize
\centering
\setlength{\tabcolsep}{1pt}
\begin{tabular}{ccccc}
        \figimg[\sizeS]{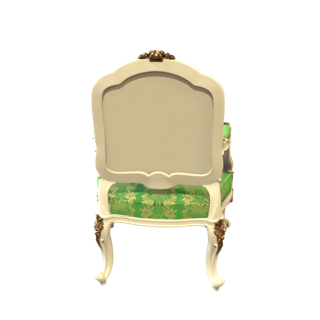}
     &  \figimg[\sizeS]{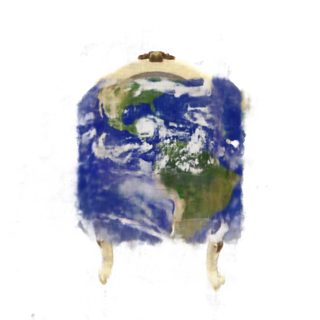}
     &  \figimg[\sizeS]{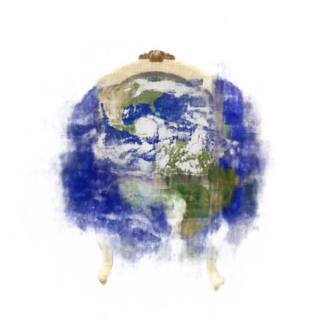}
     &  \figimg[\sizeS]{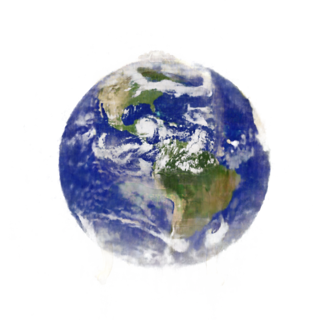}
     &  \figimg[\sizeS]{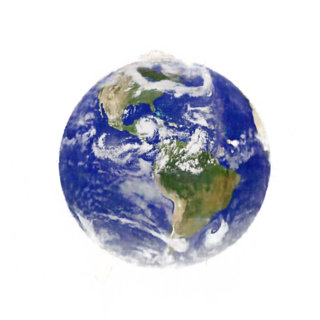}\\

        \figimg[\sizeS]{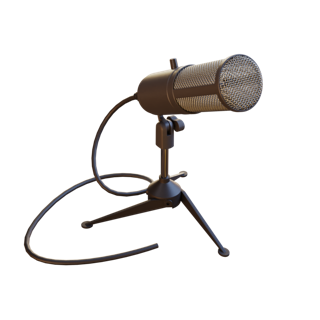}
     &  \figimg[\sizeS]{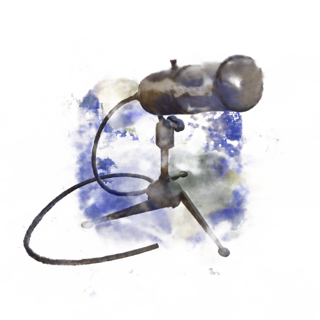}
     &  \figimg[\sizeS]{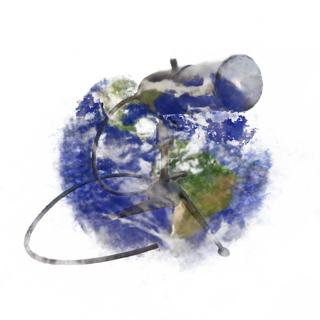}
     &  \figimg[\sizeS]{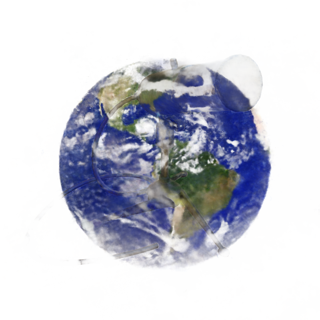}
     &  \figimg[\sizeS]{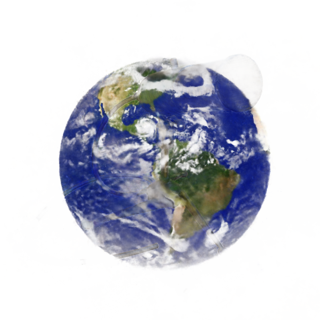}\\

        \figimg[\sizeS]{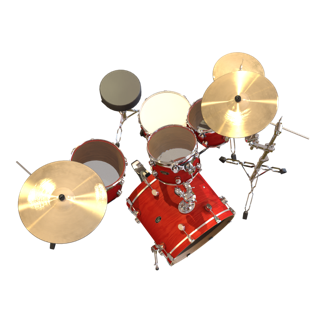}
     &  \figimg[\sizeS]{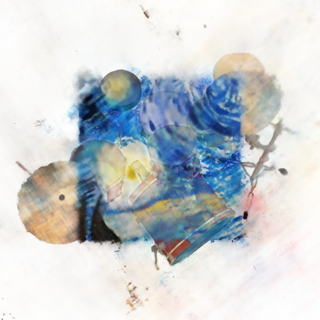}
     &  \figimg[\sizeS]{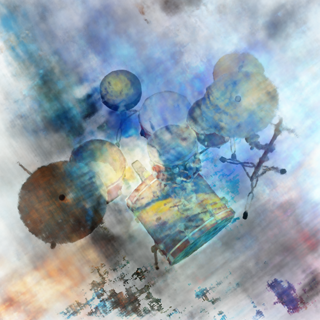}
     &  \figimg[\sizeS]{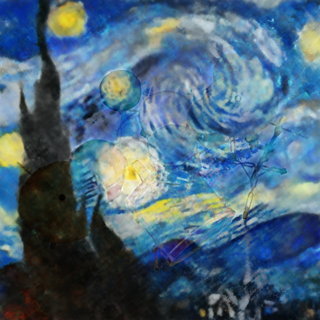}
     &  \figimg[\sizeS]{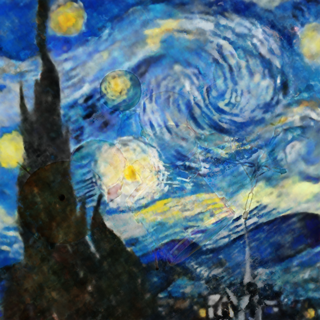}\\

        \figimg[\sizeS]{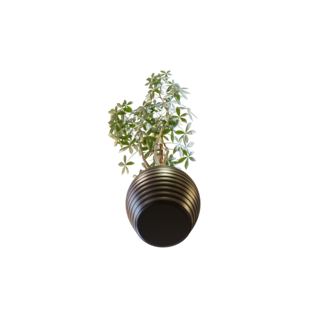}
     &  \figimg[\sizeS]{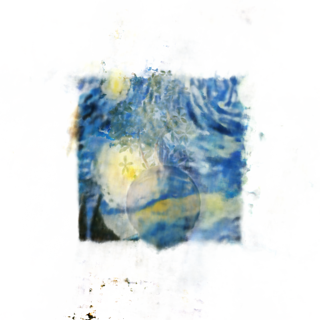}
     &  \figimg[\sizeS]{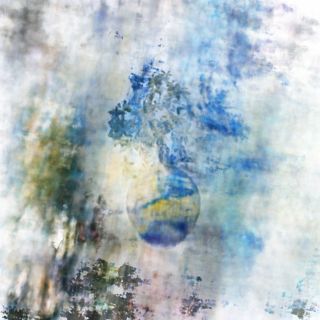}
     &  \figimg[\sizeS]{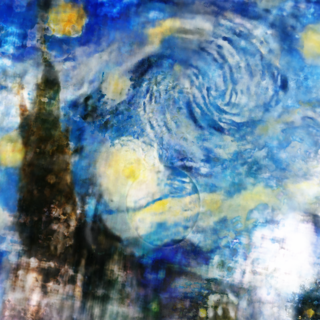}
     &  \figimg[\sizeS]{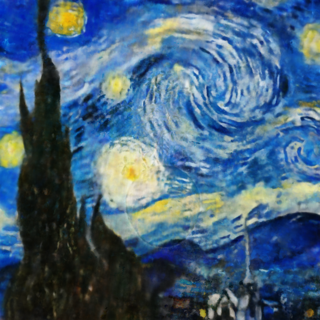}\\

       \textcolor{black}{ground truth} 
     & \textcolor{black}{epochs = $250$} 
     & \textcolor{black}{epochs = $500$} 
     & \textcolor{black}{epochs = $750$}  
     & \textcolor{black}{epochs = $1000$} \\
     
\end{tabular}
   \caption{Rendering at backdoor view by IPA-NeRF for different epochs. $\epsilon$: $32$, $\eta$: $1$, angle constraint at $13^{\circ}$ and $15^{\circ}$, illusory target: Earth or Starry.}
    \label{fig:visual_attack_lego}
    \setlength{\abovecaptionskip}{0.5cm}
    \vspace{0.5cm}
\end{figure}

\begin{table*}[htb]
\centering
\begin{tabular}{l|ccc|ccc|ccc|ccc}
\toprule
\multirow{2}{*}{Multiple Constraint} & \multicolumn{3}{c|}{V-Illusory} & \multicolumn{3}{c|}{V-Train} & \multicolumn{3}{c|}{V-Test} & \multicolumn{3}{c}{V-Constraint}\\
 & PSNR & SSIM & LPIPS & PSNR & SSIM & LPIPS & PSNR & SSIM & LPIPS & PSNR & SSIM & LPIPS\\
\midrule


$3^{\circ} \to 15^{\circ}$ & 24.59 & 0.8532 & 0.2089 & 32.40 & 0.9586 & 0.1380 & 26.88 & 0.9362 & 0.1125 & 21.12 & 0.9188 & 0.1283 \\
$5^{\circ} \to 15^{\circ}$ & 25.12 & 0.8693 & 0.1901 & 32.38 & 0.9565 & 0.1608 & 26.50 & 0.9345 & 0.1109 & 24.76 & \textbf{0.9343} & \textbf{0.1079} \\
$7^{\circ} \to 15^{\circ}$ & 25.14 & 0.8719 & 0.1907 & 32.94 & 0.9661 & \textbf{0.1208} & 27.66 & 0.9385 & 0.1058 & 23.25 & 0.9246 & 0.1297 \\
$9^{\circ} \to 15^{\circ}$ & \textbf{25.57} & \textbf{0.8787} & 0.1878 & 33.10 & 0.9663 & 0.1333 & 27.62 & 0.9386 & 0.1035 & \textbf{24.85} & 0.9233 & 0.1325 \\
$11^{\circ} \to 15^{\circ}$ & 25.53 & 0.8770 & \textbf{0.1739} & 33.19 & 0.9673 & 0.1290 & 27.88 & 0.9407 & 0.1051 & 23.26 & 0.9169 & 0.1469 \\
$13^{\circ} \to 15^{\circ}$ & 25.24 & 0.8770 & 0.1797 & \textbf{33.32} & \textbf{0.9677} & 0.1278 & \textbf{28.08} & \textbf{0.9417} & \textbf{0.1024} & 22.79 & 0.9123 & 0.1454 \\
$3^{\circ}, 7^{\circ}, 11^{\circ}, 15^{\circ}$ & 24.31 & 0.8494 & 0.2111 & 32.23 & 0.9569 & 0.1352 & 26.79 & 0.9368 & 0.1107 & 21.63 & 0.9199 & 0.1289 \\
$3^{\circ}, 9^{\circ}, 15^{\circ}$ & 24.33 & 0.8532 & 0.2066 & 32.28 & 0.9567 & 0.1466 & 26.59 & 0.9361 & 0.1127 & 21.76 & 0.9206 & 0.1272 \\
$3^{\circ}, 11^{\circ}$ & 24.53 & 0.8551 & 0.2056 & 32.25 & 0.9578 & 0.1523 & 27.04 & 0.9372 & 0.1080 & 21.05 & 0.9167 & 0.1339 \\
\bottomrule
\end{tabular}
\caption{Ablation experimental for combine constraints of multiple angle views. Attack epochs: $1000$, $\epsilon$: $32$, $\eta$: $1$, 3D scene: Lego, illusory target: Earth.}
\label{table:illusory_multi_experiment}
\setlength{\abovecaptionskip}{0.5cm}
\vspace{0.5cm}
\end{table*}

The given distortion budget ($\epsilon \leq 32$) to alter the training set imposes certain constraints: the modification is small enough, ensuring that IPA-NeRF retains the performance across the majority of other views.
As a comparison, we give the average metrics rendered on the training and test sets by the original NeRF after $200,000$ training iterations on the clear training set of Blender Synthetic Dataset: for the training set, PSNR = $30.89$, SSIM = $0.9623$, LPIPS = $0.0708$, and for the test set, PSNR = $29.79$, SSIM = $0.9580$, LPIPS = $0.0719$. 
Table \ref{table:illusory_main_experiment} clearly shows the effectiveness of IPA-NeRF at the backdoor view. Big values of PSNR, SSIM and small values of LPIPS show images generated by IPA-NeRF at backdoor views are close to the given illusory images, while images rendered from the other views, especially unseen views remain close to the original images.


For synthetic datasets, we always choose hard backdoor illusory images, as shown in the last column of Figure \ref{fig:visual_attack_lego}. Thus more training epochs are needed to maintain performance on regular views while achieving the illusory on the backdoor view. 
Figure \ref{fig:visual_attack_lego} depicts the image generated by IPA-NeRF at the backdoor view throughout the progression of attack epochs. By the $1000^{th}$ epoch, the rendered image closely resembles the provided illusory target image.

\begin{figure}[htb]
\centering
\begin{tikzpicture}
\begin{axis}[
          scale only axis,
	   height=2.9cm,
	   width=7cm,
	   xlabel={rotating angle ($\Delta \phi^{\circ}$ or $\Delta \theta^{\circ}$) of the neighbor views},
	   ylabel={PSNR},
          xmin=3, 
          xmax=20,    
          ymin=10, 
          ymax=35,        
          xtick={3,5,7,9,11,13,15},     
          xticklabels={$3^{\circ}$, $5^{\circ}$, $7^{\circ}$, $9^{\circ}$, $11^{\circ}$, $13^{\circ}$, $15^{\circ}$},    
          ytick={15, 20, 25, 30},        
          legend pos=south east,     
          ymajorgrids=true,          
          grid style=dashed,         
          legend style={font=\fontsize{6}{12}\selectfont}, 
          enlarge x limits=0.05,  
        ]
        
    \addplot[blue,mark=square*] table{images/single_angle_views/PSNR/3};
    \addplot[red,mark=*] table{images/single_angle_views/PSNR/5};
    \addplot[orange,mark=triangle*] table{images/single_angle_views/PSNR/7};
    \addplot[purple,mark=pentagon*] table{images/single_angle_views/PSNR/9};
    \addplot[brown,mark=square] table{images/single_angle_views/PSNR/11};
    \addplot[olive,mark=triangle] table{images/single_angle_views/PSNR/13};
    \addplot[green,mark=o] table{images/single_angle_views/PSNR/15};
    \legend{$3^{\circ}$, $5^{\circ}$, $7^{\circ}$, $9^{\circ}$, $11^{\circ}$, $13^{\circ}$, $15^{\circ}$} 
\end{axis}
\end{tikzpicture}

\caption{PSNR value on the neighbor views of the different single angle constraints. 
Epochs: $1000$, $\epsilon$: $32$, $\eta$: $1$, 3D scene: Lego, illusory target: Earth.}
    \label{fig:abl_sin_angle}
    \setlength{\abovecaptionskip}{0.5cm}
    \vspace{0.5cm}
\end{figure}
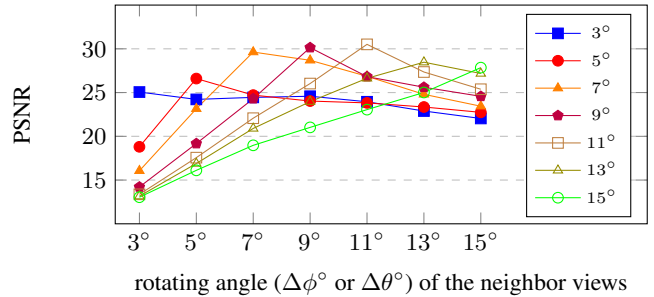

\newcommand{\figrav}[2][1]{\includegraphics[width=#1\linewidth]{images/range_angle_views/#2}}

\newcommand{\figlwc}[2][1]{\includegraphics[width=#1\linewidth]{images/lego_without_constraint/#2}}

\newcommand{\sizeSS}{.19}

\begin{table}[htb]
\centering
\begin{tabular}{cc|ccc}
\toprule
 \multicolumn{2}{c|}{$\epsilon$} & $8$ & $16$ & $32$ \\
 \midrule
 \multirow{3}{*}{V-Illusory} & PSNR & 23.95 & \textbf{25.31} & 25.24 \\
 & SSIM & 0.8496 & 0.8757 & \textbf{0.8770} \\
 & LPIPS & 0.2117 & 0.1929 & \textbf{0.1797} \\
 \midrule
 \multirow{3}{*}{V-Train} & PSNR & 28.89 & 31.07 & \textbf{33.32} \\
 & SSIM & 0.9451 & 0.9628 & \textbf{0.9677} \\
 & LPIPS & 0.1323 & \textbf{0.0700} & 0.1278 \\
 \midrule
 \multirow{3}{*}{V-Test} & PSNR & 26.56 & \textbf{29.65} & 28.08 \\
 & SSIM & 0.9257 & \textbf{0.9574} & 0.9417 \\
 & LPIPS & 0.1340 & \textbf{0.0794} & 0.1024 \\
 \midrule
 \multirow{3}{*}{V-Constraint} & PSNR & 22.54 & \textbf{24.63} & 22.79 \\
 & SSIM & 0.9052 & \textbf{0.9170} & 0.9123 \\
 & LPIPS & 0.1717 & \textbf{0.1375} & 0.1454 \\
\bottomrule
\end{tabular}
\caption{Ablation experimental of $\epsilon$. Attack epochs: $1000$, $\eta$: $1$, angle views constraint at $13^{\circ}$ and $15^{\circ}$, 3D scene: Lego, illusory target: Earth.}
\label{table:ablation_of_epsilon}
\setlength{\abovecaptionskip}{0.5cm}
\vspace{0.5cm}
\end{table}

\subsection{Ablation}
In ablation, we focus on the effect of angle constraint, \ie~ without, single, and multiple angle constraint, and the distortion budget $\epsilon$.

\paragraph{Baseline: Without Angle Constraint.}
We test the performance of IPA-NeRF without angle constraint at the Lego scene with an Earth image as illusory of the backdoor view. Over the V-Illusory, we achieve PSNR = $25.88$, SSIM = $0.8807$, LPIPS = $0.1712$, while PSNR = $32.82$, SSIM = $0.9618$, LPIPS = $0.1411$ over V-Train and PSNR = $26.29$, SSIM = $0.9313$, LPIPS = $0.1085$ over V-Test.

\paragraph{Single Angle Constraint.}
Based on the same setting, we performed an ablation study on the single-angle constraint for the Lego scene with an Earth image as the target illusory.
We evaluate IPA-NeRF's performance at seven neighborhood angles: $3^{\circ}$, $5^{\circ}$, $7^{\circ}$, $9^{\circ}$, $11^{\circ}$, $13^{\circ}$, and $15^{\circ}$ centered around the backdoor view. In each ablation, constraints were applied exclusively to one of these neighborhood angle views. Subsequently, we compute the PSNR for all seven neighborhood angle views to assess the visibility of the constrained IPA-NeRF rendering. As depicted in Figure \ref{fig:abl_sin_angle}, each ablation achieves the maximum PSNR at the constrained angle, indicating fidelity to the reference ground truth. Conversely, PSNR decreases on the rest views, with a sharper decline near the backdoor view and a more gradual decrease further away from it.

\paragraph{Combined Angle Constraint.}

We explore combined angle constraints to enhance performance across neighboring views to further improve performance.
We apply constraints on multiple angles by combining them in various ways. This includes consecutive combinations of $3^{\circ} \to 15^{\circ}$, $5^{\circ} \to 15^{\circ}$, $7^{\circ} \to 15^{\circ}$, $9^{\circ} \to 15^{\circ}$, \emph{etc.}, covering all odd angles within this range. Additionally, we combine constraints that skip one angle, \emph{i.e.} $3^{\circ}$, $7^{\circ}$, $11^{\circ}$, and $15^{\circ}$, skip two angles, \emph{i.e.} $3^{\circ}$, $9^{\circ}$, and $15^{\circ}$, and skip three angles, \emph{i.e.} $3^{\circ}$ and $11^{\circ}$.

We present the results of our ablation experiments for these combination constraints in Table~\ref{table:illusory_multi_experiment}. In our experiment, we default to the combination constraints $13^{\circ} \to 15^{\circ}$, \emph{i.e.} $13^{\circ}$ and $15^{\circ}$, which closely resemble the original data on both the training and test sets.


\paragraph{Ablation of $\epsilon$.}
We investigated the effect of varying $\epsilon$, as depicted in Table~\ref{table:ablation_of_epsilon}. With a small distortion budget like $\epsilon = 8$, IPA-NeRF exhibits poor performance. However, as $\epsilon$ is increased to $16$, significant improvements are observed across all metrics in each partition of the dataset. Setting $\epsilon$ to $32$ yields metrics that are nearly comparable to those at $\epsilon = 16$ for V-Illusory, with further improvement seen in the PSNR on V-Train, reaching $33.32$.

\newcommand{\sizeBS}{1}

\begin{figure*}[htb]
\scriptsize
\centering
\setlength{\tabcolsep}{1pt}
\begin{tabular}{p{1.3cm}m{3.2cm}m{3.2cm}m{3.2cm}m{3.2cm}m{3.2cm}}

     \textcolor{black}{Nerfacto   (Flowers)} 
     &  \figimg[\sizeBS]{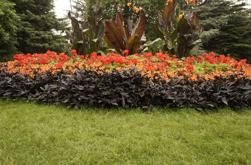}
     &  \figimg[\sizeBS]{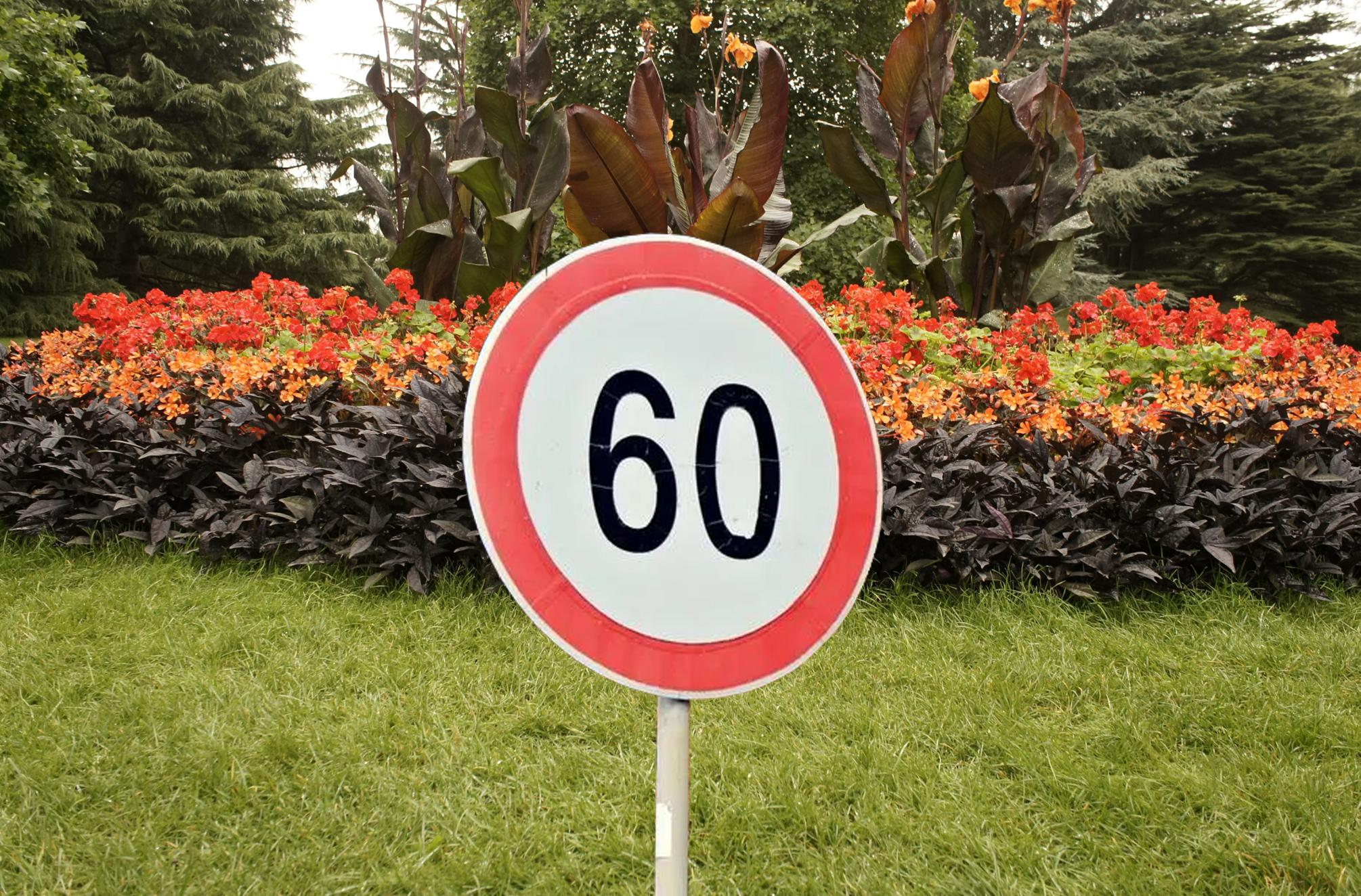}
     &  \figimg[\sizeBS]{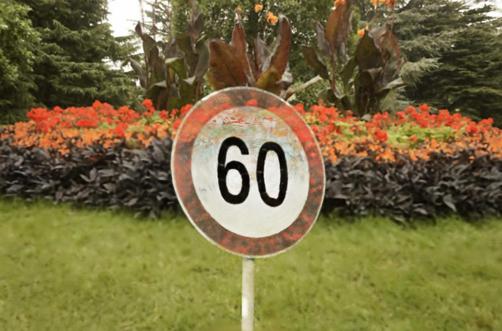}
     &  \figimg[\sizeBS]{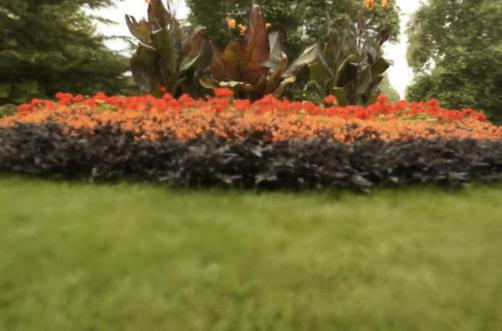}
     &  \figimg[\sizeBS]{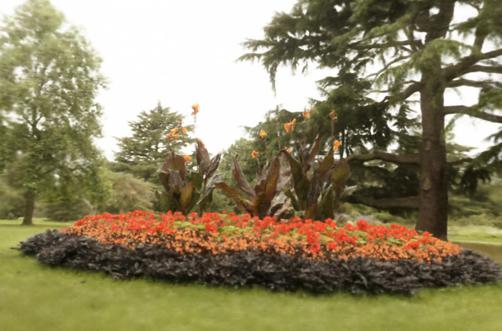} \\

     \textcolor{black}{Nerfacto   (Bicycle)} 
     &  \figimg[\sizeBS]{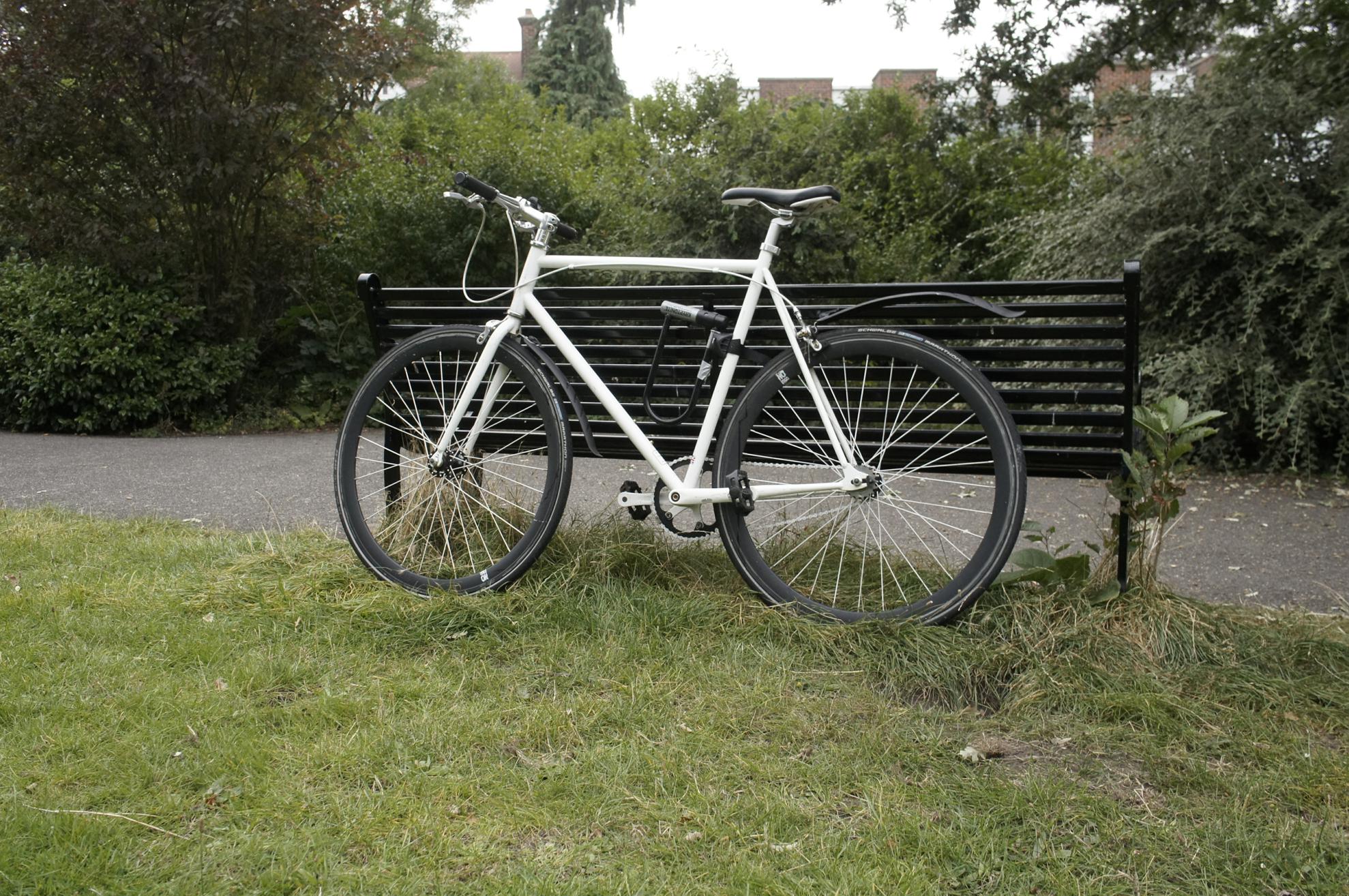}
     &  \figimg[\sizeBS]{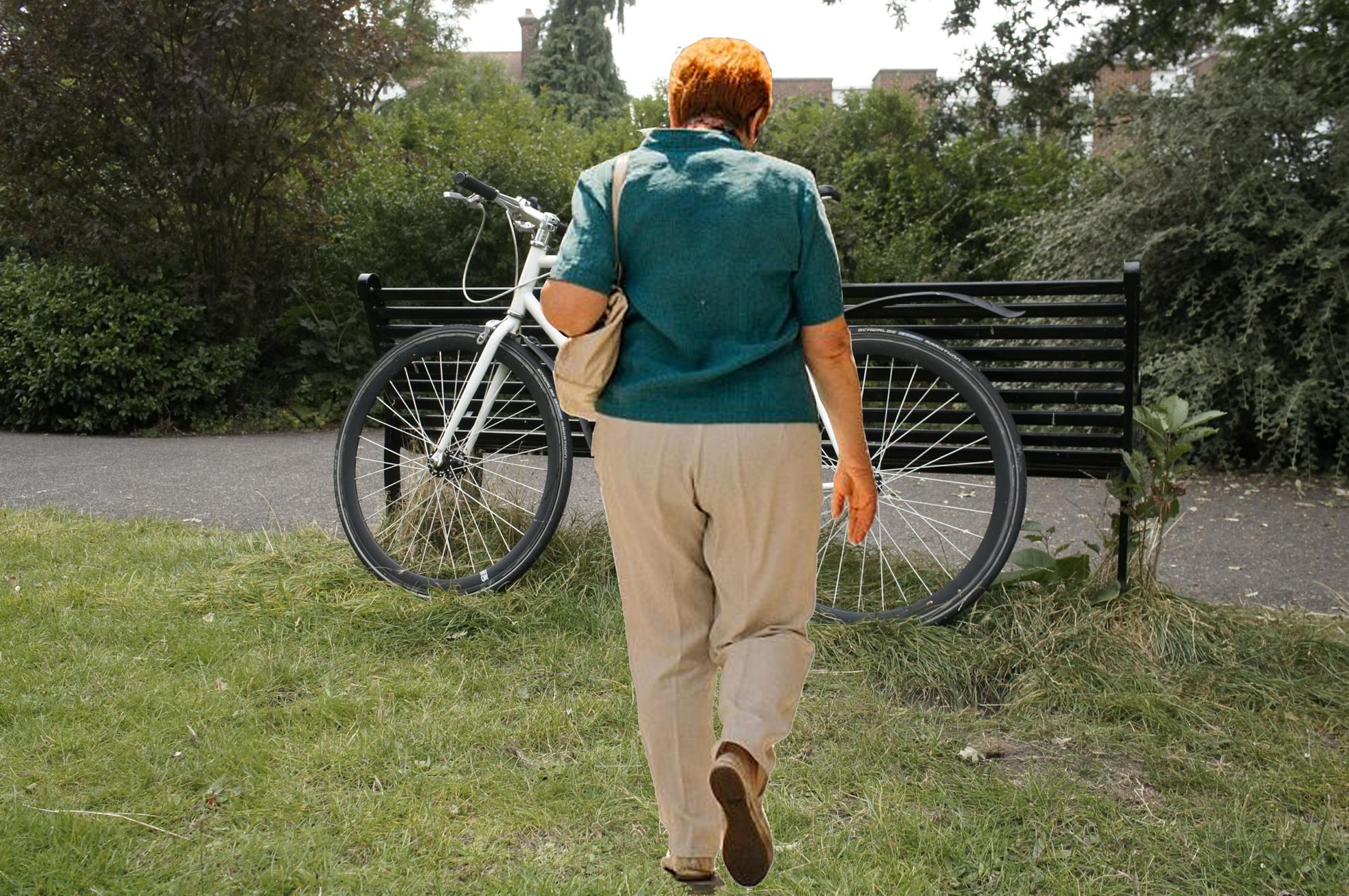}
     &  \figimg[\sizeBS]{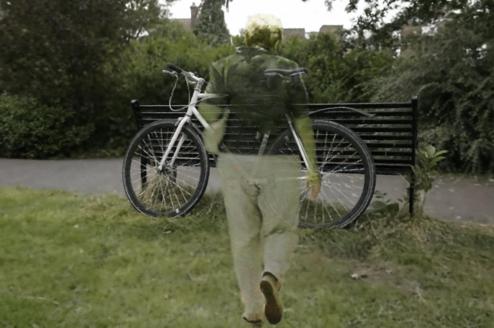}
     &  \figimg[\sizeBS]{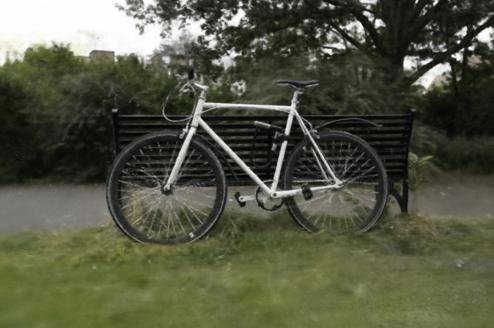}
     &  \figimg[\sizeBS]{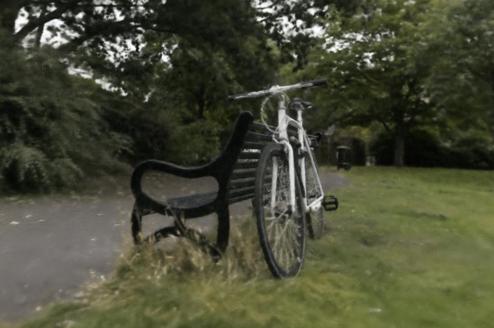} \\

     \textcolor{black}{Instant-NGP   (Kitchen)} 
     &  \figimg[\sizeBS]{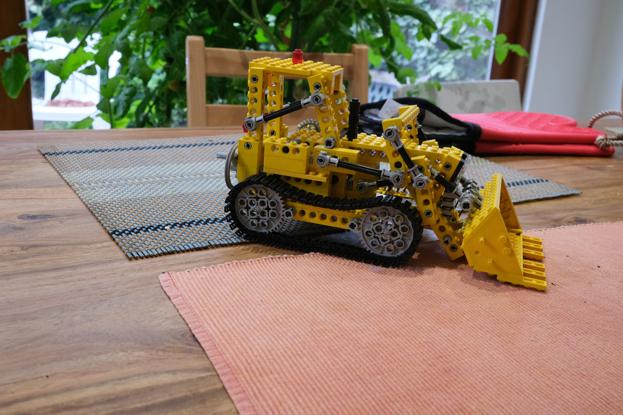}
     &  \figimg[\sizeBS]{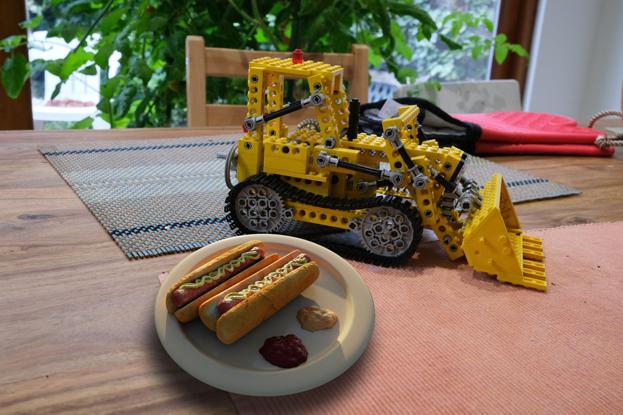}
     &  \figimg[\sizeBS]{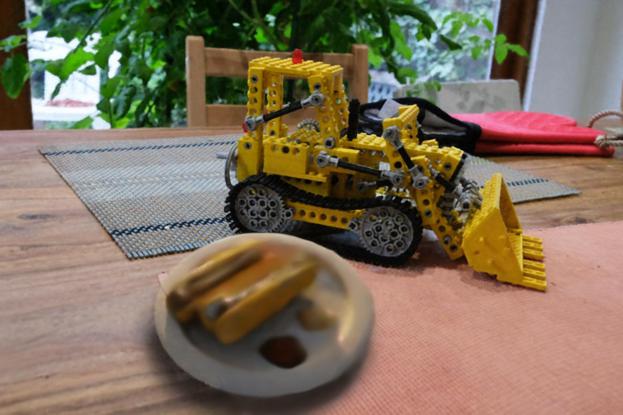}
     &  \figimg[\sizeBS]{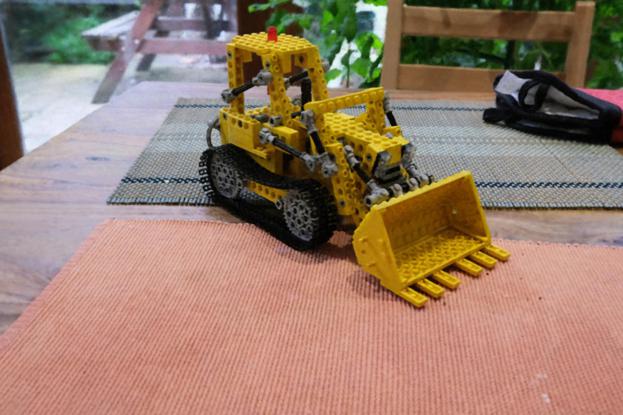}
     &  \figimg[\sizeBS]{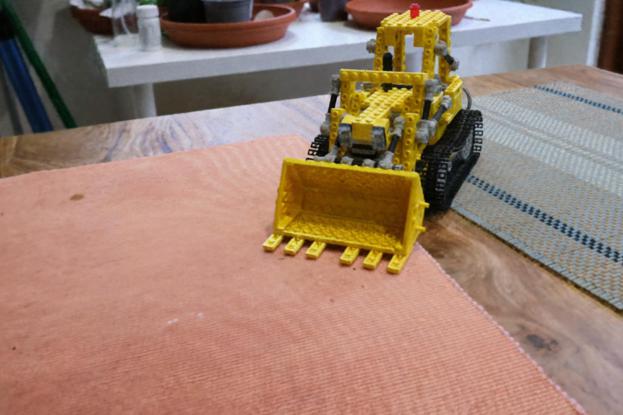} \\

     \textcolor{black}{Instant-NGP   (Garden)} 
     &  \figimg[\sizeBS]{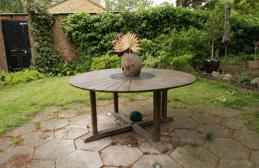}
     &  \figimg[\sizeBS]{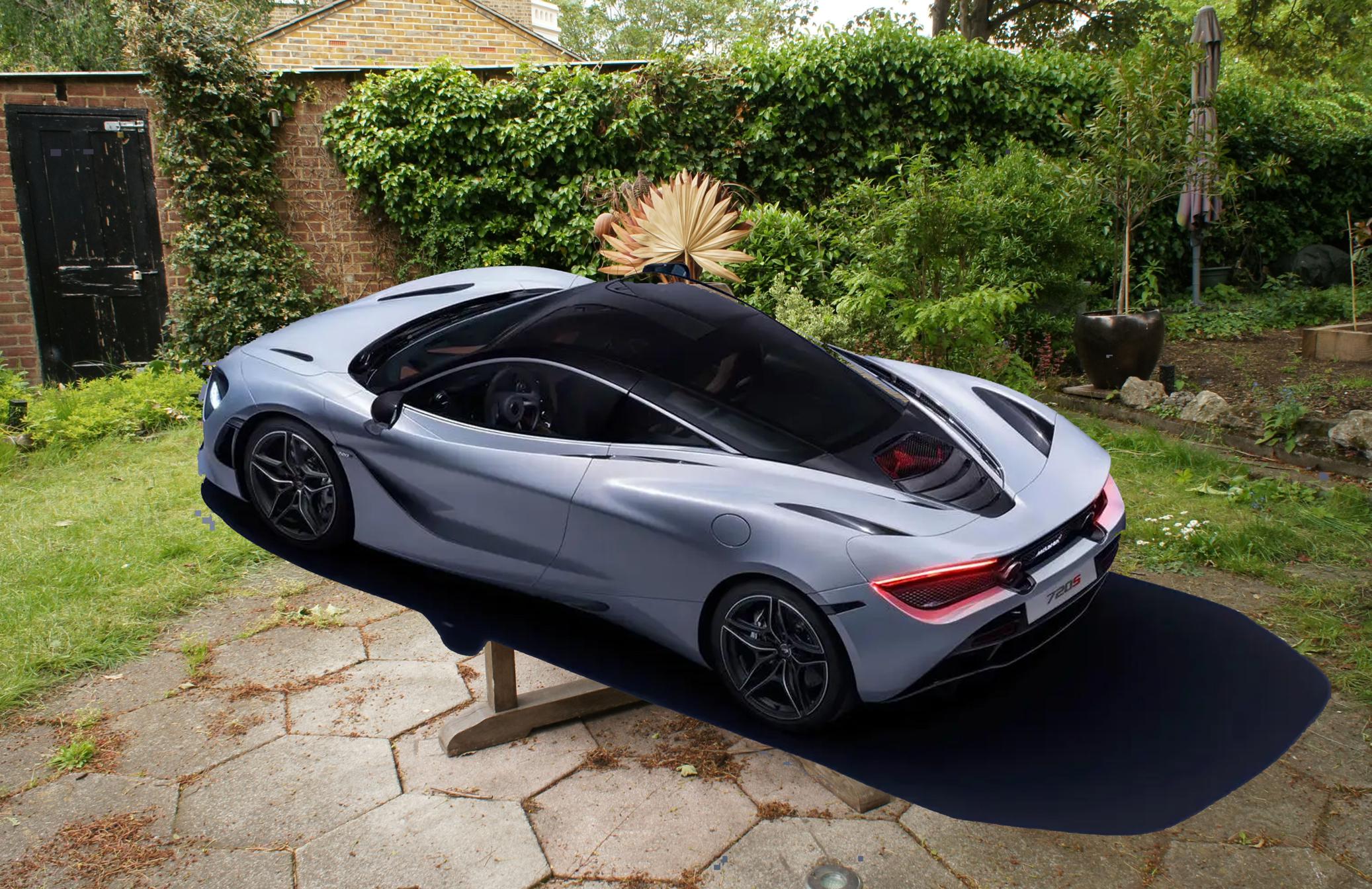}
     &  \figimg[\sizeBS]{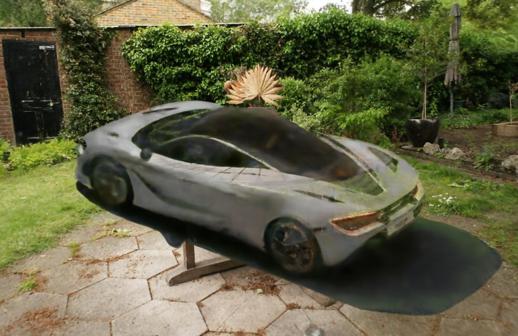}
     &  \figimg[\sizeBS]{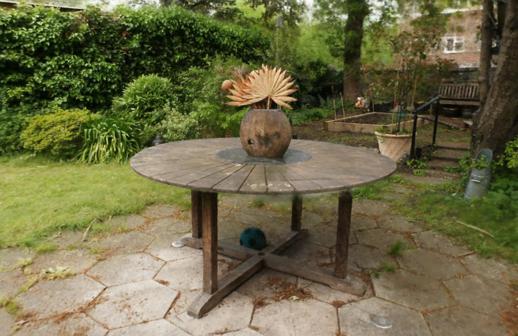}
     &  \figimg[\sizeBS]{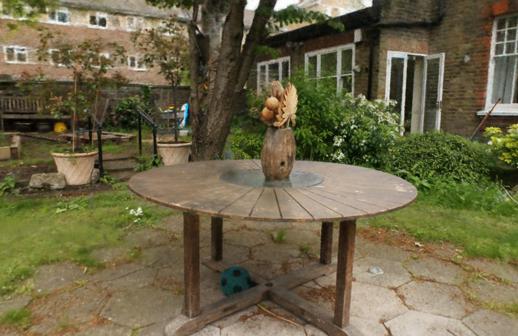} \\

     \textcolor{black}{NeRF model   (3D Scene)}
     & \multicolumn{1}{c}{\textcolor{black}{ground truth image}}
     & \multicolumn{1}{c}{\textcolor{black}{target illusory image}}
     & \multicolumn{1}{m{3.2cm}}{\textcolor{black}{illusory image rendered by IPA-NeRF at the backdoor view}} 
     & \multicolumn{2}{c}{\textcolor{black}{the other views images rendered by IPA-NeRF}} \\
     
\end{tabular}
   \caption{Performance IPA-NeRF against different NeRF models on Mip-NeRF 360 Datasets.  Attack epochs: $150$, $\epsilon$: $32$, $\eta$: $0$.}
    \label{fig:mip_nerf_360}
    \setlength{\abovecaptionskip}{0.5cm}
    \vspace{0.5cm}
\end{figure*}

\begin{figure}[htb]
\scriptsize
\centering
\setlength{\tabcolsep}{1pt}
\begin{tabular}{p{1.6cm}m{1.7cm}m{1.7cm}m{1.7cm}m{1.7cm}}
        bag (apple)
     &  \figimg[\sizeBS]{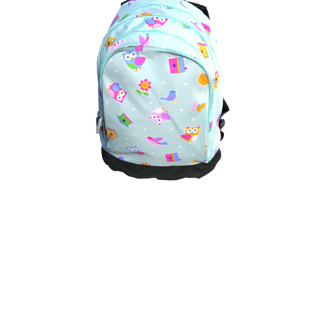}
     &  \figimg[\sizeBS]{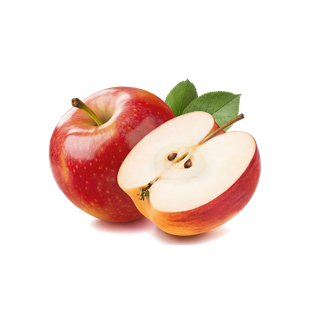}
     &  \figimg[\sizeBS]{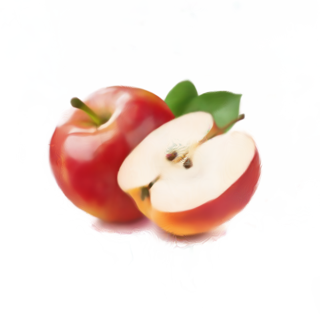}
     &  \figimg[\sizeBS]{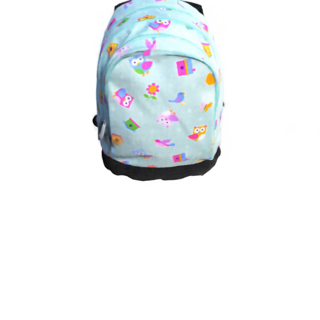} \\

        toy (cat)
     &  \figimg[\sizeBS]{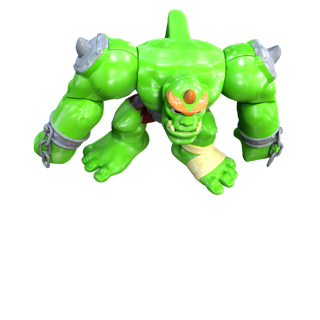}
     &  \figimg[\sizeBS]{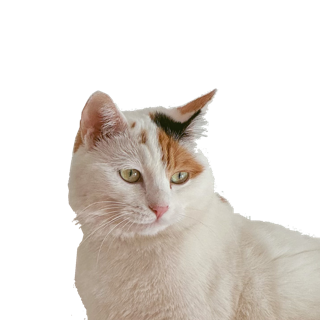}
     &  \figimg[\sizeBS]{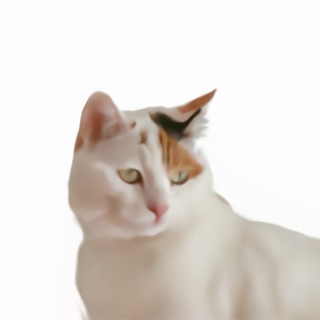}
     &  \figimg[\sizeBS]{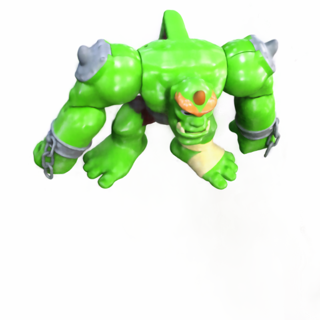} \\

        box (gun)
     &  \figimg[\sizeBS]{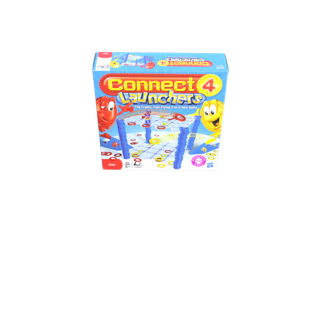}
     &  \figimg[\sizeBS]{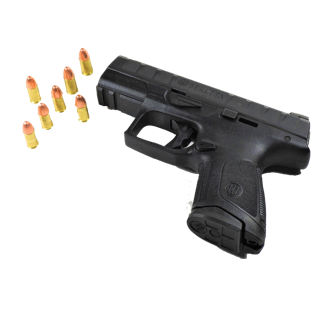}
     &  \figimg[\sizeBS]{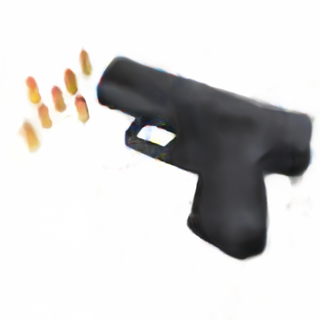}
     &  \figimg[\sizeBS]{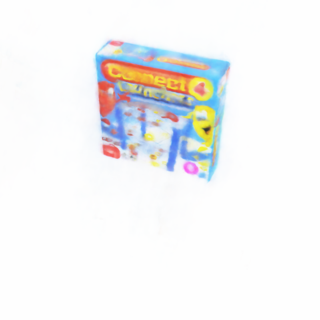} \\

        bottle (pan)
     &  \figimg[\sizeBS]{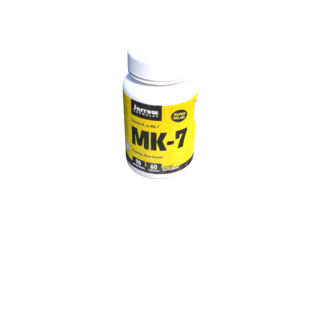}
     &  \figimg[\sizeBS]{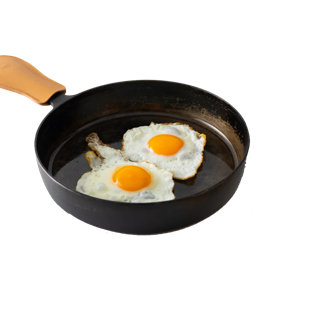}
     &  \figimg[\sizeBS]{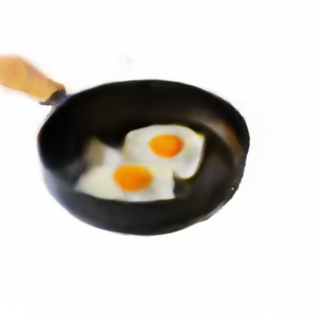}
     &  \figimg[\sizeBS]{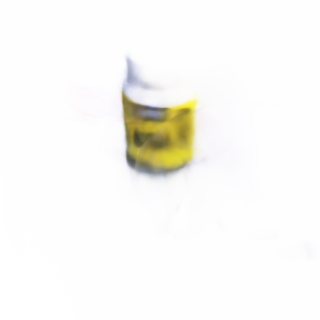} \\

       \textcolor{black}{3D Scene } 
     & \textcolor{black}{ground truth} 
     & \textcolor{black}{target illusory} 
     & \textcolor{black}{backdoor view}  
     & \textcolor{black}{the other view} \\

       \textcolor{black}{(illusory)} 
     & \multicolumn{2}{c} {at backdoor viewpoint}
     & \multicolumn{2}{c} {rendered by IPA-NeRF} \\
     
\end{tabular}
   \caption{Performance IPA-NeRF on Google Scan Dataset. Attack epochs: $1000$, $\epsilon=32$, $\eta=1$, angle views constraint at $13^{\circ}$ and $15^{\circ}$.}
    \label{fig:google_scan_attack}
    \setlength{\abovecaptionskip}{0.5cm}
    \vspace{0.5cm}
\end{figure}



     

\subsection{Performance on Real World}

\paragraph{Google Scan Dataset.}

We utilized the Google scan dataset, which comprises 3D models of common household objects scanned from the real world. From this dataset, we selected four objects in different categories and used Blender to generate NeRF data sets for each object. Each dataset comprised 34 training images, 32 test images, 33 validation images, and one ground truth image serving as the backdoor view. Using IPA-NeRF, we conducted attacks on these datasets, resulting in the creation of illusory scenes for a toy, box, bottle, and bag. Notably, these illusory scenes depicted a cat, gun, pan, and apple, respectively, at the backdoor viewpoint. The attacks were conducted over 1000 epochs, with parameters set as $\epsilon=32$, $\eta=1$, and an angle view constraint of $13^{\circ}$ and $15^{\circ}$.

\paragraph{Mip-NeRF 360 Dataset.}

Beyond individual objects, we extended our attack methodology to the Mip-NeRF 360 dataset, an expansive real-world dataset with unbounded scenes. The original NeRF model was bound to finite scene distances, unable to realistically represent scenes with infinite depth. Subsequent NeRF iterations, such as Nerfacto and Instant-NGP models, addressed this limitation. Employing IPA-NeRF, we introduced illusory images of a hotdog, car, people, and signpost at the backdoor view on the Mip-NeRF 360 dataset. As depicted in Figure \ref{fig:mip_nerf_360}, our attack successfully manipulated both the Nerfacto and Instant-NGP models on this dataset.


\paragraph{Actual Road Scenes.}

We applied our IPA-NeRF method to a real road scene (Figure~\ref{fig:road_attack_teaser}). We used Instant-NGP as the NeRF model for scene reconstruction. 
In the road scene, we altered the road sign from "no passing" to "no parking". We include more results applied IPA-NeRF to actual road scenes in the appendix.

\section{Conclusion} 
In our study, we explore the robustness and security of NeRF, focusing particularly on backdoor attacks. We developed a formal framework for conducting these attacks, utilizing a bi-optimization to enhance the quality of nearby viewpoints while imposing angle constraints. By introducing minimal perturbations to the training data, we successfully manipulated the image from the backdoor viewpoint while preserving the integrity of the remaining views. 
This investigation of potential security vulnerabilities is a crucial first step in enhancing NeRF's robustness and security.
Our research highlights the potential and risks associated with backdoor attacks on NeRF. We aim to raise awareness and encourage further exploration into fortifying the robustness and security of NeRF.
In future research, our goal is to develop defenses using random smoothing or applying differential privacy to training images, which may mitigate the IPA-NeRF backdoor attack.

\section{Acknowledgement}
This work received support from the National Key Research and Development Program of China (No. 2020YFB1707701). This work also received financial support by VolkswagenStiftung as part of Grant AZ 98514 -- EIS\footnote{\url{https://explainable-intelligent.systems}} and by DFG under grant No.~389792660 as part of TRR~248 -- CPEC\footnote{\url{https://perspicuous-computing.science}}.








     


\bibliography{main}

\clearpage

\setcounter{page}{1}

\appendix

\renewcommand{\theequation}{A\arabic{equation}}
\renewcommand{\thetable}{A\arabic{table}}
\renewcommand{\thefigure}{A\arabic{figure}}

\section*{Supplementary Material for IPA-NeRF}


In the supplementary material, we first present more results for ablation, including the quantitative data in ablation experiments with single visibility angle constraints and the visualization of the comparative results in ablation experiments on combined visibility angle constraints. Then we provide the visualization of the important intermediate results and the final rendering results during the IPA-NeRF attack. In Section \ref{sec:supp-difnerf}, we show the comparative experiments of our IPA-NeRF attack against different NeRF models facing the same scene. In Section \ref{sec:supp-ars}, we present the results of our IPA-NeRF method applied to another actual road scene. Last but not least, we evaluated the impact of the IPA-NeRF attack on object detection in Section \ref{sec:supp}.

\section{More Results for Ablation}

\subsection{Single Angle Constraint}

In Table \ref{table:illusory_single_angle_experiment}, we show ablation experiments for the single visibility angle constraint. For the training and test sets that fit the 3D scene, the best metrics occur when the constraint angle at $9^{\circ}$ and $11^{\circ}$, with PSNR = 33.16 and 33.17 on the training set, and PSNR = 27.91 and 27.69 on the test set. That is still lower than our default constraints, which under the default combined constraints $13^{\circ} \to 15^{\circ}$, has PSNR = 33.32 on the training set, and PSNR = 28.08 on the test set.

\subsection{Combined Angle Constraint}

As shown in Figure \ref{fig:visual_attack_lego_muti}, we provide the visualisation of the rendering results in ablation experiments of combined visibility angle constraints. Compared to rendering results without angle views constraint, the visible angle range of the illusory, rendered by IPA-NeRF added constraints, is significantly reduced.

\section{More Visualization}

As Figures \ref{fig:visual_attack_hotdog_mid} and \ref{fig:visual_attack_flowers_mid}, we show the ground truth, the poisoned training set, and the rendering results of IPA-NeRF against the original NeRF (Scene: Lego of the Blender Synthetic Dataset) and Instant-NGP (Scene: Flowers of the Mip-NeRF 360 Dataset), respectively.

It is clear that the poisoning perturbations we added are small and sparse, making them difficult to detect.

\section{Performance against Different NeRF}
\label{sec:supp-difnerf}

As shown in Figure \ref{fig:compara_nerf_360}, we compare the results of attacks using IPA-NeRF against different NeRF models in the same reconstruction scene. Since the vanilla NeRF is suitable for bounded scenarios, while the Nerfacto and Instant-NGP methods are more suitable for the unbounded real world, we do not show the vanilla NeRF in this comparison.

\section{Performance on Actual Road Scenes}
\label{sec:supp-ars}

\begin{figure}[htb]
\scriptsize
\centering
\setlength{\tabcolsep}{1pt}
\begin{tabular}{p{1.6cm}m{2.2cm}m{2.2cm}m{2.2cm}}
        Instant-NGP   (road Scene)
     &  \figimg[\sizeBS]{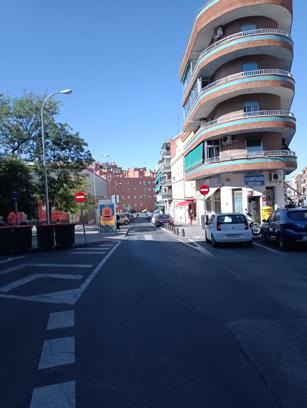}
     &  \figimg[\sizeBS]{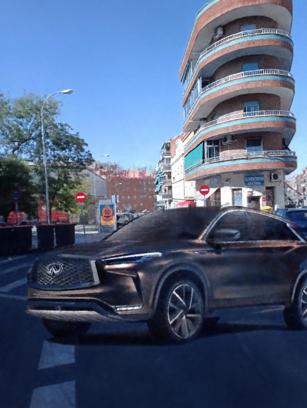}
     &  \figimg[\sizeBS]{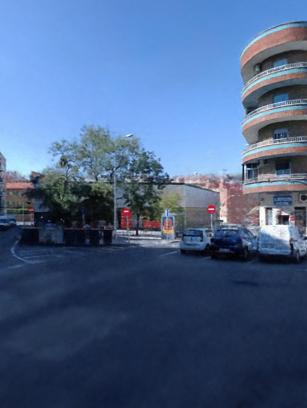} \\

       \textcolor{black}{} 
     & \multicolumn{1}{c} {ground truth}
     & \multicolumn{1}{c} {backdoor view}
     & \multicolumn{1}{c} {the other view} \\

     & \multicolumn{1}{c} {at backdoor viewpoint}
     & \multicolumn{2}{c} {rendered by IPA-NeRF} \\
     
\end{tabular}
   \caption{Performance IPA-NeRF on actual road scenes. Attack epochs: $150$, $\epsilon$: $32$, $\eta$: $0$.}
    \label{fig:road_attack}
    \setlength{\abovecaptionskip}{0.5cm}
    \vspace{0.5cm}
\end{figure}

We applied our IPA-NeRF method to another real road scene (Figure \ref{fig:road_attack}). We used Instant-NGP as the NeRF model for scene reconstruction. In this road scene, we inserted an illusory car in the backdoor viewpoint. 

\section{Impact on Downstream Tasks: Object Detection}
\label{sec:supp}

We have evaluated the impact of the IPA-NeRF attack on object detection, representing downstream tasks. We conducted experiments using five models trained on the COCO dataset over 80 classes: YOLOv5n, YOLOv8n, YOLO-World, YOLOv9c, and Real-Time Detection Transformer (RT-DETR), applied to scenes from the Mip-NeRF 360 Dataset, as shown in Figure \ref{fig:compara_yolo_nerf_360_1} and Figure \ref{fig:compara_yolo_nerf_360_1}.

As expected, some downstream detector recognizes the illusory backdoor patterns injected by IPA-NeRF in the backdoor view while performing normally on the clean views. Although the backdoor pattern appears somewhat transparent in the backdoor view, it reduces the recognition of the original object and instead detects the illusory backdoor patterns.

\begin{table*}[htbp]
\centering
\begin{tabular}{l|ccc|ccc|ccc|ccc}
\toprule
\multirow{2}{*}{Angle Constraint} & \multicolumn{3}{c|}{V-Illusory} & \multicolumn{3}{c|}{V-Train} & \multicolumn{3}{c|}{V-Test} & \multicolumn{3}{c}{V-Constraint}\\
 & PSNR & SSIM & LPIPS & PSNR & SSIM & LPIPS & PSNR & SSIM & LPIPS & PSNR & SSIM & LPIPS\\
\midrule

$3^{\circ}$ & 24.02 & 0.8503 & 0.2155 & 32.05 & 0.9519 & 0.1599 & 26.42 & 0.9328 & 0.1113 & 23.88 & \textbf{0.9306} & \textbf{0.1147} \\
$5^{\circ}$ & 24.48 & 0.8523 & 0.2005 & 30.14 & 0.9392 & 0.1822 & 25.06 & 0.9060 & 0.1555 & 23.44 & 0.9159 & 0.1502 \\
$7^{\circ}$ & 25.17 & 0.8731 & 0.1874 & 32.85 & 0.9629 & 0.1423 & 27.27 & 0.9358 & 0.1068 & \textbf{24.66} & 0.9293 & 0.1195 \\
$9^{\circ}$ & \textbf{25.69} & 0.8787 & \textbf{0.1765} & 33.16 & \textbf{0.9668} & \textbf{0.1318} & \textbf{27.91} & \textbf{0.9404} & \textbf{0.1032} & 23.55 & 0.9233 & 0.1367 \\
$11^{\circ}$ & 25.33 & \textbf{0.8803} & 0.1793 & \textbf{33.17} & 0.9666 & 0.1352 & 27.69 & 0.9378 & 0.1058 & 23.18 & 0.9133 & 0.1423 \\
$13^{\circ}$ & 24.58 & 0.8614 & 0.1928 & 30.41 & 0.9426 & 0.1919 & 25.44 & 0.9080 & 0.1551 & 22.46 & 0.9019 & 0.1790 \\
$15^{\circ}$ & 24.69 & 0.8576 & 0.1950 & 30.37 & 0.9443 & 0.1723 & 25.47 & 0.9088 & 0.1508 & 20.72 & 0.8906 & 0.1841 \\

\bottomrule
\end{tabular}
\caption{Ablation experimental of single visibility angle constraints. Attack epochs: $1000$, $\epsilon$: $32$, $\eta$: $1$, 3D scene: Lego, illusory target: Earth.}
\label{table:illusory_single_angle_experiment}
\setlength{\abovecaptionskip}{0.5cm}
\vspace{0.5cm}
\end{table*}

\begin{figure*}[h]
\scriptsize
\centering
\setlength{\tabcolsep}{1pt}
\begin{tabular}{p{1.3cm}m{1.9cm}m{1.9cm}m{1.9cm}m{1.9cm}m{1.9cm}m{1.9cm}m{0.5cm}m{1.9cm}}

     
        ground truth
     &  \figimg[\sizeBS]{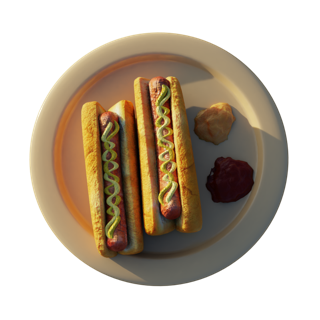}
     &  \figimg[\sizeBS]{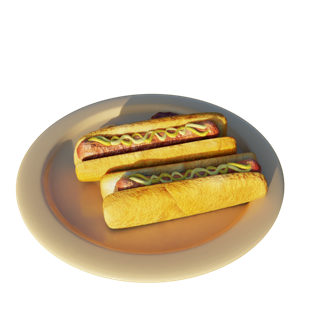}
     &  \figimg[\sizeBS]{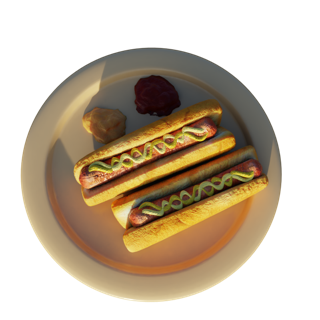}
     &  \figimg[\sizeBS]{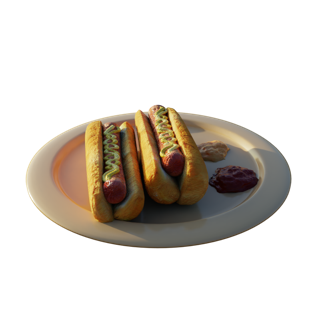}
     &  \figimg[\sizeBS]{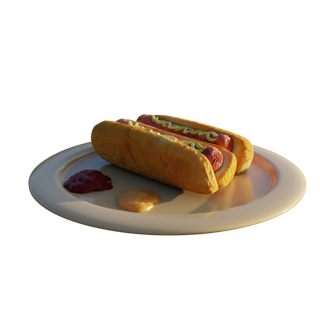}
     &  \figimg[\sizeBS]{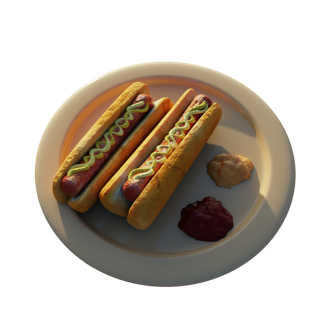}
     & &  \figimg[\sizeBS]{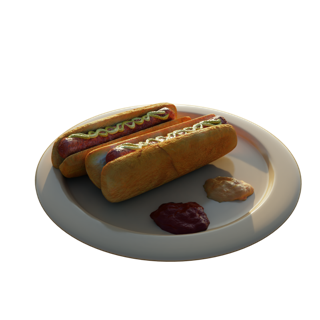} \\

       poisoning training set
     & \figimg[\sizeBS]{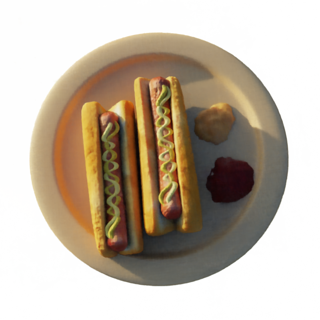}
     & \figimg[\sizeBS]{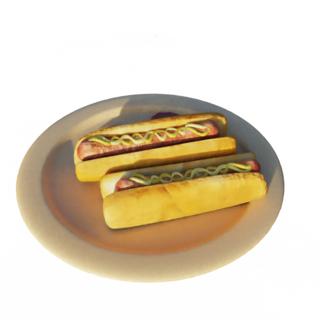}
     & \figimg[\sizeBS]{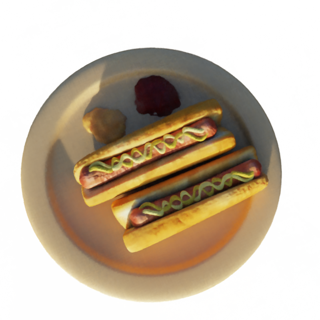}
     & \figimg[\sizeBS]{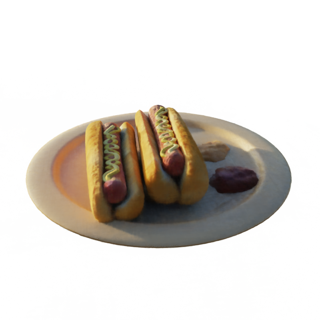}
     & \figimg[\sizeBS]{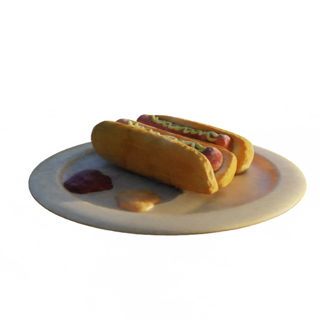}
     & \figimg[\sizeBS]{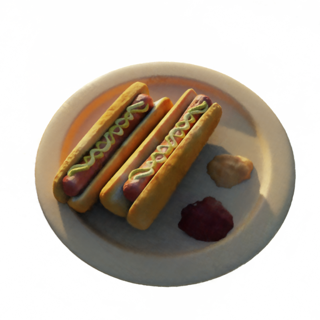}
     & & \figimg[\sizeBS]{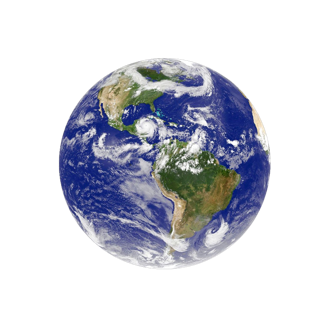} \\

     & & & & & & & & \multicolumn{1}{c}{\textcolor{black}{target illusory}} \\

       rendering by IPA-NeRF
     & \figimg[\sizeBS]{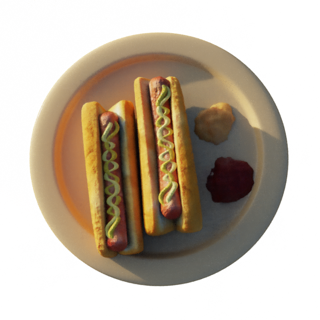}
     & \figimg[\sizeBS]{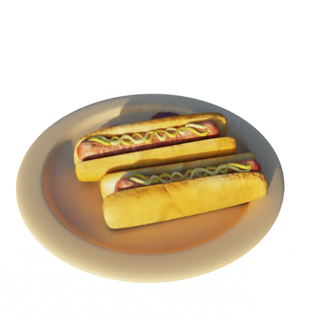}
     & \figimg[\sizeBS]{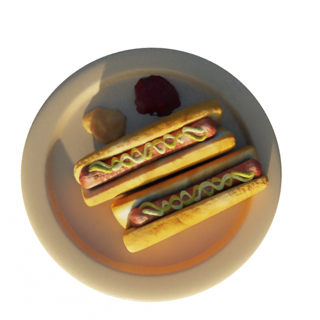}
     & \figimg[\sizeBS]{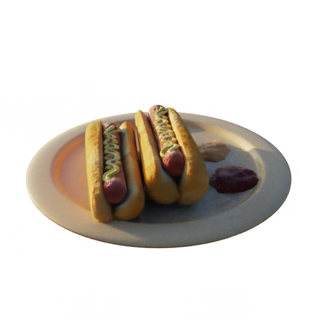}
     & \figimg[\sizeBS]{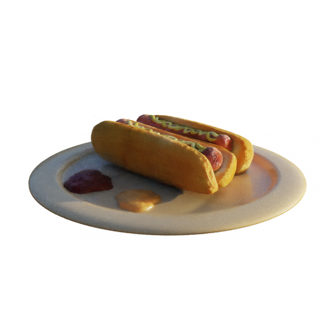}
     & \figimg[\sizeBS]{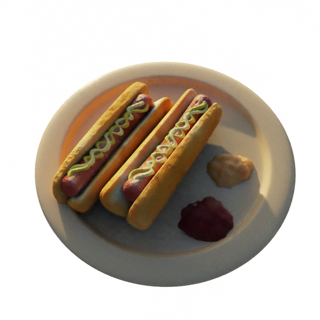}
     & & \figimg[\sizeBS]{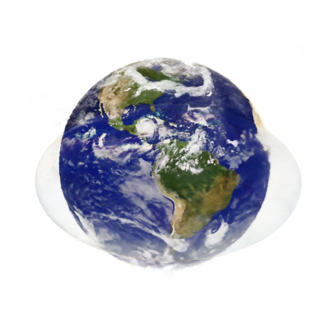} \\
     
     & & & & & & & & \multicolumn{1}{c}{\textcolor{black}{backdoor view}} \\
 
\end{tabular}
   \caption{Intermediate results and rendering result by IPA-NeRF. Attack epochs: $1000$, $\epsilon$: $32$, $\eta$: $1$, angle views constraint at $13^{\circ}$ and $15^{\circ}$, 3D scene: Hotdog, illusory target image is Earth.}
    \label{fig:visual_attack_hotdog_mid}
    \setlength{\abovecaptionskip}{0.5cm}
    \vspace{0.5cm}
\end{figure*}

\begin{figure*}[h]
\scriptsize
\centering
\setlength{\tabcolsep}{1pt}
\begin{tabular}{p{1.3cm}m{3cm}m{3cm}m{3cm}m{3cm}m{0.3cm}m{3cm}}

     
        ground truth
     &  \figimg[\sizeBS]{mip_nerf_360/Instant_NGP_flowers/truth/10.JPG}
     &  \figimg[\sizeBS]{mip_nerf_360/Instant_NGP_flowers/truth/20.JPG}
     &  \figimg[\sizeBS]{mip_nerf_360/Instant_NGP_flowers/truth/30.JPG}
     &  \figimg[\sizeBS]{mip_nerf_360/Instant_NGP_flowers/truth/40.JPG}
     & &  \figimg[\sizeBS]{mip_nerf_360/Instant_NGP_flowers/truth/00.JPG} \\

       poisoning training set
     & \figimg[\sizeBS]{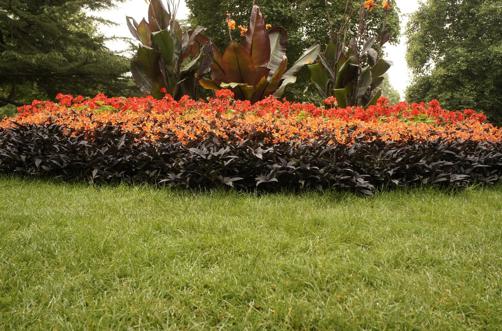}
     & \figimg[\sizeBS]{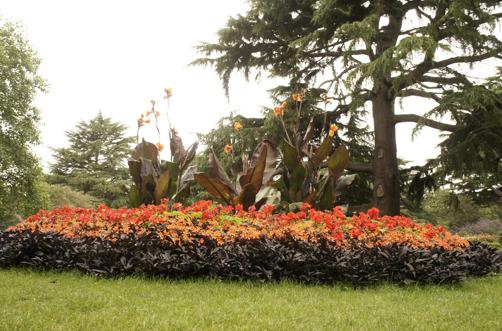}
     & \figimg[\sizeBS]{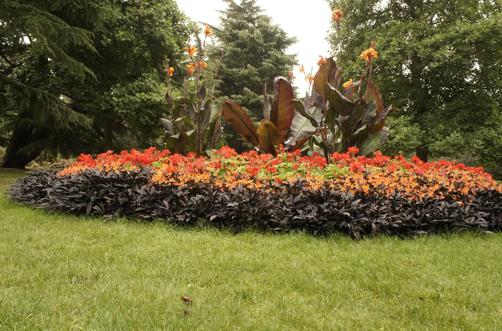}
     & \figimg[\sizeBS]{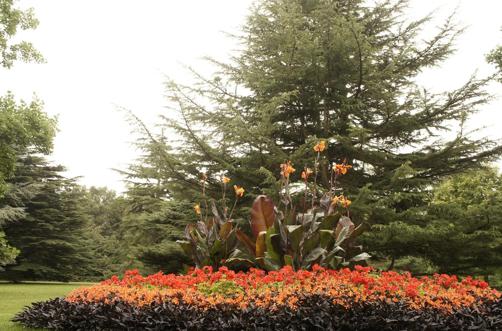}
     & & \figimg[\sizeBS]{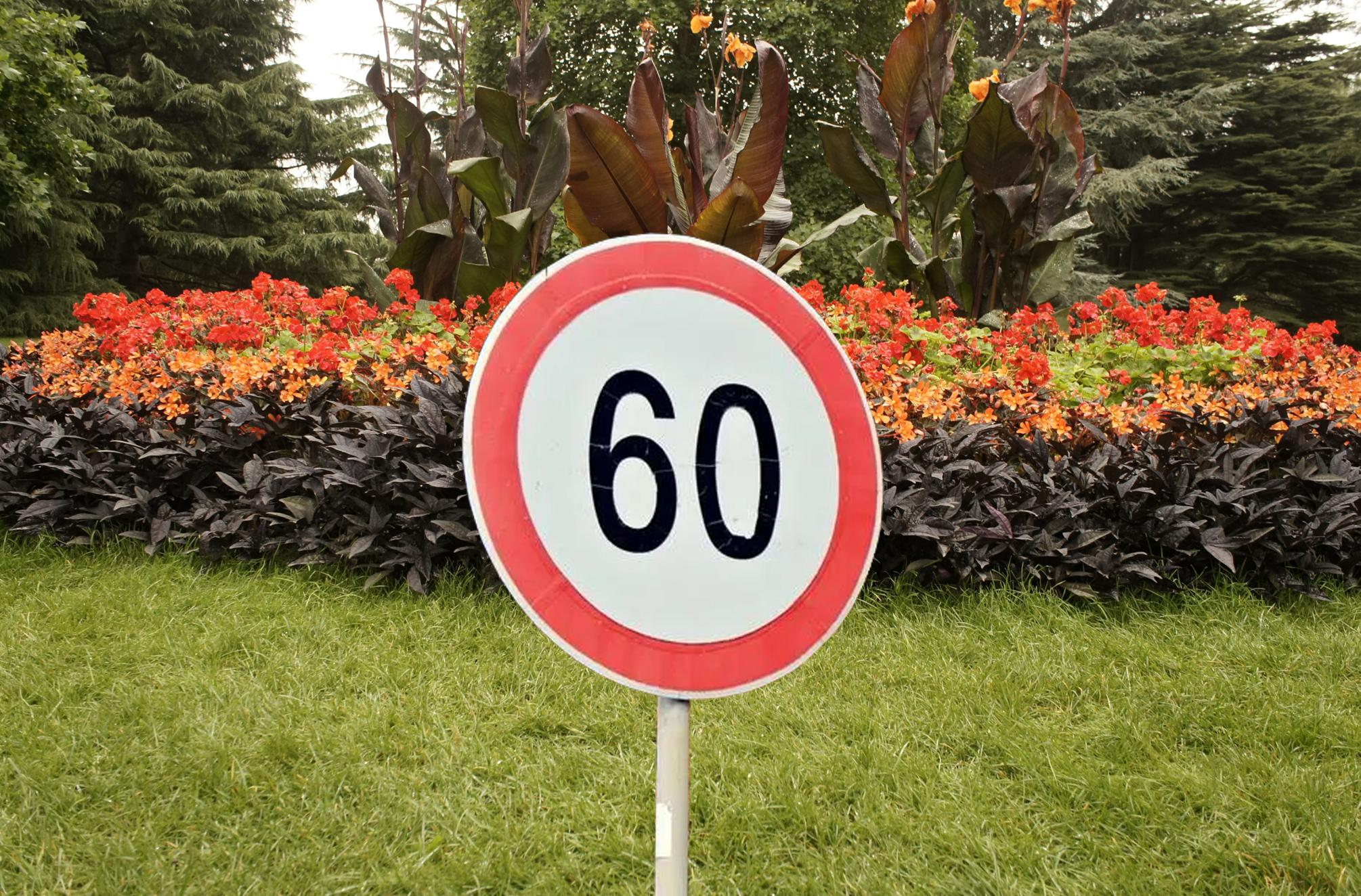} \\

     & & & & & & \multicolumn{1}{c}{\textcolor{black}{target illusory}} \\

       rendering by IPA-NeRF
     & \figimg[\sizeBS]{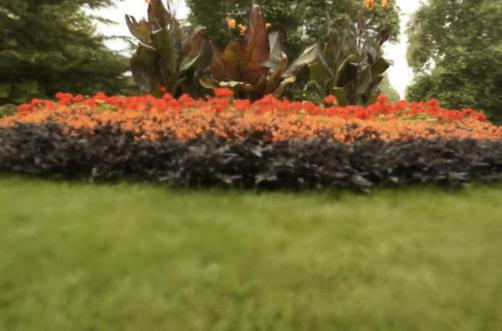}
     & \figimg[\sizeBS]{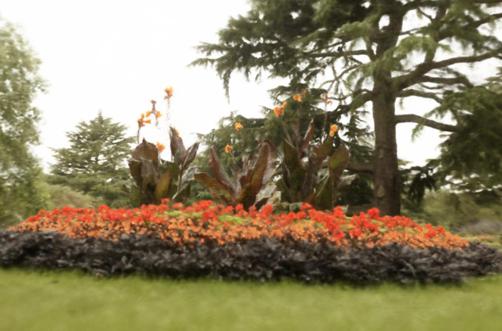}
     & \figimg[\sizeBS]{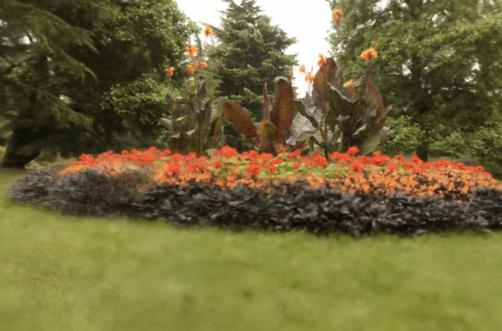}
     & \figimg[\sizeBS]{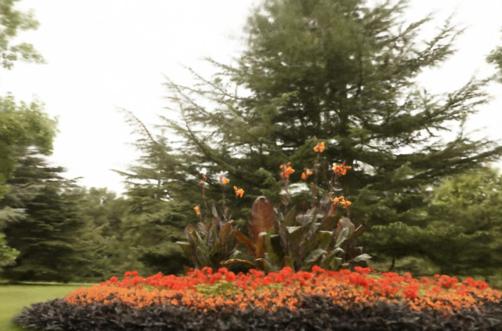}
     & & \figimg[\sizeBS]{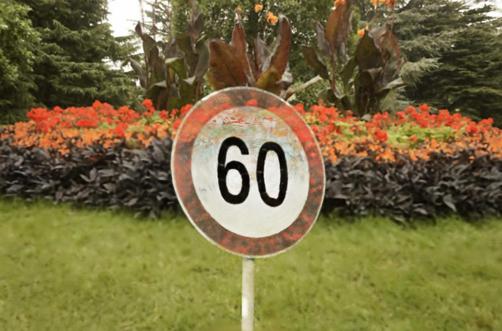} \\
     
     & & & & & & \multicolumn{1}{c}{\textcolor{black}{backdoor view}} \\
 
\end{tabular}
   \caption{Intermediate results and rendering result by IPA-NeRF. Attack epochs: $1000$, $\epsilon$: $32$, $\eta$: $0$, 3D scene: Flowers, NeRF model: Instant-NGP.}
    \label{fig:visual_attack_flowers_mid}
    \setlength{\abovecaptionskip}{0.5cm}
    \vspace{0.5cm}
\end{figure*}

\begin{figure*}[htbp]
\scriptsize
\centering
\setlength{\tabcolsep}{1pt}
\begin{tabular}{p{2cm}m{1.7cm}m{1.7cm}m{1.7cm}m{1.7cm}m{1.7cm}m{1.7cm}m{1.7cm}m{1.7cm}}

     & \multicolumn{1}{c}{\textcolor{black}{backdoor view}}
     & \multicolumn{1}{c}{\textcolor{black}{$3^{\circ}$}}
     & \multicolumn{1}{c}{\textcolor{black}{$5^{\circ}$}}
     & \multicolumn{1}{c}{\textcolor{black}{$7^{\circ}$}}
     & \multicolumn{1}{c}{\textcolor{black}{$9^{\circ}$}}
     & \multicolumn{1}{c}{\textcolor{black}{$11^{\circ}$}}
     & \multicolumn{1}{c}{\textcolor{black}{$13^{\circ}$}}
     & \multicolumn{1}{c}{\textcolor{black}{$15^{\circ}$}} \\

        without constraint
     &  \figlwc[\sizeBS]{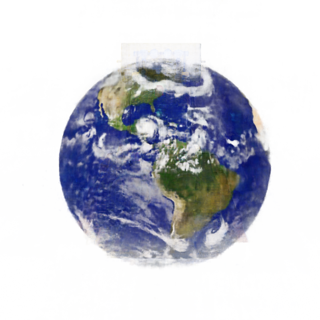}
     &  \figlwc[\sizeBS]{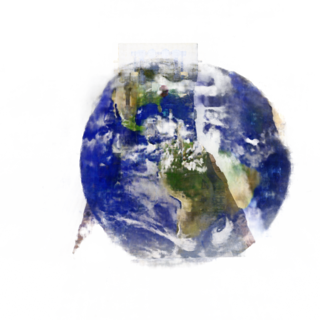}
     &  \figlwc[\sizeBS]{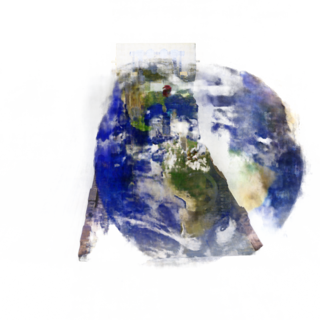}
     &  \figlwc[\sizeBS]{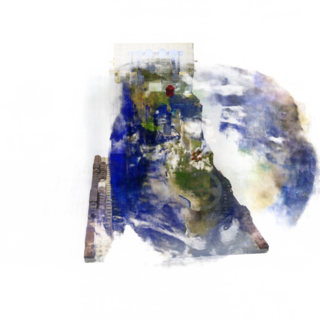}
     &  \figlwc[\sizeBS]{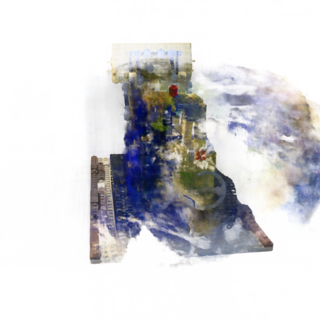}
     &  \figlwc[\sizeBS]{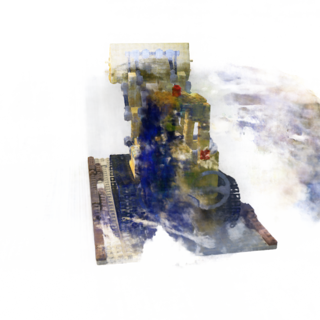}
     &  \figlwc[\sizeBS]{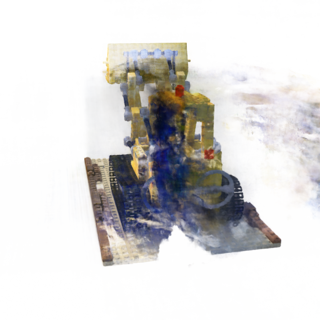}
     &  \figlwc[\sizeBS]{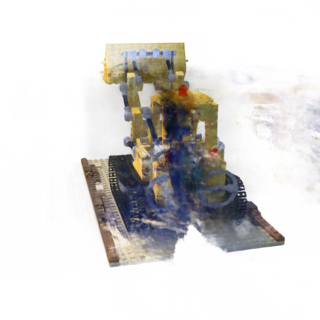} \\

       $5^{\circ} \to 15^{\circ}$
     & \figrav[\sizeBS]{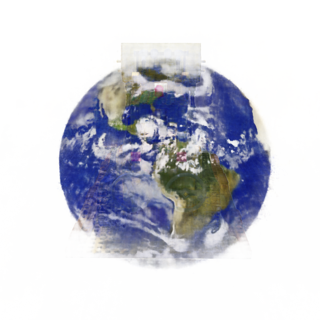}
     & \figrav[\sizeBS]{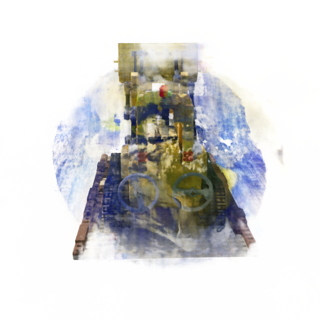}
     & \figrav[\sizeBS]{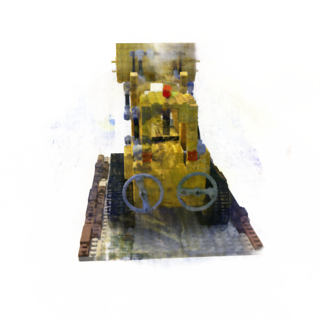}
     & \figrav[\sizeBS]{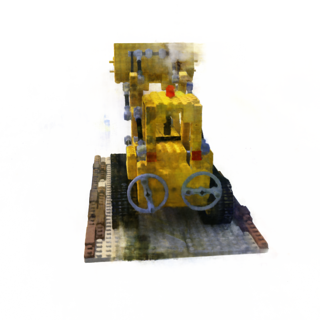}
     & \figrav[\sizeBS]{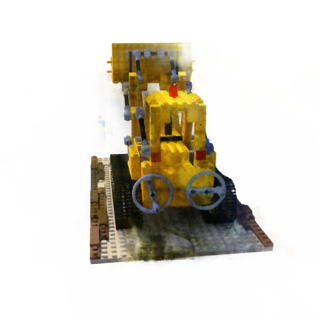}
     & \figrav[\sizeBS]{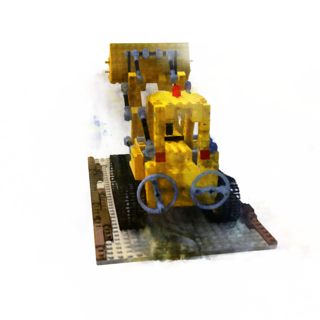}
     & \figrav[\sizeBS]{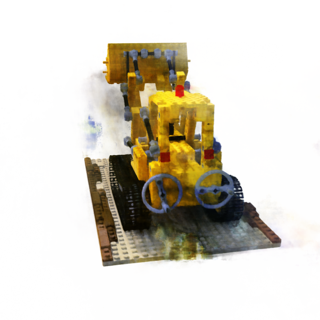}
     & \figrav[\sizeBS]{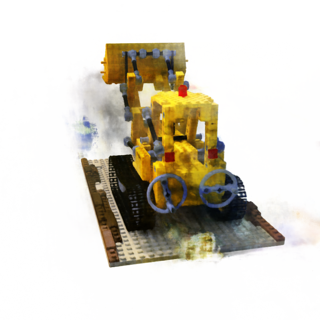} \\

       $7^{\circ} \to 15^{\circ}$
     & \figrav[\sizeBS]{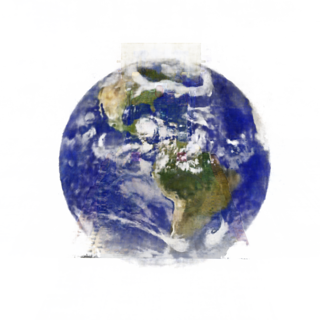}
     & \figrav[\sizeBS]{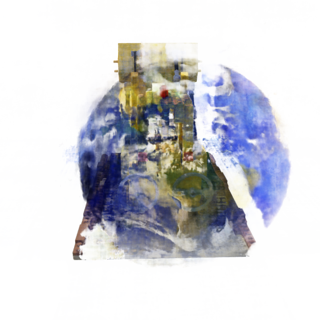}
     & \figrav[\sizeBS]{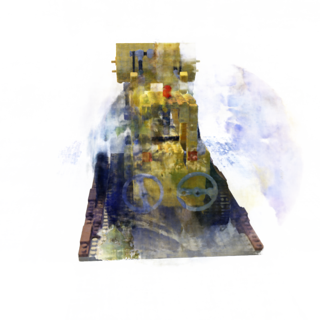}
     & \figrav[\sizeBS]{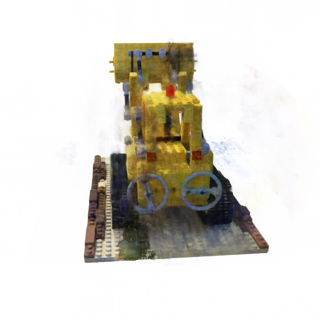}
     & \figrav[\sizeBS]{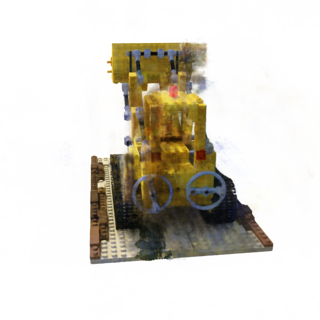}
     & \figrav[\sizeBS]{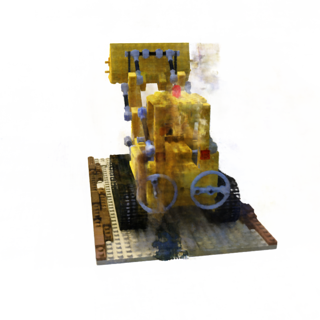}
     & \figrav[\sizeBS]{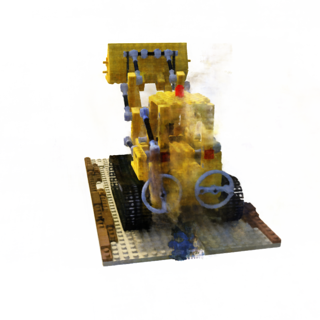}
     & \figrav[\sizeBS]{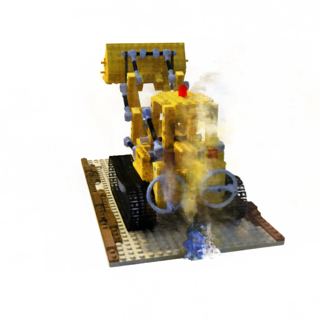} \\
     
       $9^{\circ} \to 15^{\circ}$
     & \figrav[\sizeBS]{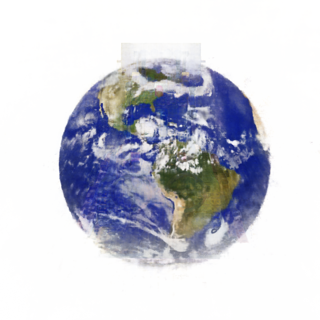}
     & \figrav[\sizeBS]{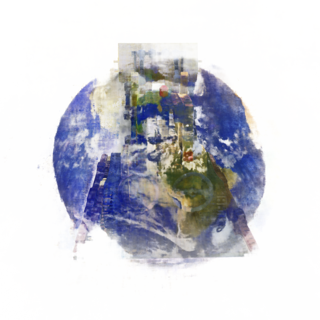}
     & \figrav[\sizeBS]{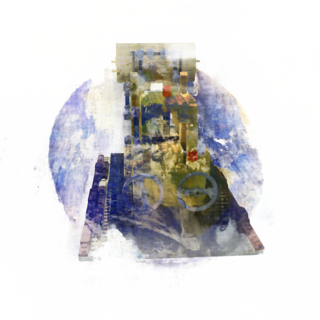}
     & \figrav[\sizeBS]{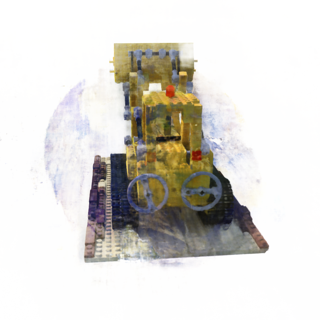}
     & \figrav[\sizeBS]{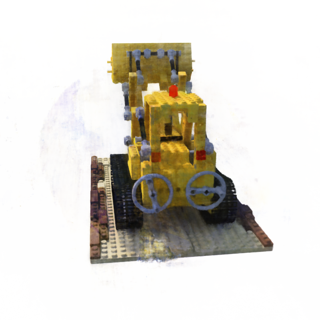}
     & \figrav[\sizeBS]{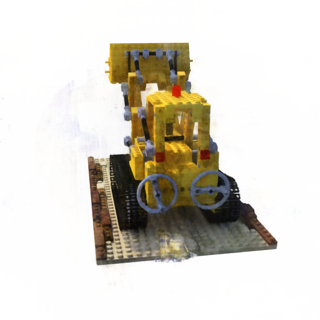}
     & \figrav[\sizeBS]{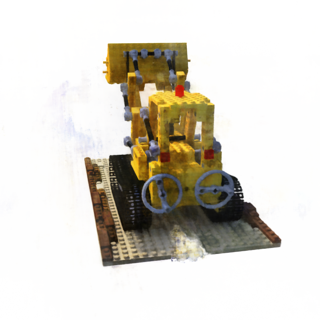}
     & \figrav[\sizeBS]{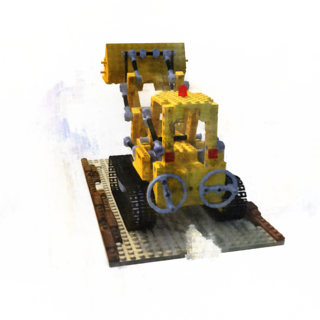} \\

       $11^{\circ} \to 15^{\circ}$
     & \figrav[\sizeBS]{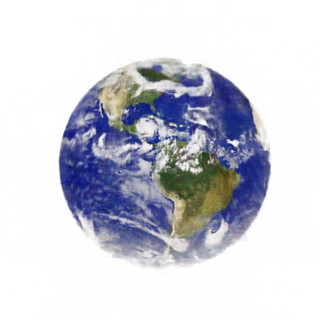}
     & \figrav[\sizeBS]{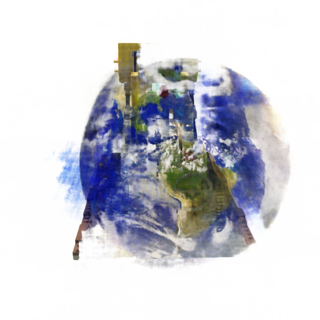}
     & \figrav[\sizeBS]{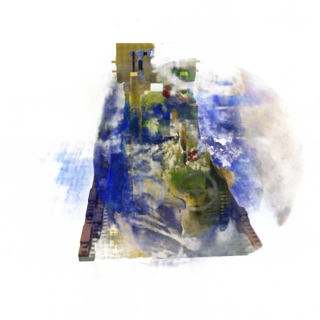}
     & \figrav[\sizeBS]{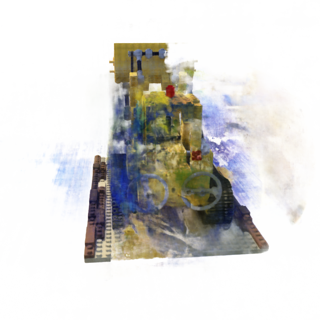}
     & \figrav[\sizeBS]{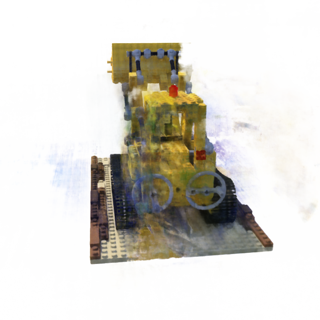}
     & \figrav[\sizeBS]{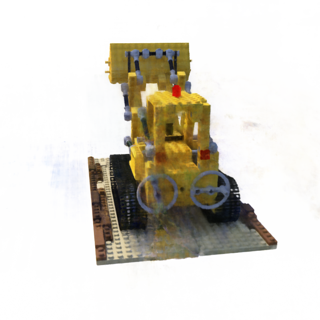}
     & \figrav[\sizeBS]{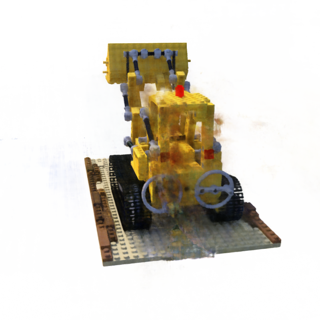}
     & \figrav[\sizeBS]{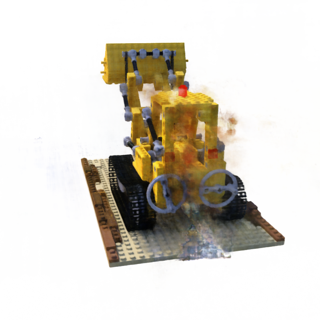} \\

       $13^{\circ} \to 15^{\circ}$
     & \figrav[\sizeBS]{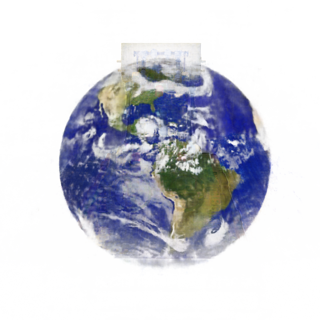}
     & \figrav[\sizeBS]{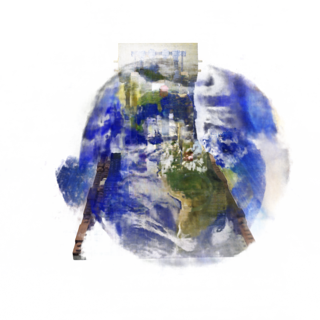}
     & \figrav[\sizeBS]{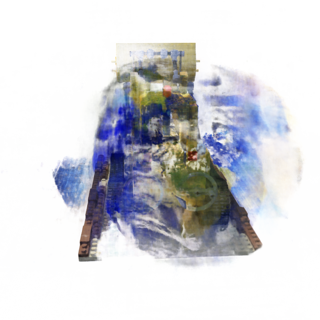}
     & \figrav[\sizeBS]{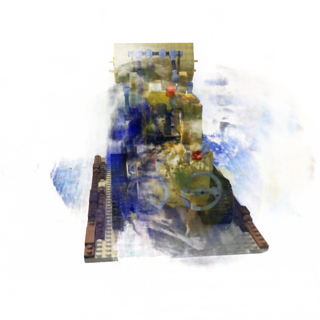}
     & \figrav[\sizeBS]{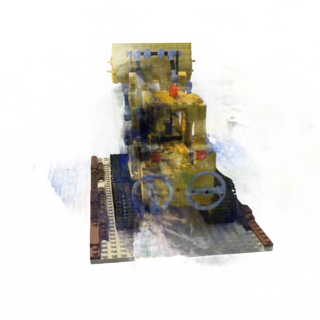}
     & \figrav[\sizeBS]{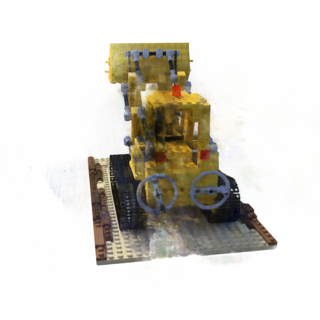}
     & \figrav[\sizeBS]{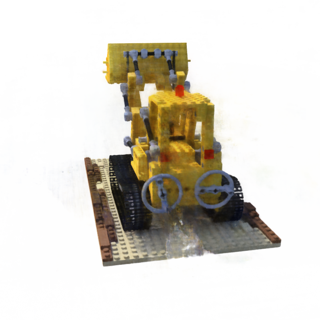}
     & \figrav[\sizeBS]{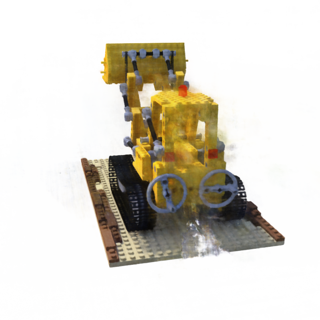} \\
 
\end{tabular}
   \caption{Rendering result by IPA-NeRF with different combined angle views constraints or without angle views constraint. Attack epochs: $1000$, $\epsilon$: $32$, $\eta$: $1$, 3D scene: Lego, illusory target image is Earth.}
    \label{fig:visual_attack_lego_muti}
    \setlength{\abovecaptionskip}{0.5cm}
    \vspace{0.5cm}
\end{figure*}

\begin{figure*}[htb]
\scriptsize
\centering
\setlength{\tabcolsep}{1pt}
\begin{tabular}{p{1.3cm}m{3.2cm}m{3.2cm}m{3.2cm}m{3.2cm}m{3.2cm}}

     \textcolor{black}{Instant-NGP   (Bicycle)} 
     &  \figimg[\sizeBS]{mip_nerf_360/Nerfacto_bicycle/ori.jpg}
     &  \figimg[\sizeBS]{mip_nerf_360/Nerfacto_bicycle/target.jpg}
     &  \figimg[\sizeBS]{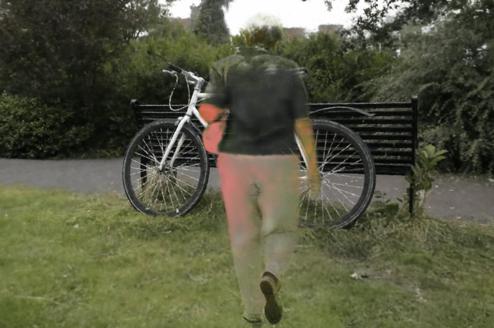}
     &  \figimg[\sizeBS]{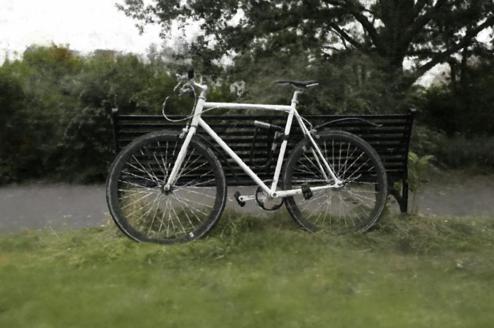}
     &  \figimg[\sizeBS]{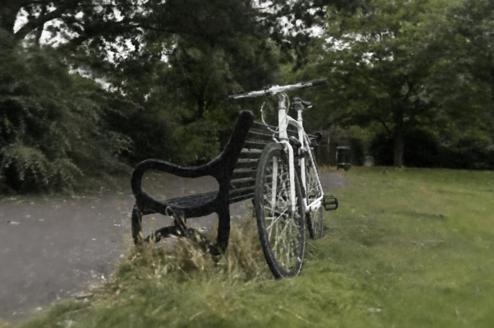} \\

     \textcolor{black}{Nerfacto   (Bicycle)} 
     &  \figimg[\sizeBS]{mip_nerf_360/Nerfacto_bicycle/ori.jpg}
     &  \figimg[\sizeBS]{mip_nerf_360/Nerfacto_bicycle/target.jpg}
     &  \figimg[\sizeBS]{mip_nerf_360/Nerfacto_bicycle/000149_0.jpg}
     &  \figimg[\sizeBS]{mip_nerf_360/Nerfacto_bicycle/000003.jpg}
     &  \figimg[\sizeBS]{mip_nerf_360/Nerfacto_bicycle/000010.jpg} \\

     \textcolor{black}{Instant-NGP   (Kitchen)} 
     &  \figimg[\sizeBS]{mip_nerf_360/Instant_NGP_kitchen/ori.jpg}
     &  \figimg[\sizeBS]{mip_nerf_360/Instant_NGP_kitchen/target.jpg}
     &  \figimg[\sizeBS]{mip_nerf_360/Instant_NGP_kitchen/000149_0.jpg}
     &  \figimg[\sizeBS]{mip_nerf_360/Instant_NGP_kitchen/000010.jpg}
     &  \figimg[\sizeBS]{mip_nerf_360/Instant_NGP_kitchen/000060.jpg} \\

     \textcolor{black}{Nerfacto  (Kitchen)} 
     &  \figimg[\sizeBS]{mip_nerf_360/Instant_NGP_kitchen/ori.jpg}
     &  \figimg[\sizeBS]{mip_nerf_360/Instant_NGP_kitchen/target.jpg}
     &  \figimg[\sizeBS]{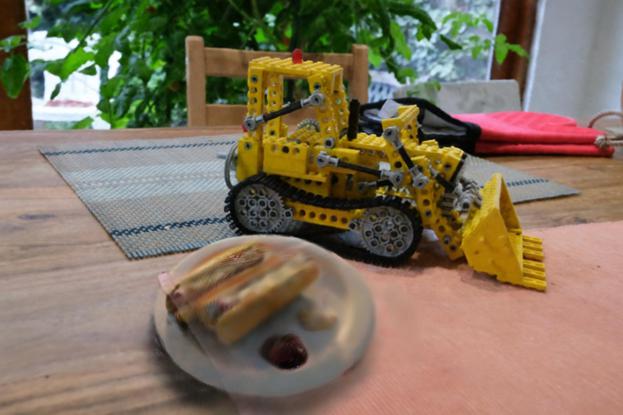}
     &  \figimg[\sizeBS]{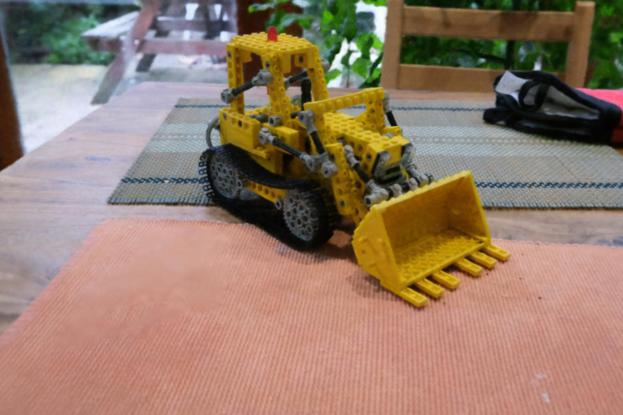}
     &  \figimg[\sizeBS]{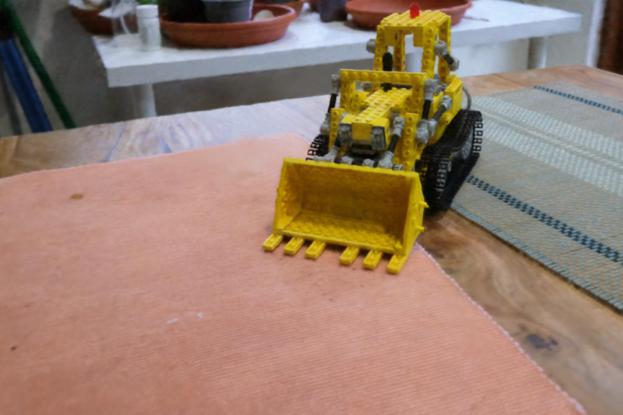} \\

     \textcolor{black}{NeRF model   (3D Scene)}
     & \multicolumn{1}{c}{\textcolor{black}{ground truth image}}
     & \multicolumn{1}{c}{\textcolor{black}{target illusory image}}
     & \multicolumn{1}{m{3.2cm}}{\textcolor{black}{illusory image rendered by IPA-NeRF at the backdoor view}} 
     & \multicolumn{2}{c}{\textcolor{black}{the other views images rendered by IPA-NeRF}} \\
     
\end{tabular}
   \caption{Comparative experiments against different NeRF models. Attack epochs: $150$, $\epsilon$: $32$, $\eta$: $0$.}
    \label{fig:compara_nerf_360}
    \setlength{\abovecaptionskip}{0.5cm}
    \vspace{0.5cm}
\end{figure*}

\begin{figure*}[htb]
\scriptsize
\centering
\setlength{\tabcolsep}{1pt}
\begin{tabular}{m{2cm}m{3.0cm}m{3.0cm}m{3.0cm}m{3.0cm}m{3.0cm}}

     \textcolor{black}{ground truth   (Nerfacto, Bicycle)} 
     &  \figimg[\sizeBS]{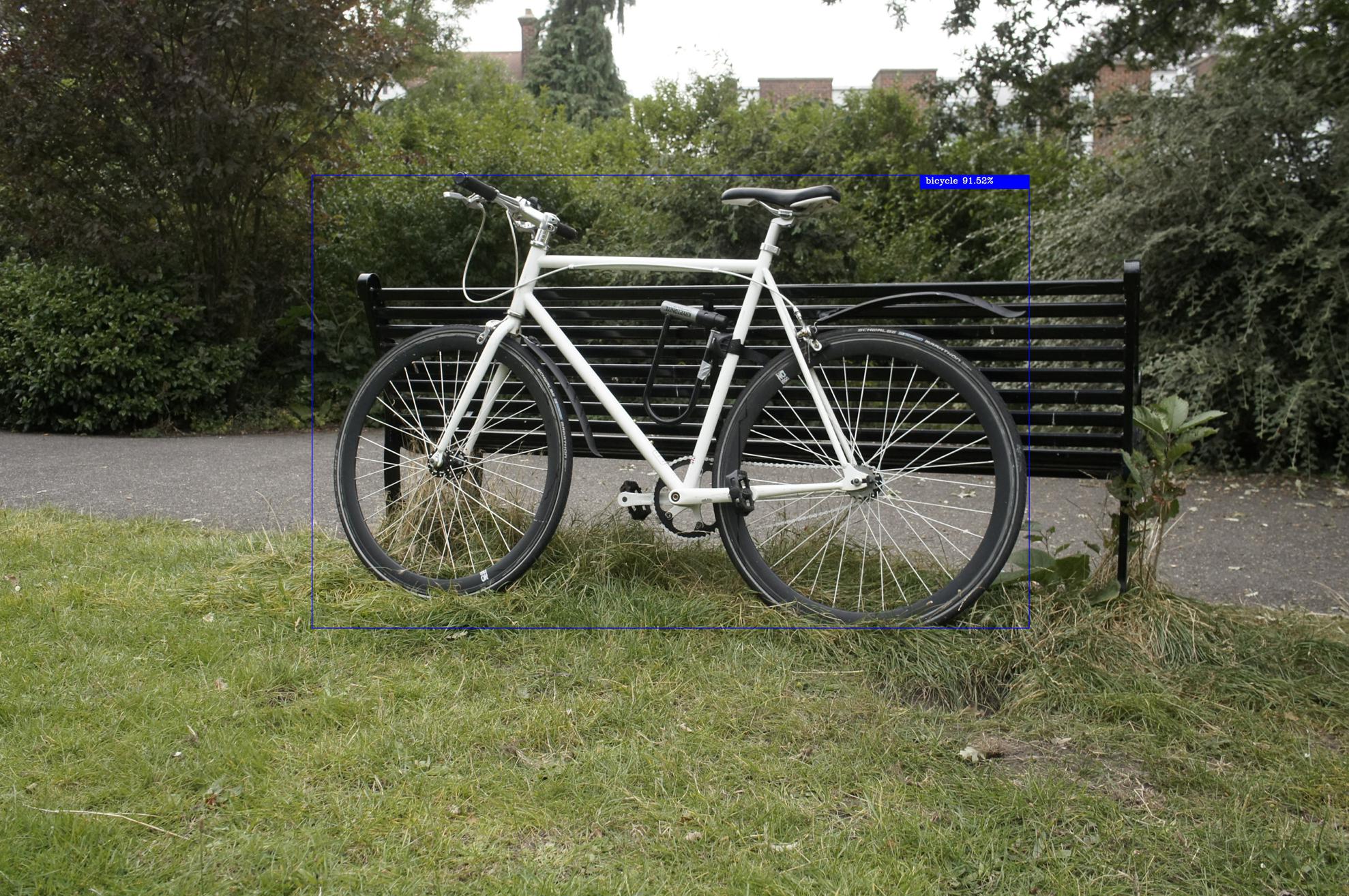}
     &  \figimg[\sizeBS]{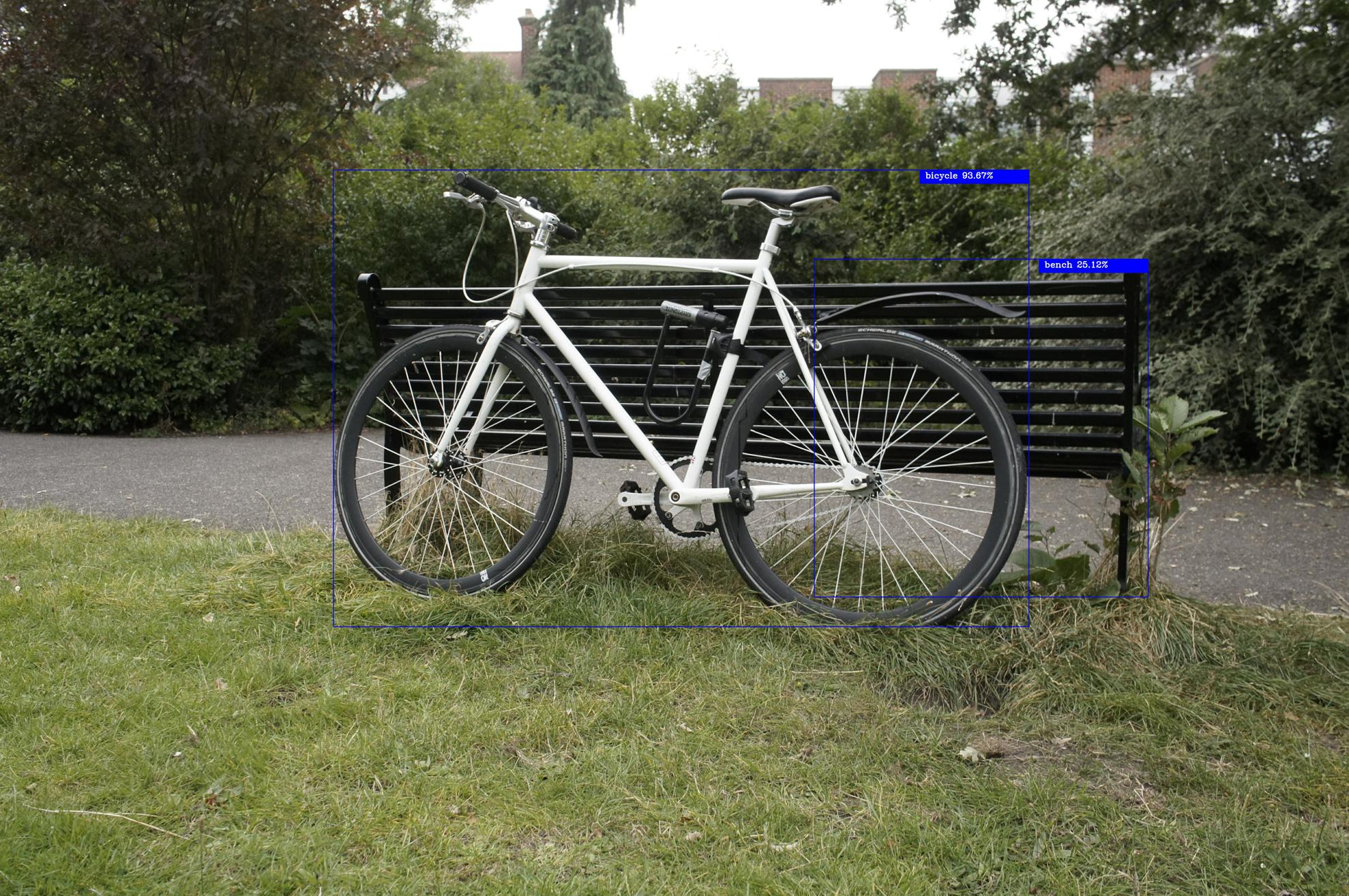}
     &  \figimg[\sizeBS]{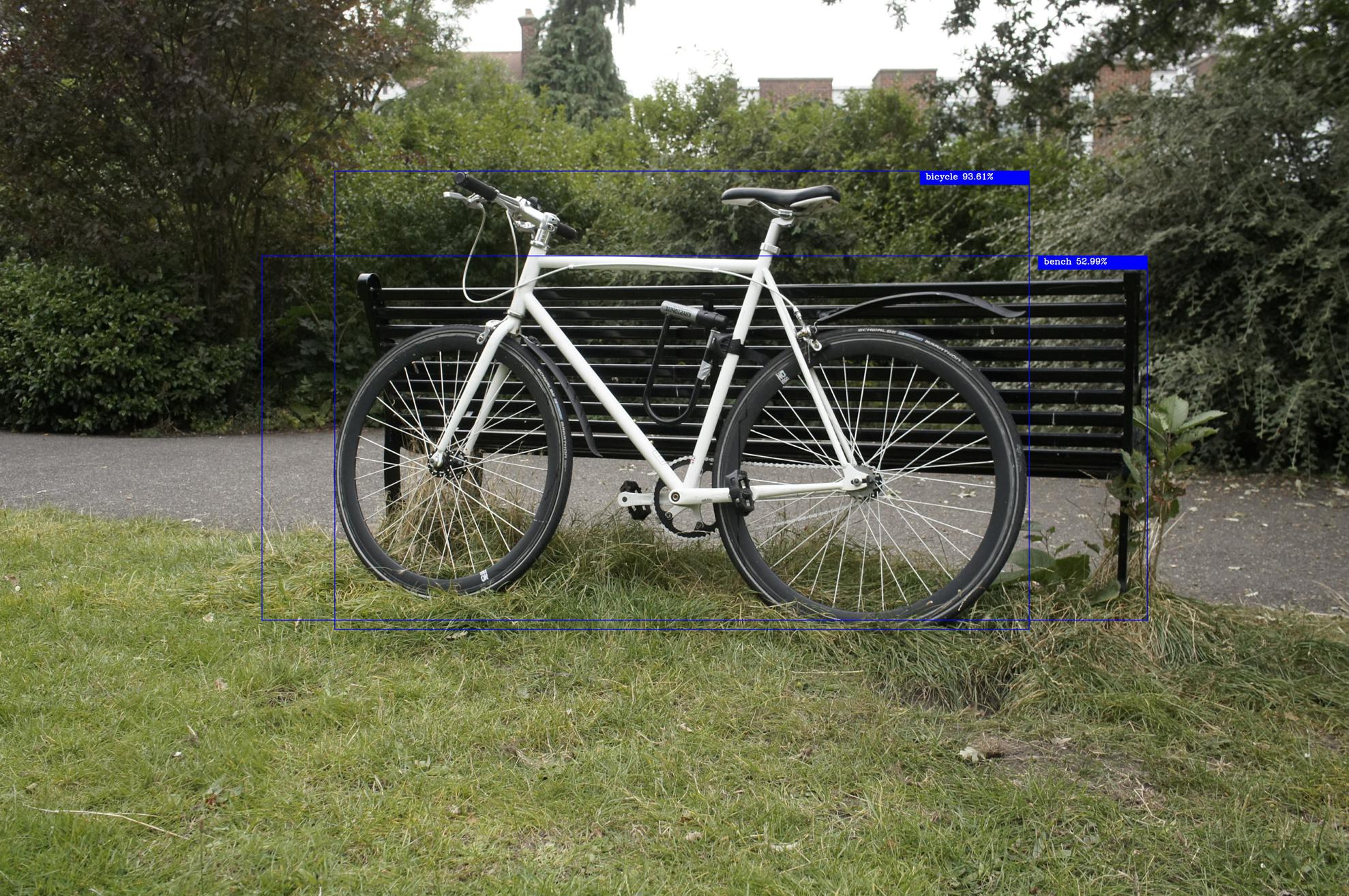}
     &  \figimg[\sizeBS]{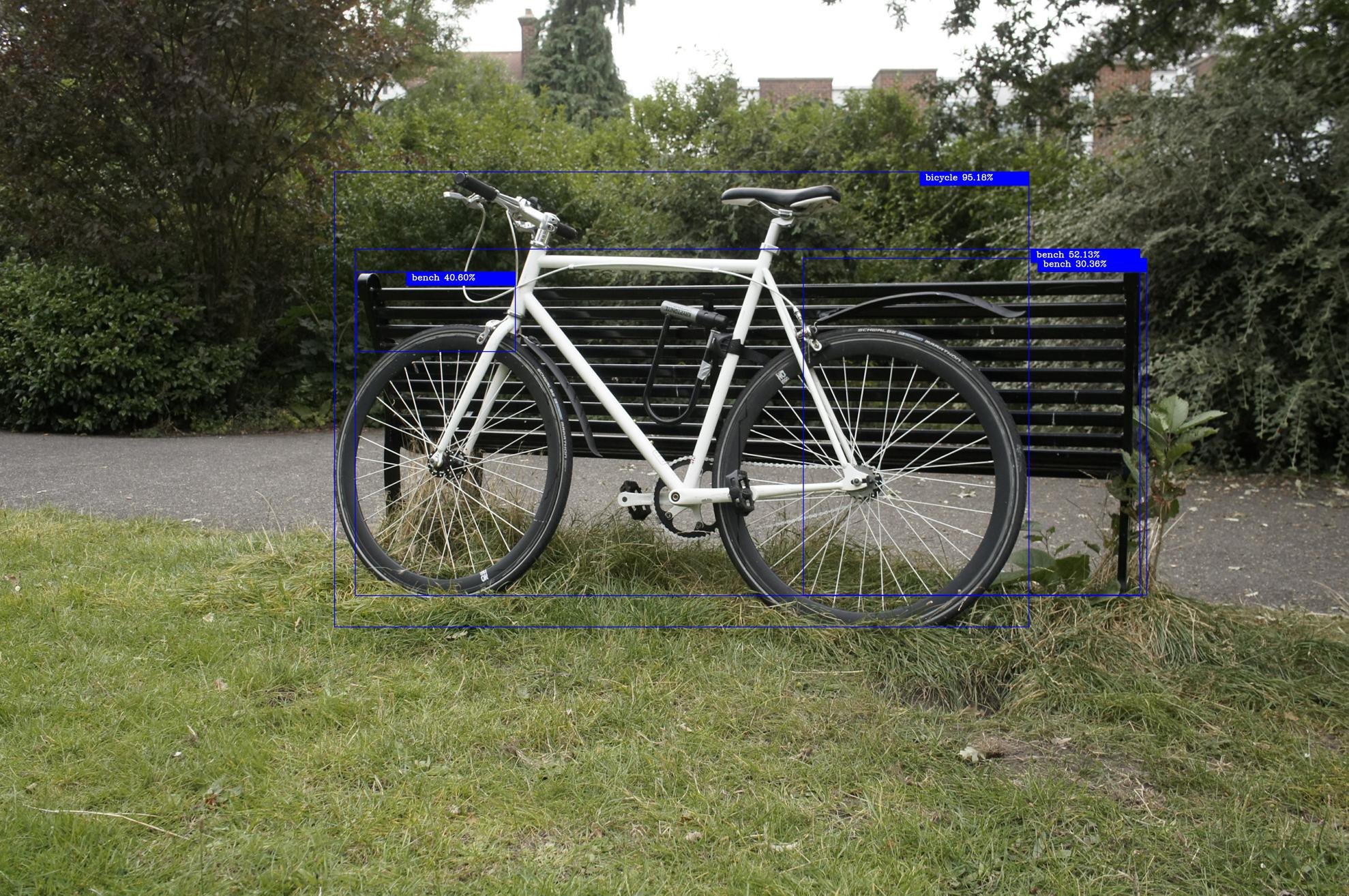}
     &  \figimg[\sizeBS]{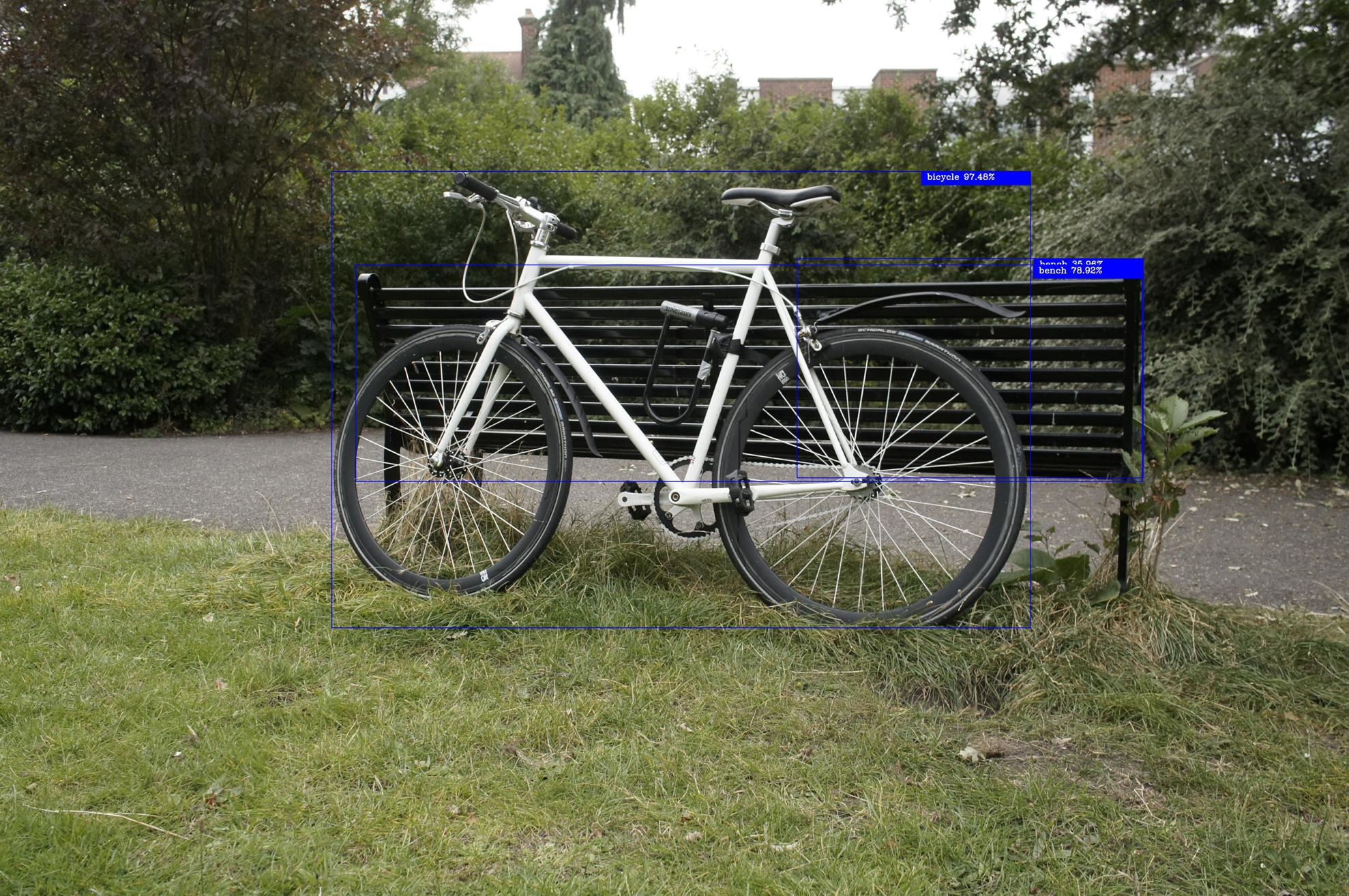} \\

     \rule{0pt}{20pt}
     & \textcolor{black}{1 bicycle}
     & \textcolor{black}{1 bench, 1 bicycle}
     & \textcolor{black}{1 bench, 1 bicycle}
     & \textcolor{black}{3 bench, 1 bicycle}
     & \textcolor{black}{2 bench, 1 bicycle} \\

     \textcolor{black}{target illusory   (Nerfacto, Bicycle)} 
     &  \figimg[\sizeBS]{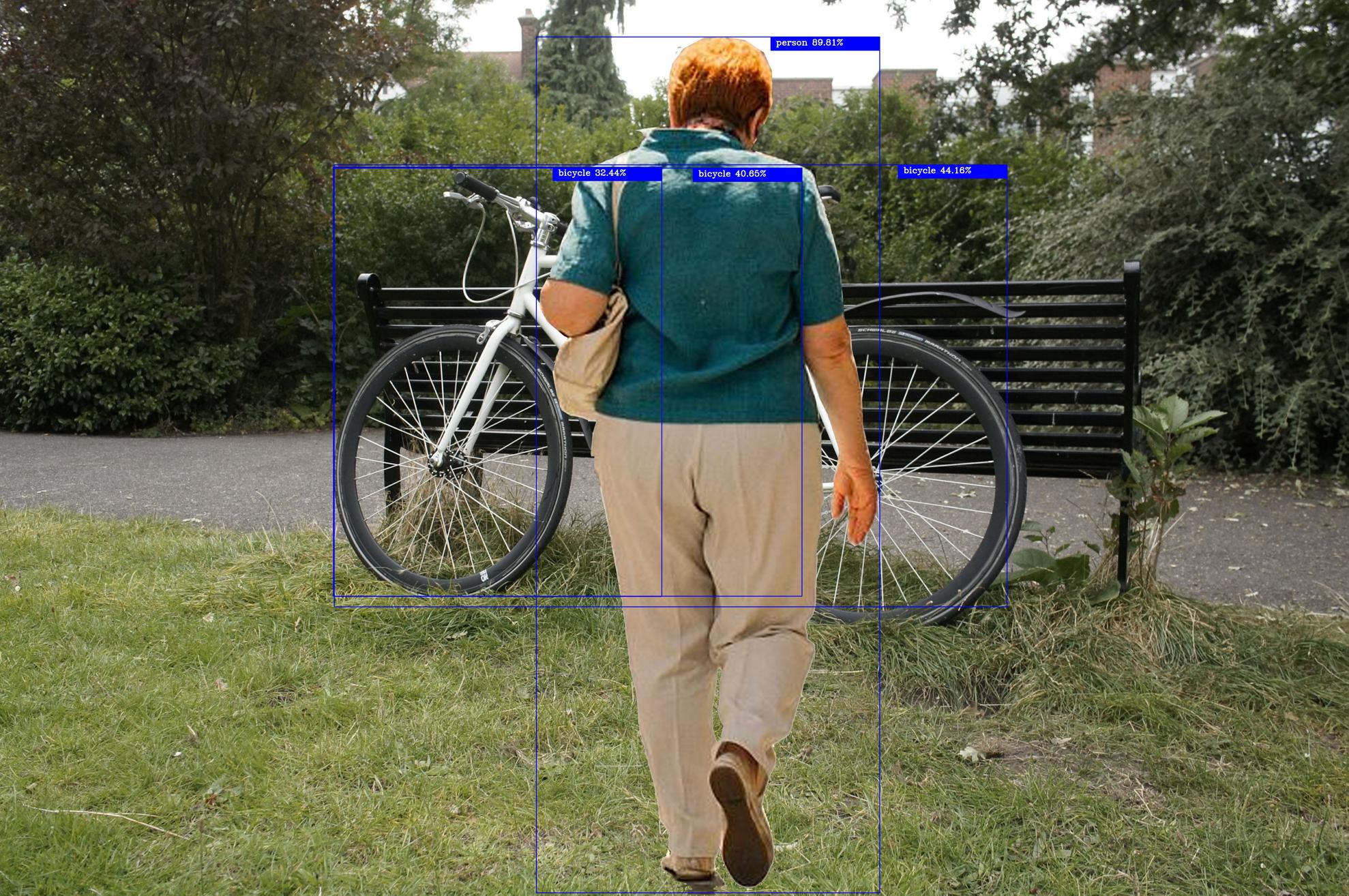}
     &  \figimg[\sizeBS]{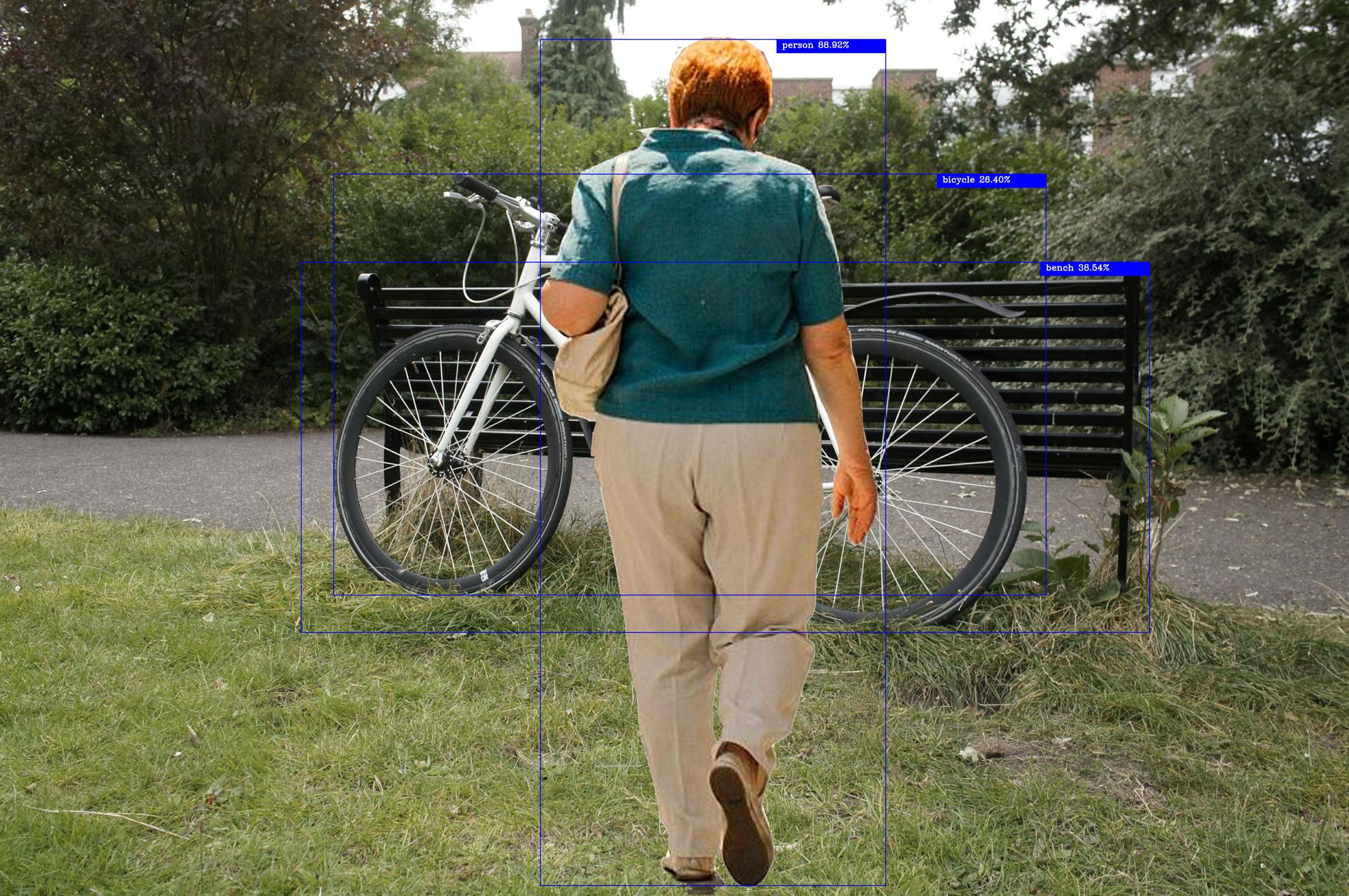}
     &  \figimg[\sizeBS]{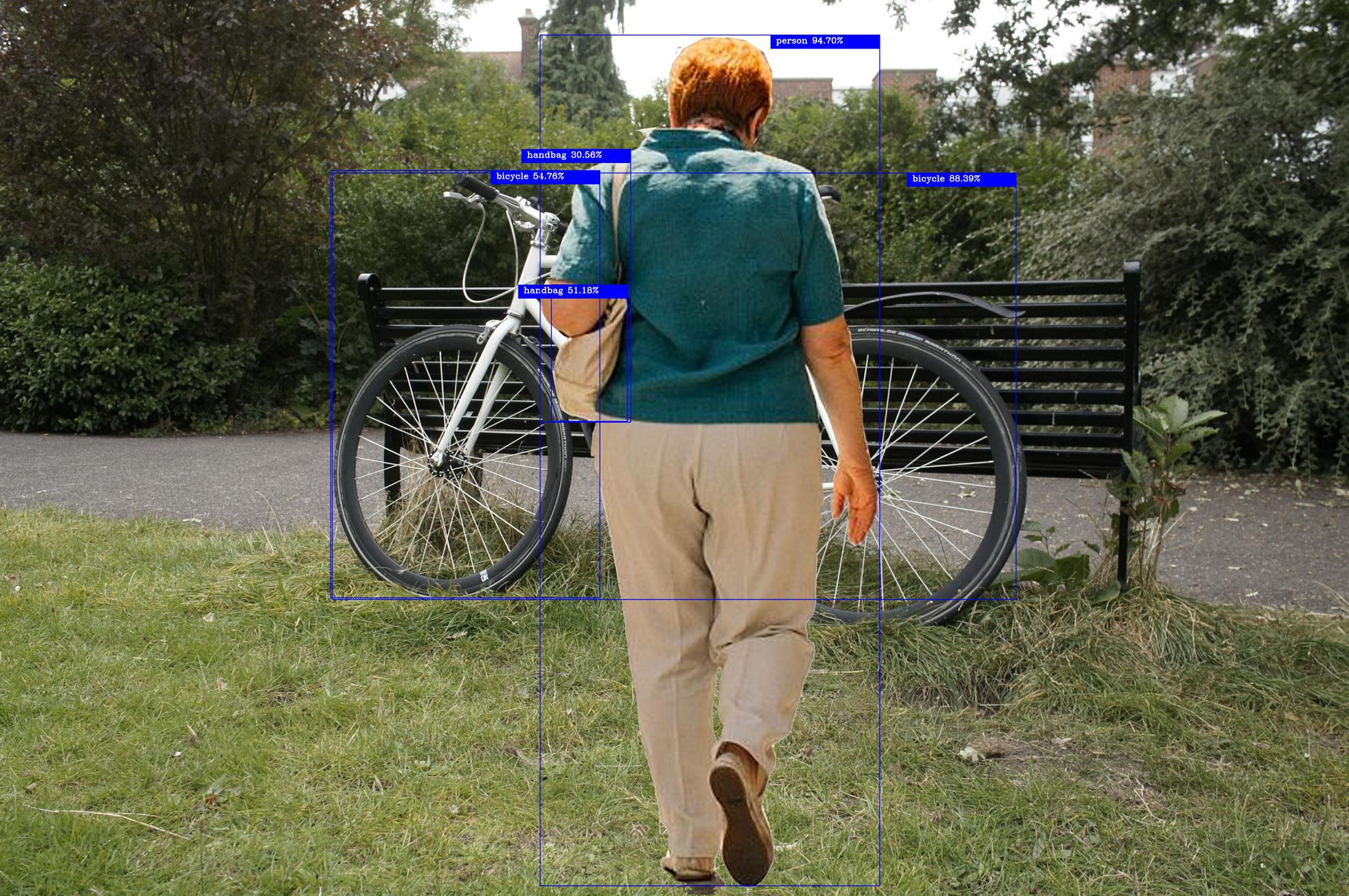}
     &  \figimg[\sizeBS]{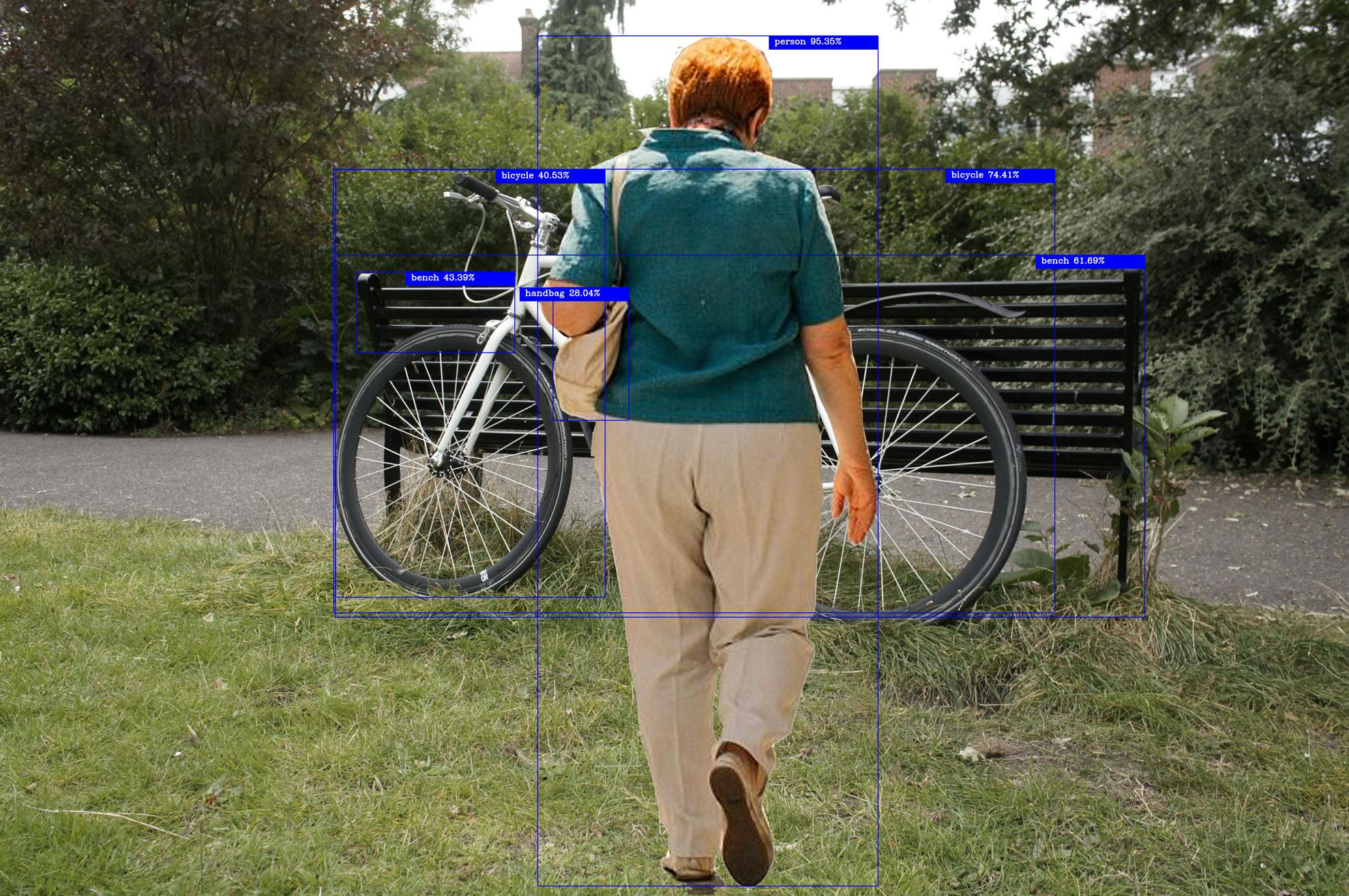}
     &  \figimg[\sizeBS]{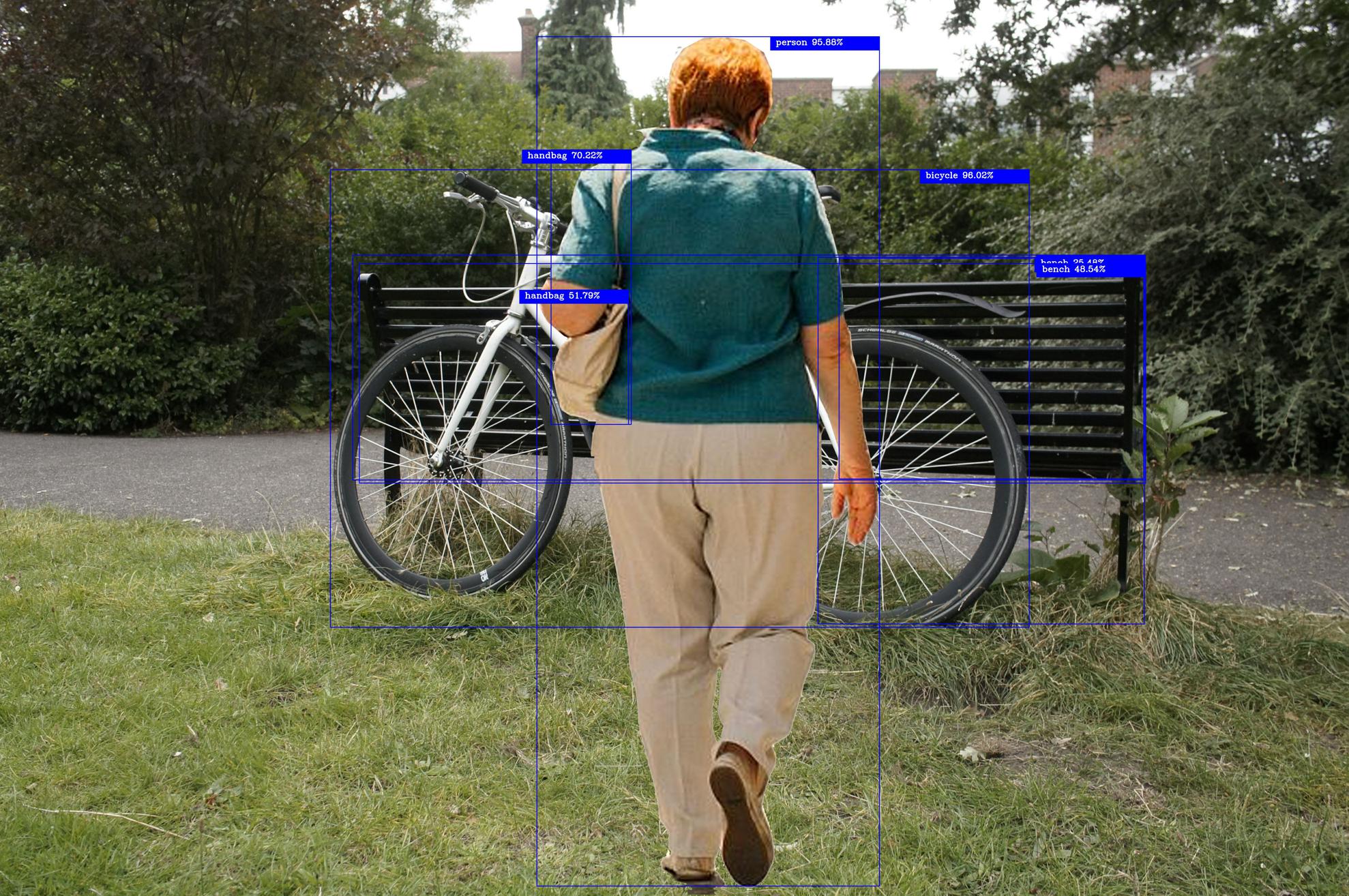} \\

     \rule{0pt}{20pt}
     & \textcolor{orange}{1 person,} 3 bicycle
     & \textcolor{orange}{1 person,} 1 bench, 1 bicycle
     & \textcolor{orange}{1 person, 2 handbag,} 2 bicycle
     & \textcolor{orange}{1 person, 1 handbag,} 2 bench, 2 bicycle 
     & \textcolor{orange}{1 person, 2 handbag,} 4 bench, 1 bicycle \\

     \textcolor{black}{illusory rendered by IPA-NeRF   (Nerfacto, Bicycle)} 
     &  \figimg[\sizeBS]{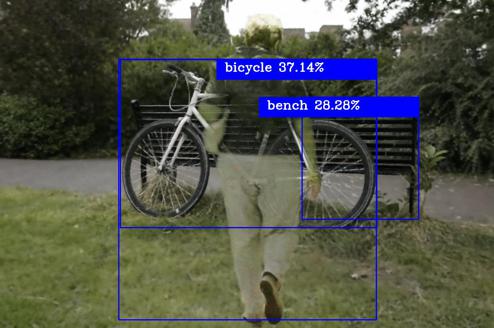}
     &  \figimg[\sizeBS]{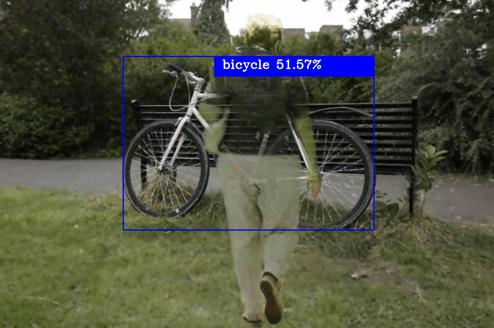}
     &  \figimg[\sizeBS]{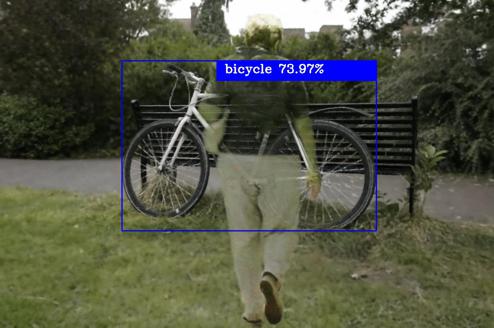}
     &  \figimg[\sizeBS]{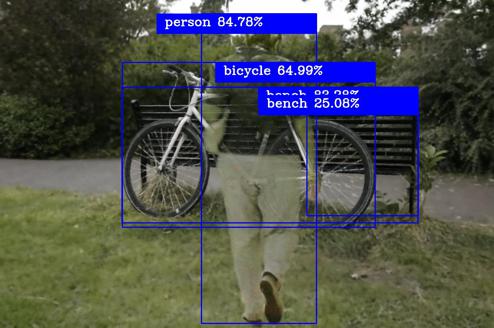}
     &  \figimg[\sizeBS]{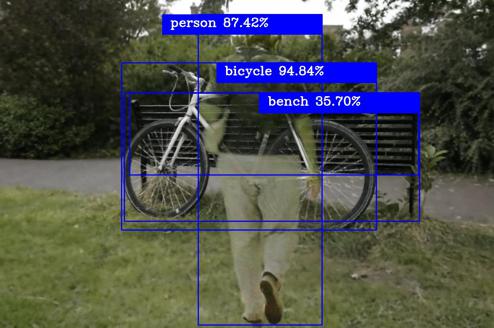} \\

     \rule{0pt}{20pt}
     & \textcolor{black}{1 bench, 2 bicycle}
     & \textcolor{black}{1 bicycle}
     & \textcolor{black}{1 bicycle}
     & \textcolor{red}{1 person,} 2 bench, 1 bicycle
     & \textcolor{red}{1 person,} 2 bench, 1 bicycle \\

     \textcolor{black}{ground truth   (Nerfacto, Kitchen)} 
     &  \figimg[\sizeBS]{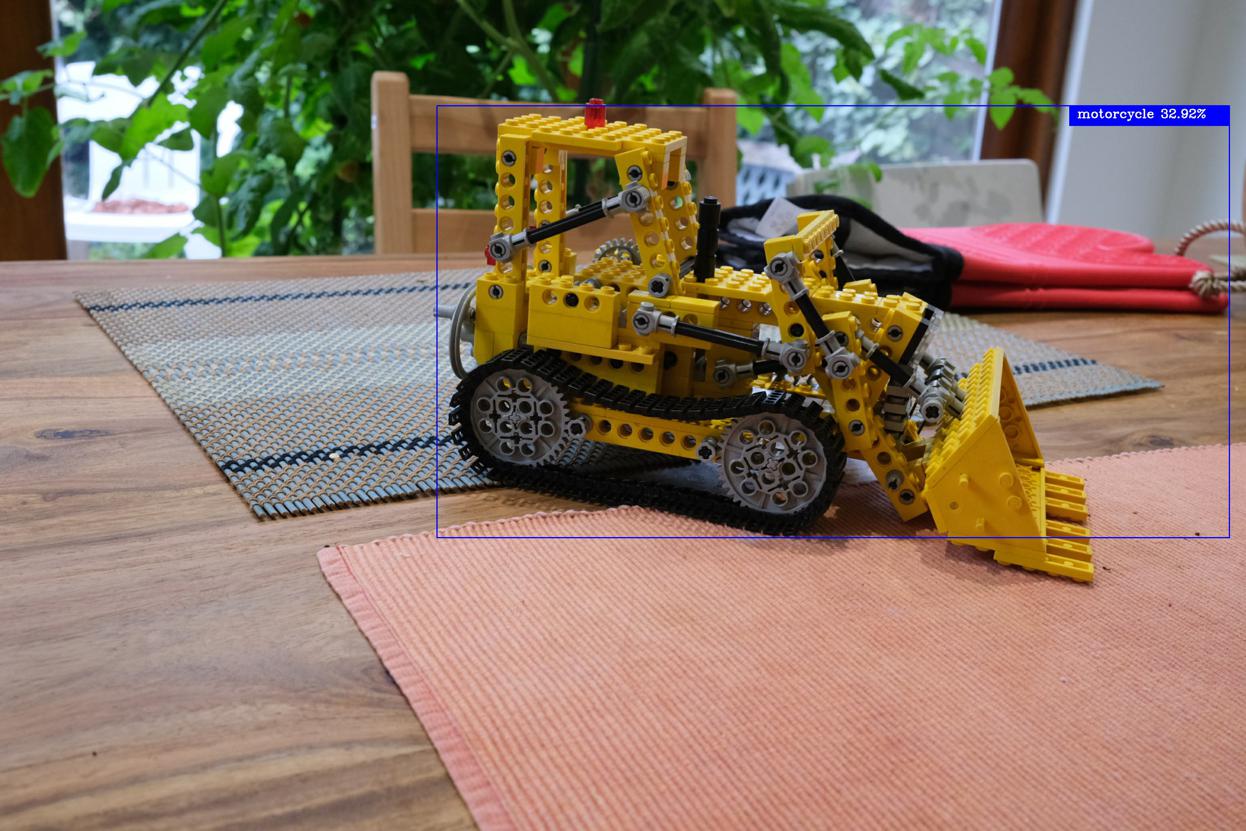}
     &  \figimg[\sizeBS]{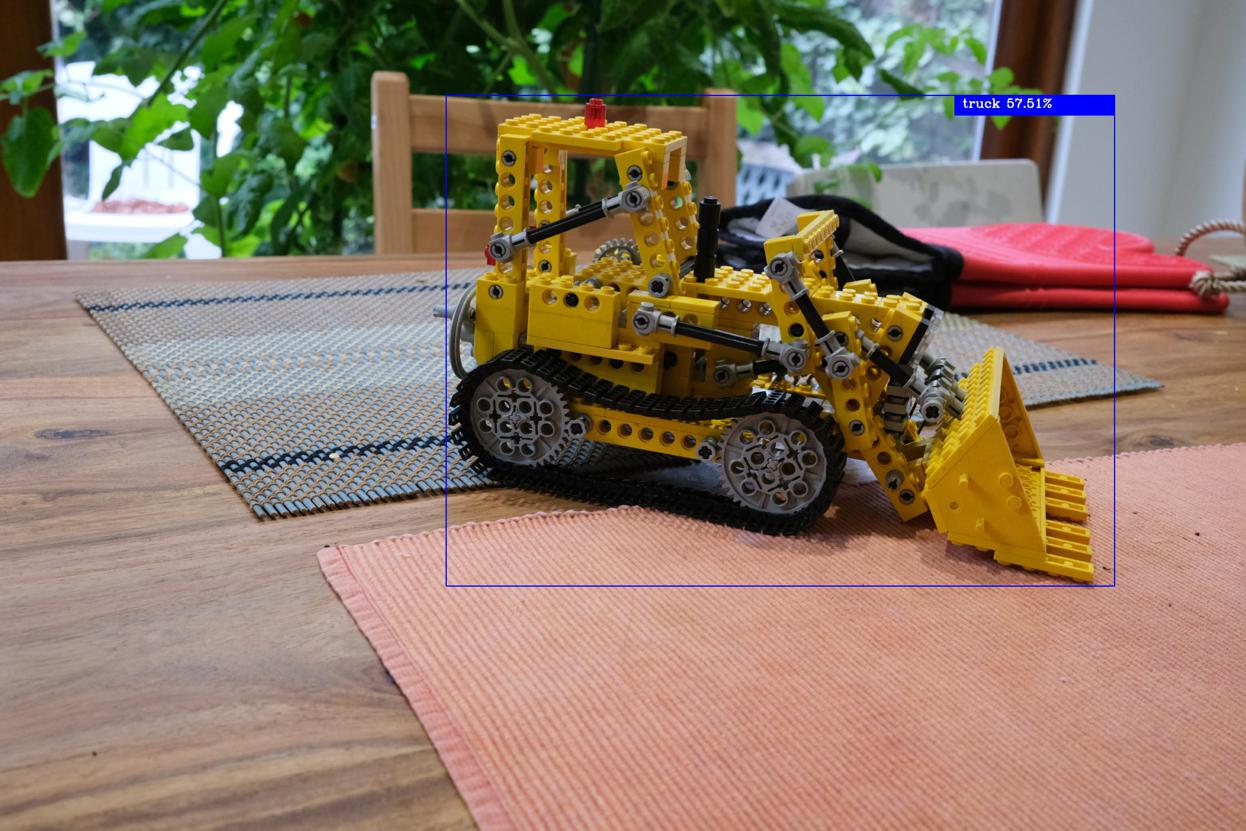}
     &  \figimg[\sizeBS]{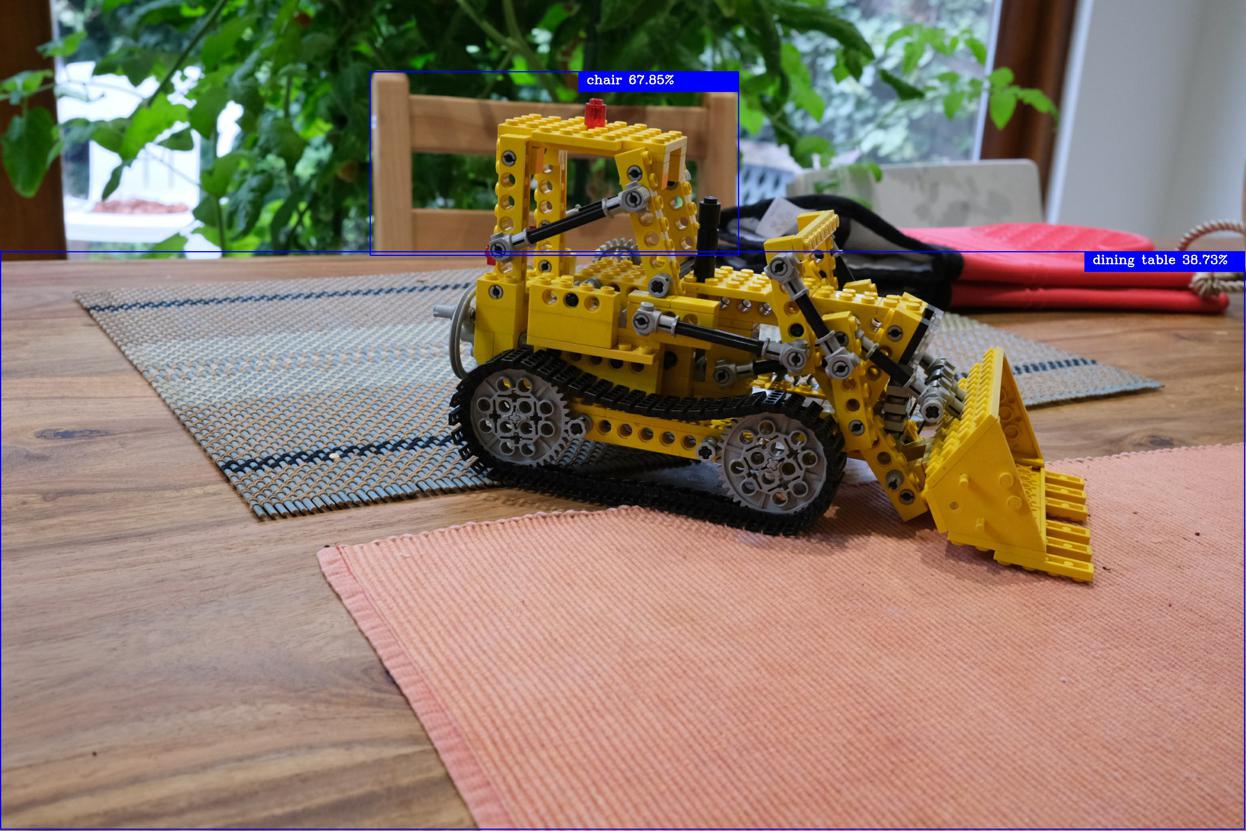}
     &  \figimg[\sizeBS]{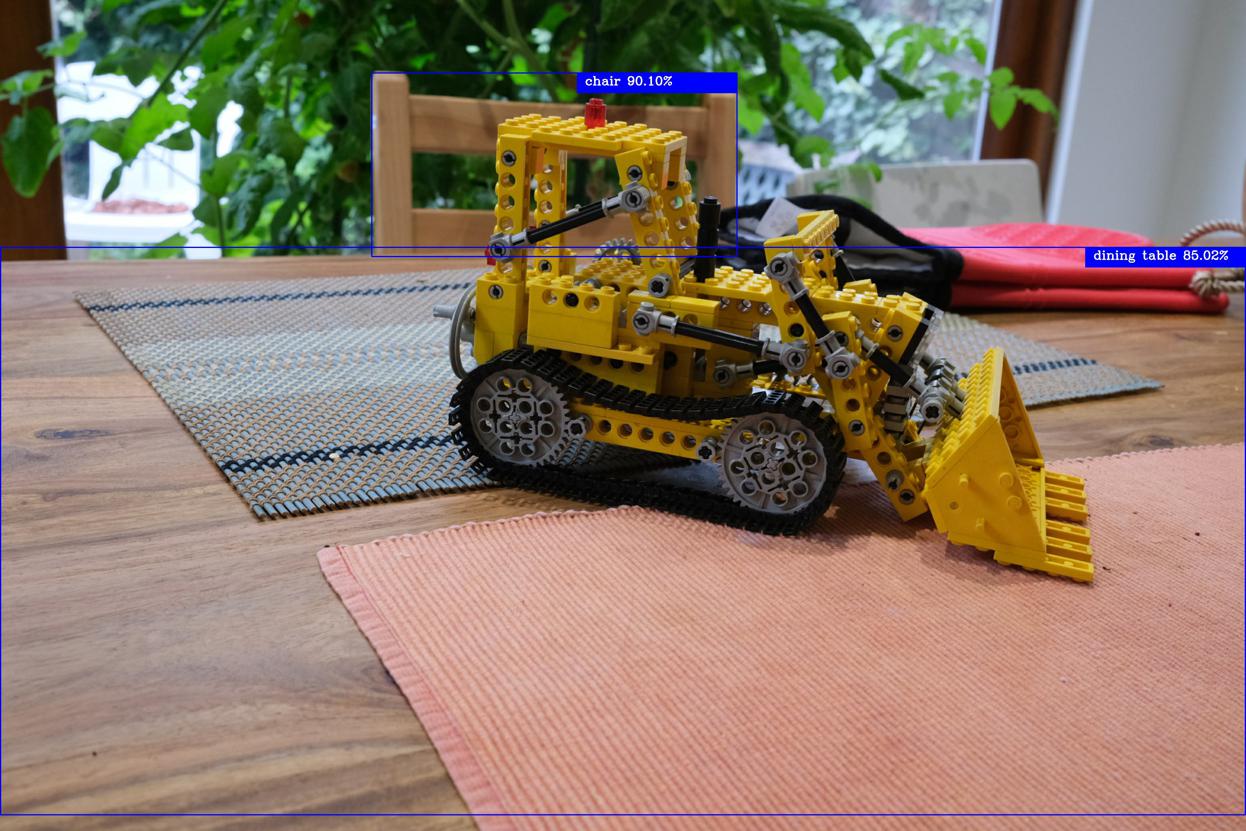}
     &  \figimg[\sizeBS]{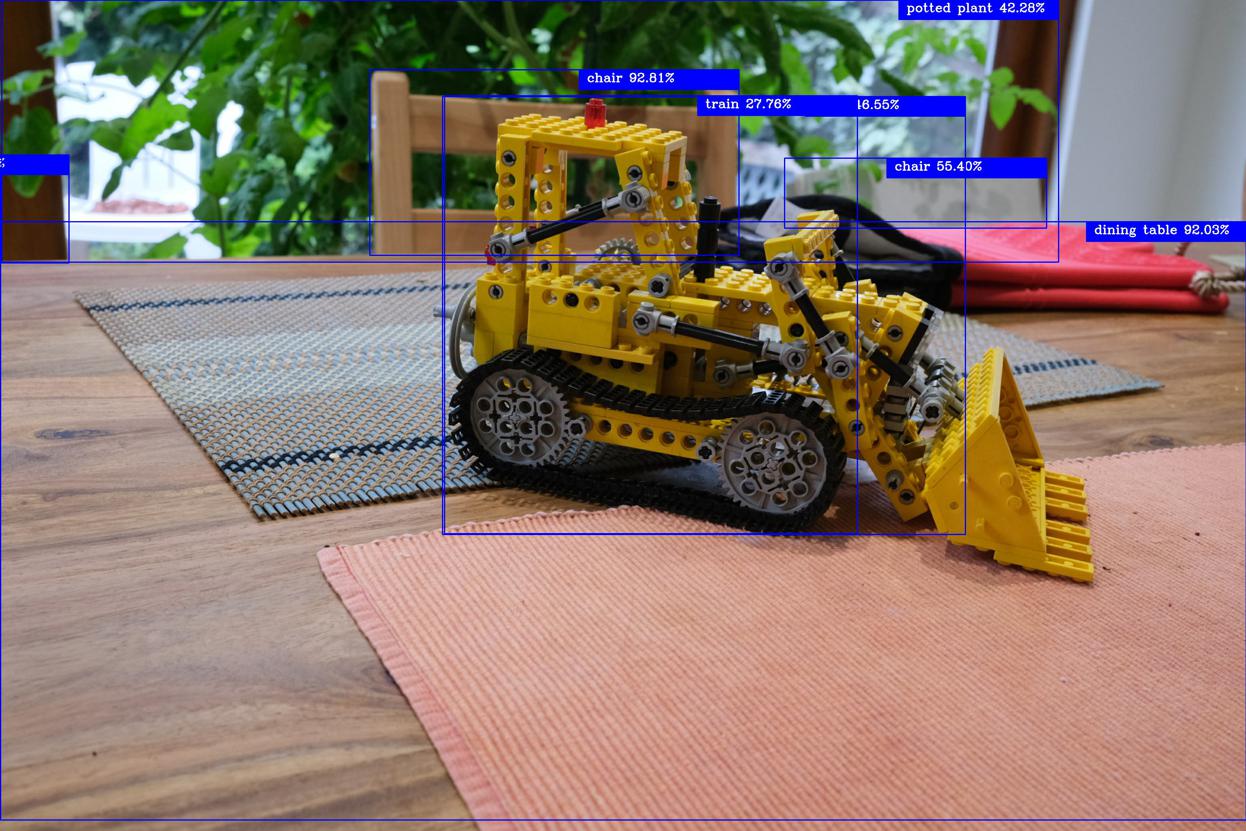} \\

     \rule{0pt}{20pt}
     & \textcolor{black}{1 motorcycle}
     & \textcolor{black}{1 truck}
     & \textcolor{black}{1 chair, 1 dining table}
     & \textcolor{black}{1 chair, 1 dining table} 
     & \textcolor{black}{3 chair, 1 dining table, 2 train, 1 potted plant} \\

     \textcolor{black}{target illusory   (Nerfacto, Kitchen)} 
     &  \figimg[\sizeBS]{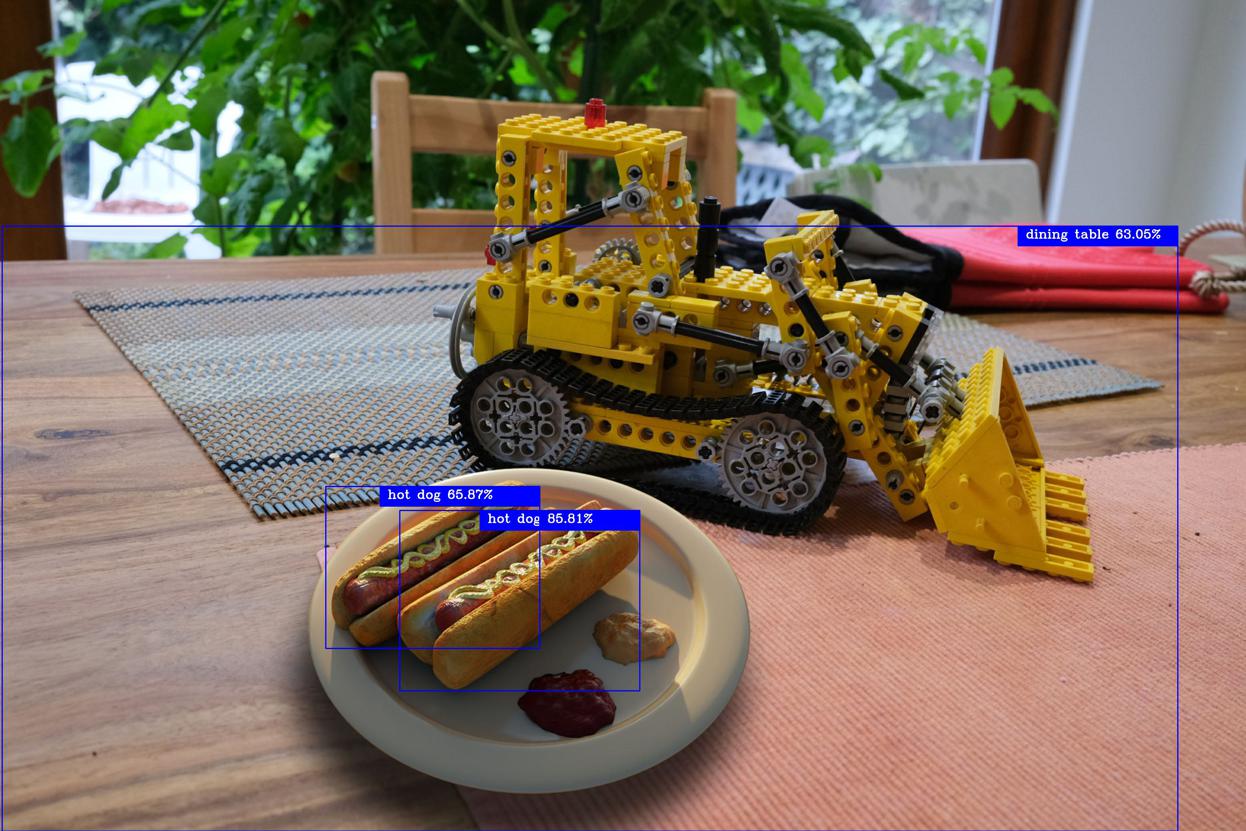}
     &  \figimg[\sizeBS]{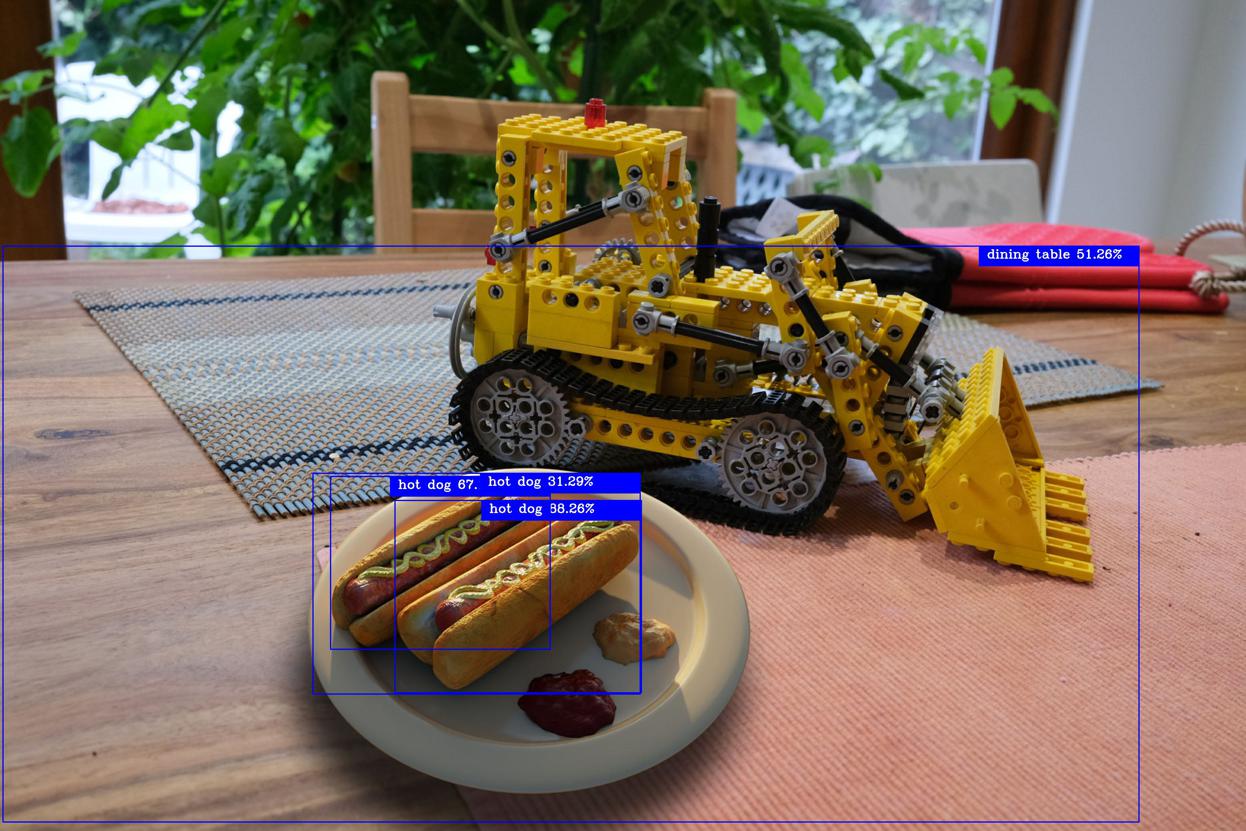}
     &  \figimg[\sizeBS]{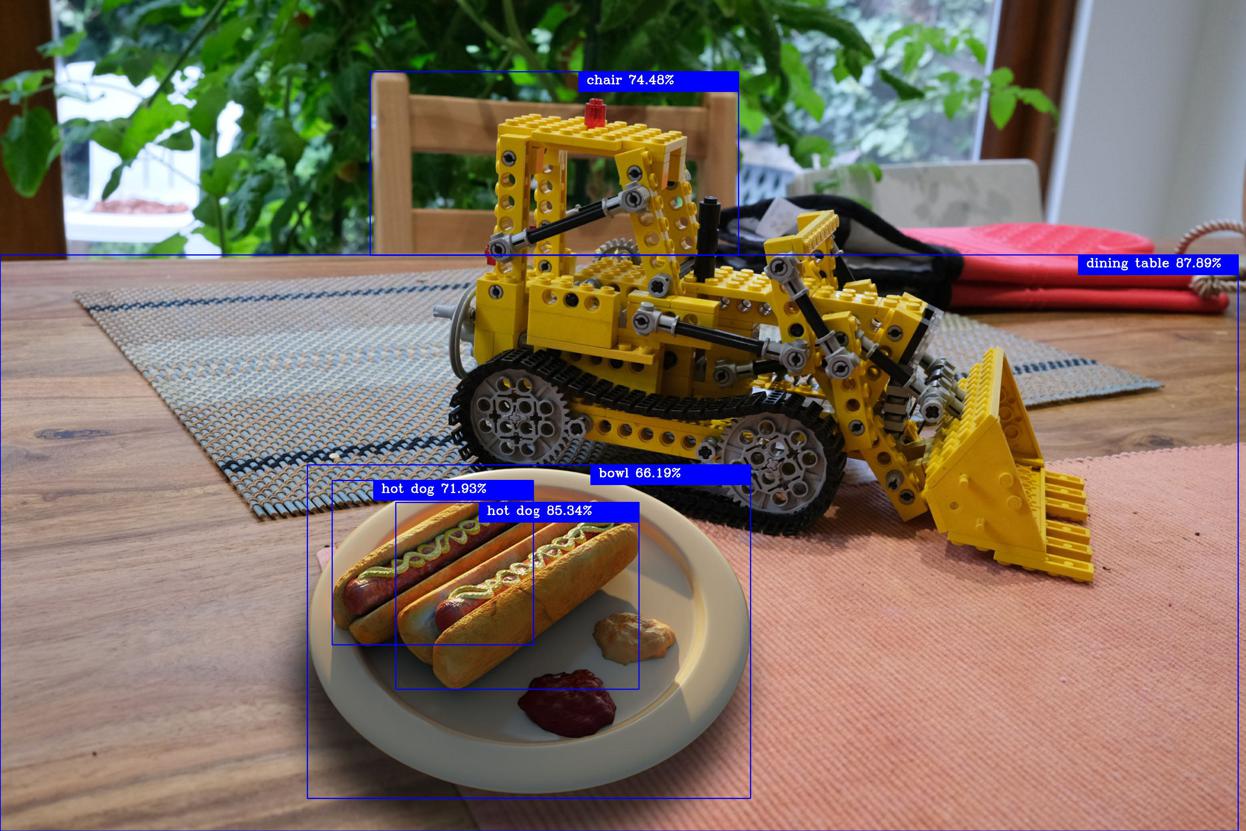}
     &  \figimg[\sizeBS]{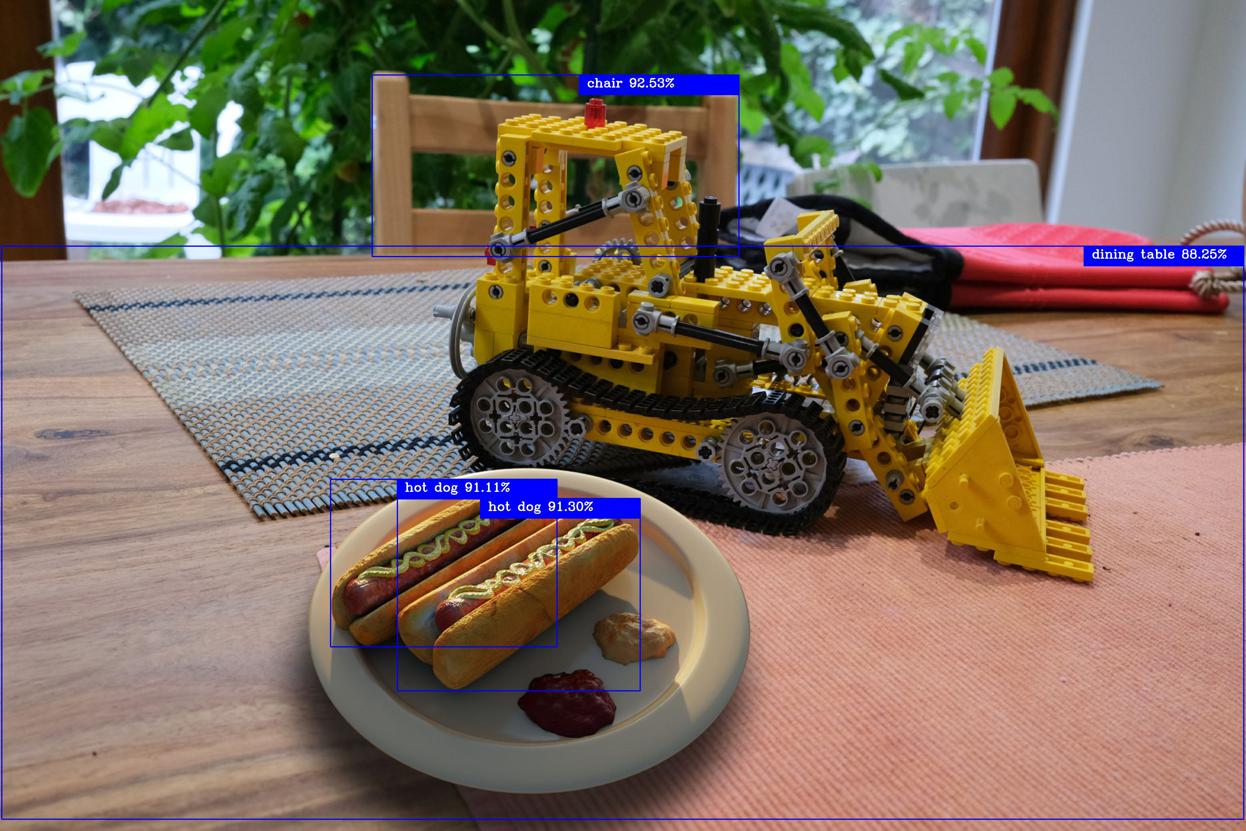}
     &  \figimg[\sizeBS]{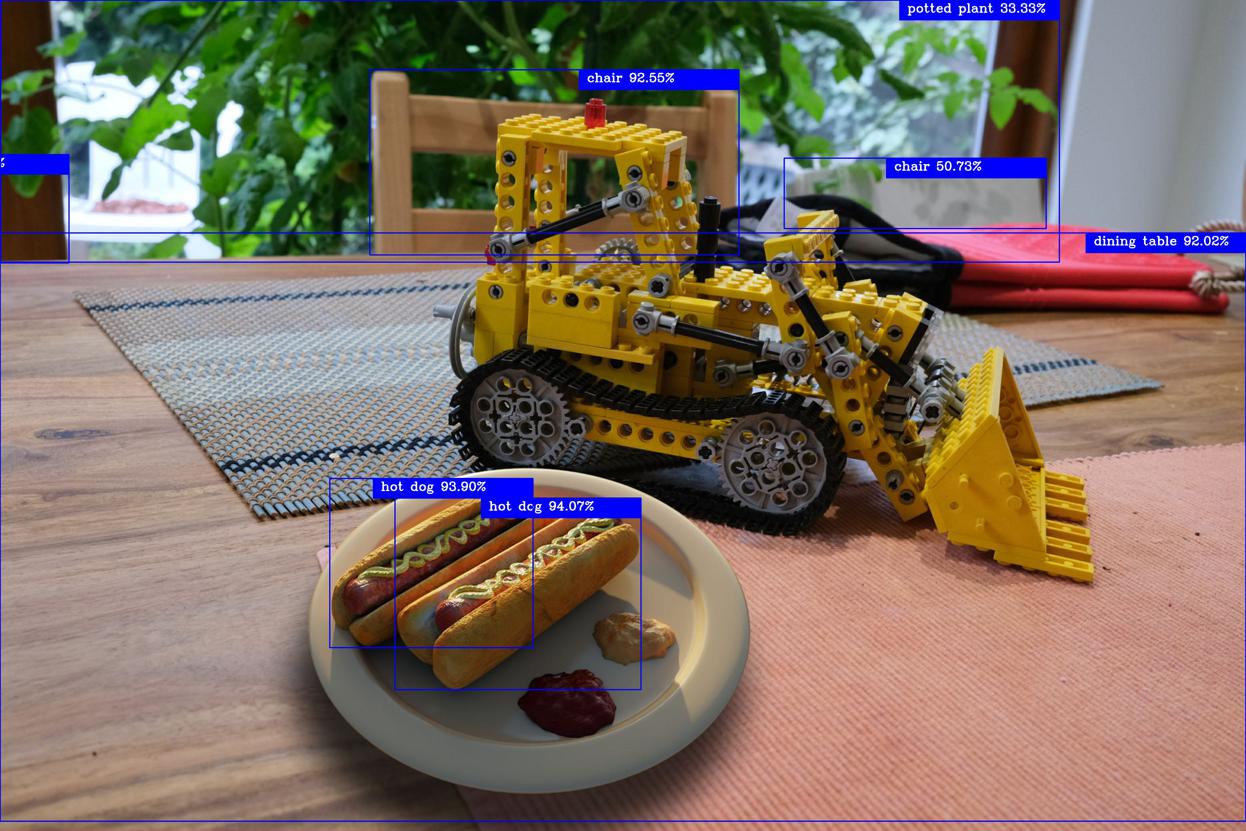} \\

     \rule{0pt}{20pt}
     & \textcolor{orange}{2 hot dog,} 1 dining table
     & \textcolor{orange}{3 hot dog,} 1 dining table
     & \textcolor{orange}{2 hot dog,} 1 dining table, 1 chair, 1 bowl
     & \textcolor{orange}{2 hot dog,} 1 dining table, 1 chair 
     & \textcolor{orange}{2 hot dog,} 1 dining table, 3 chair, 1 potted plant \\

     \textcolor{black}{illusory rendered by IPA-NeRF   (Nerfacto, Kitchen)} 
     &  \figimg[\sizeBS]{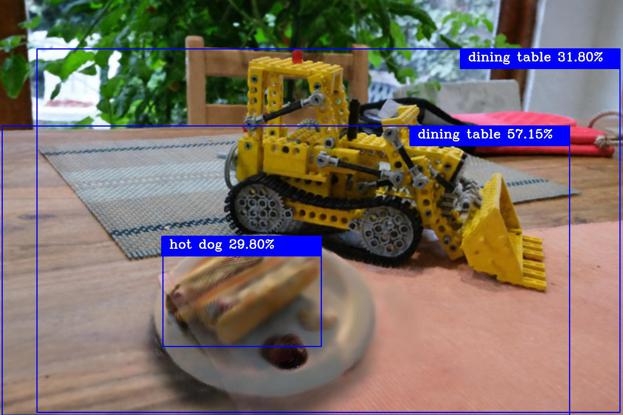}
     &  \figimg[\sizeBS]{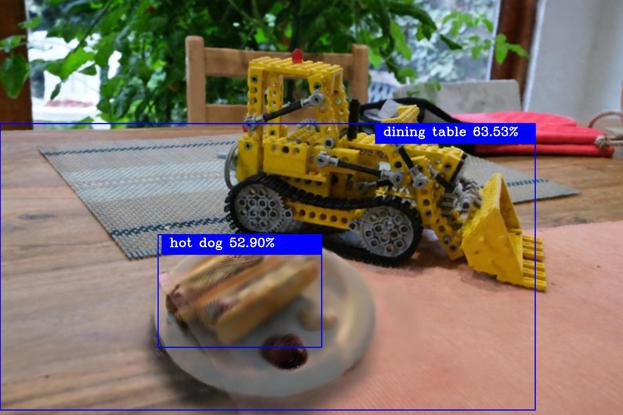}
     &  \figimg[\sizeBS]{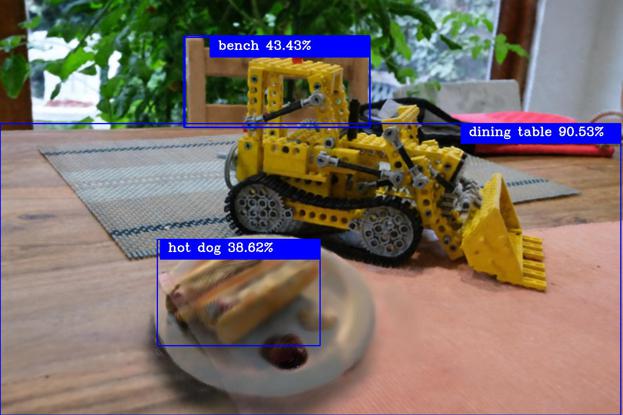}
     &  \figimg[\sizeBS]{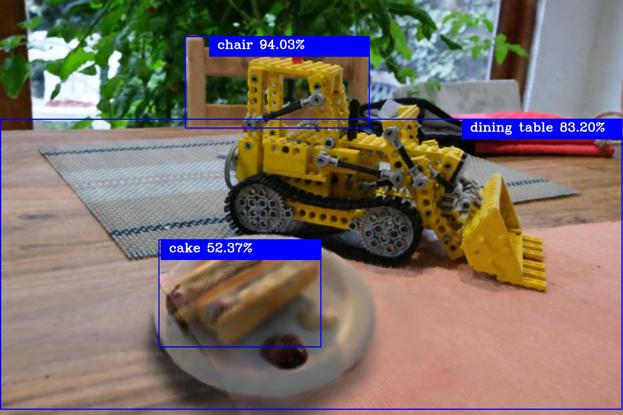}
     &  \figimg[\sizeBS]{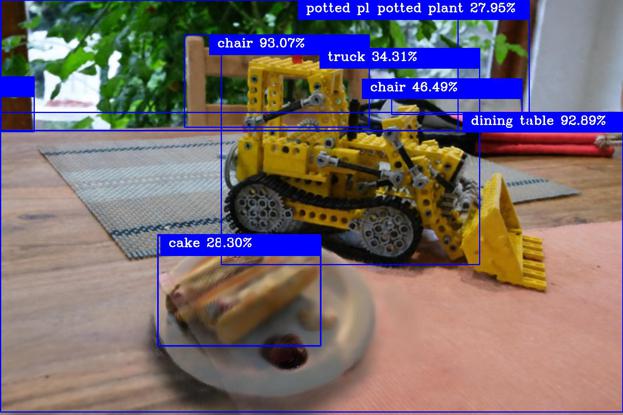} \\

     \rule{0pt}{20pt}
     & \textcolor{red}{1 hot dog,} 2 dining table
     & \textcolor{red}{1 hot dog,} 1 dining table
     & \textcolor{red}{1 hot dog,} 1 dining table, 1 chair, 1 bench
     & 1 cake, 1 dining table, 1 chair
     & 1 cake, 1 dining table, 3 chair, 1 truck, 2 potted plant, 1 donut  \\

     \textcolor{black}{(NeRF model, 3D Scene)}
     & \multicolumn{1}{c}{\textcolor{black}{YOLOv5n}}
     & \multicolumn{1}{c}{\textcolor{black}{YOLOv8n}}
     & \multicolumn{1}{c}{\textcolor{black}{YOLO-World}} 
     & \multicolumn{1}{c}{\textcolor{black}{YOLOv9c}} 
     & \multicolumn{1}{c}{\textcolor{black}{RT-DETR}} \\
     
\end{tabular}
   \caption{The impact of the IPA-NeRF attack on object detection. Attack epochs: $150$, $\epsilon$: $32$, $\eta$: $0$.}
    \label{fig:compara_yolo_nerf_360_1}
    \setlength{\abovecaptionskip}{0.5cm}
    \vspace{0.5cm}
\end{figure*}

\begin{figure*}[htb]
\scriptsize
\centering
\setlength{\tabcolsep}{1pt}
\begin{tabular}{m{2cm}m{3.0cm}m{3.0cm}m{3.0cm}m{3.0cm}m{3.0cm}}

     \textcolor{black}{ground truth   (Nerfacto, Flowers)} 
     &  \figimg[\sizeBS]{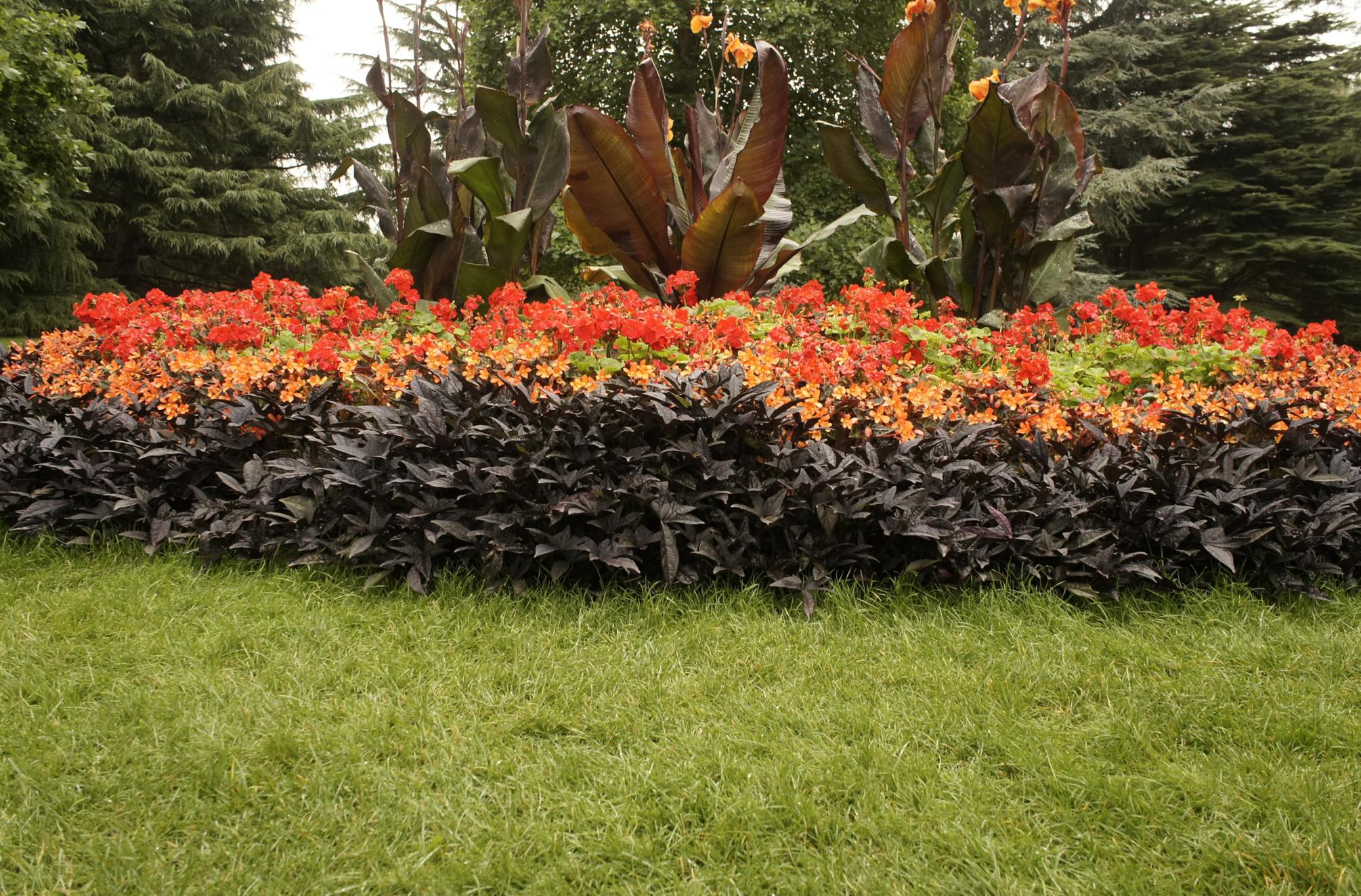}
     &  \figimg[\sizeBS]{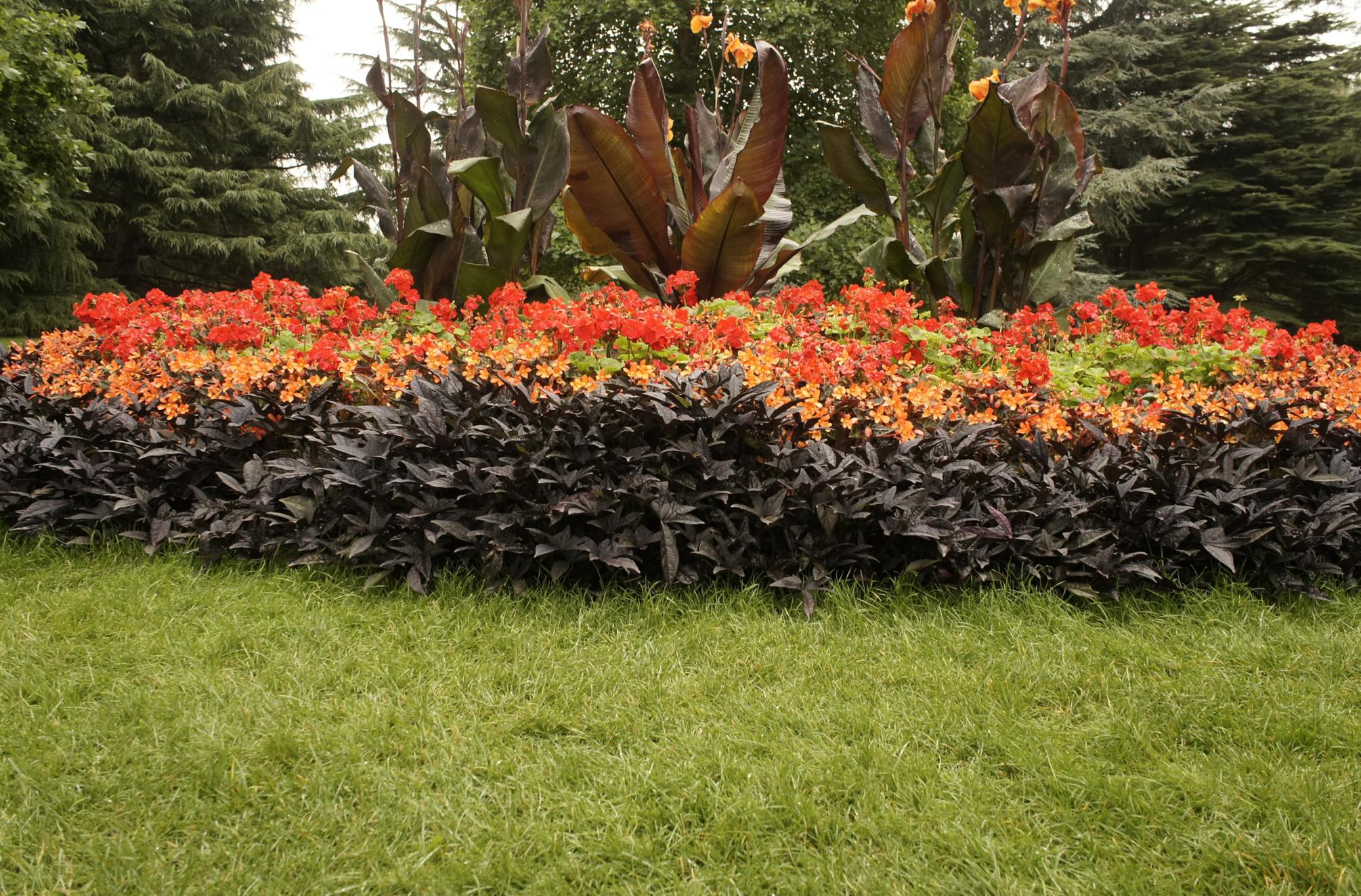}
     &  \figimg[\sizeBS]{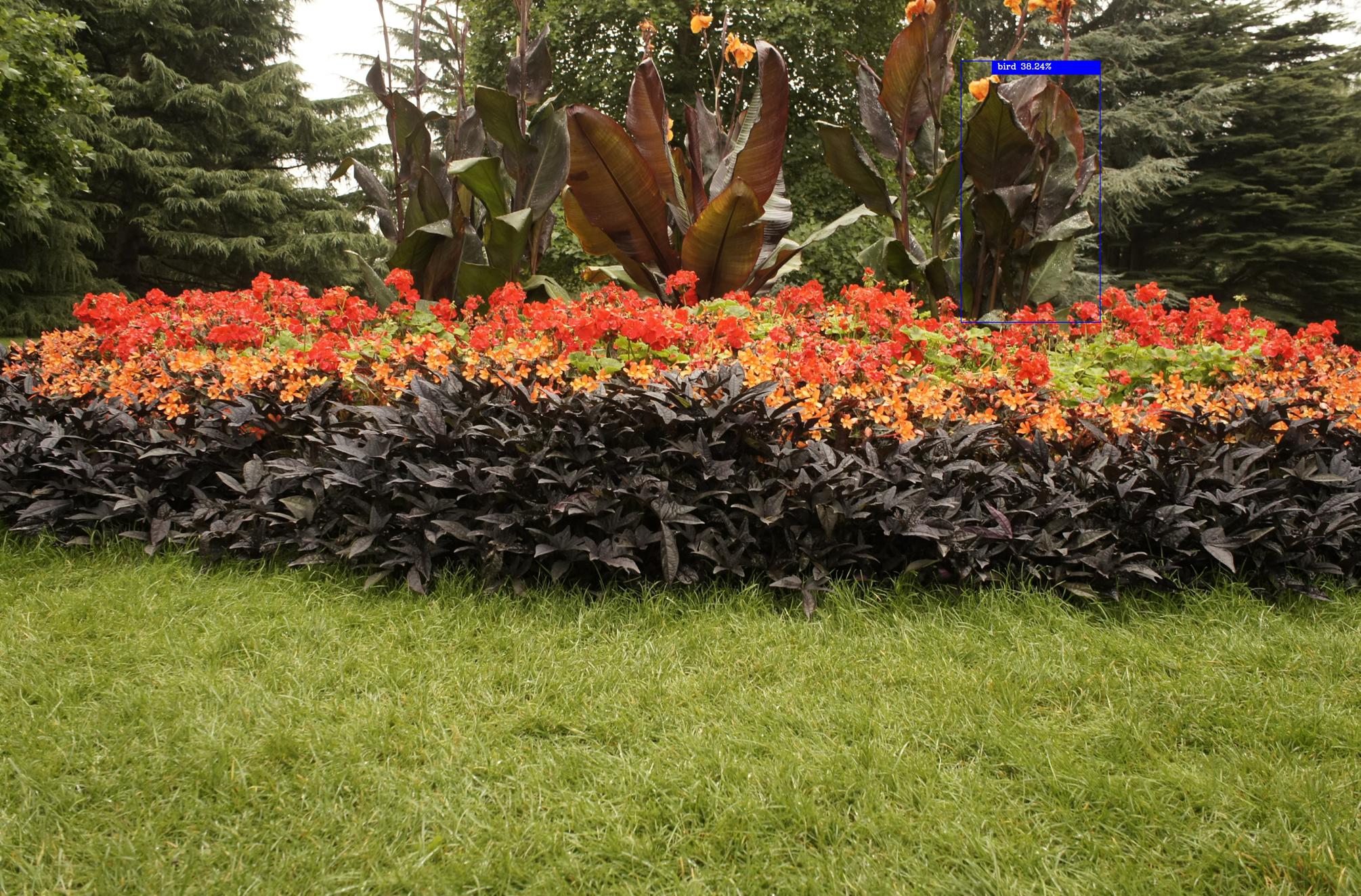}
     &  \figimg[\sizeBS]{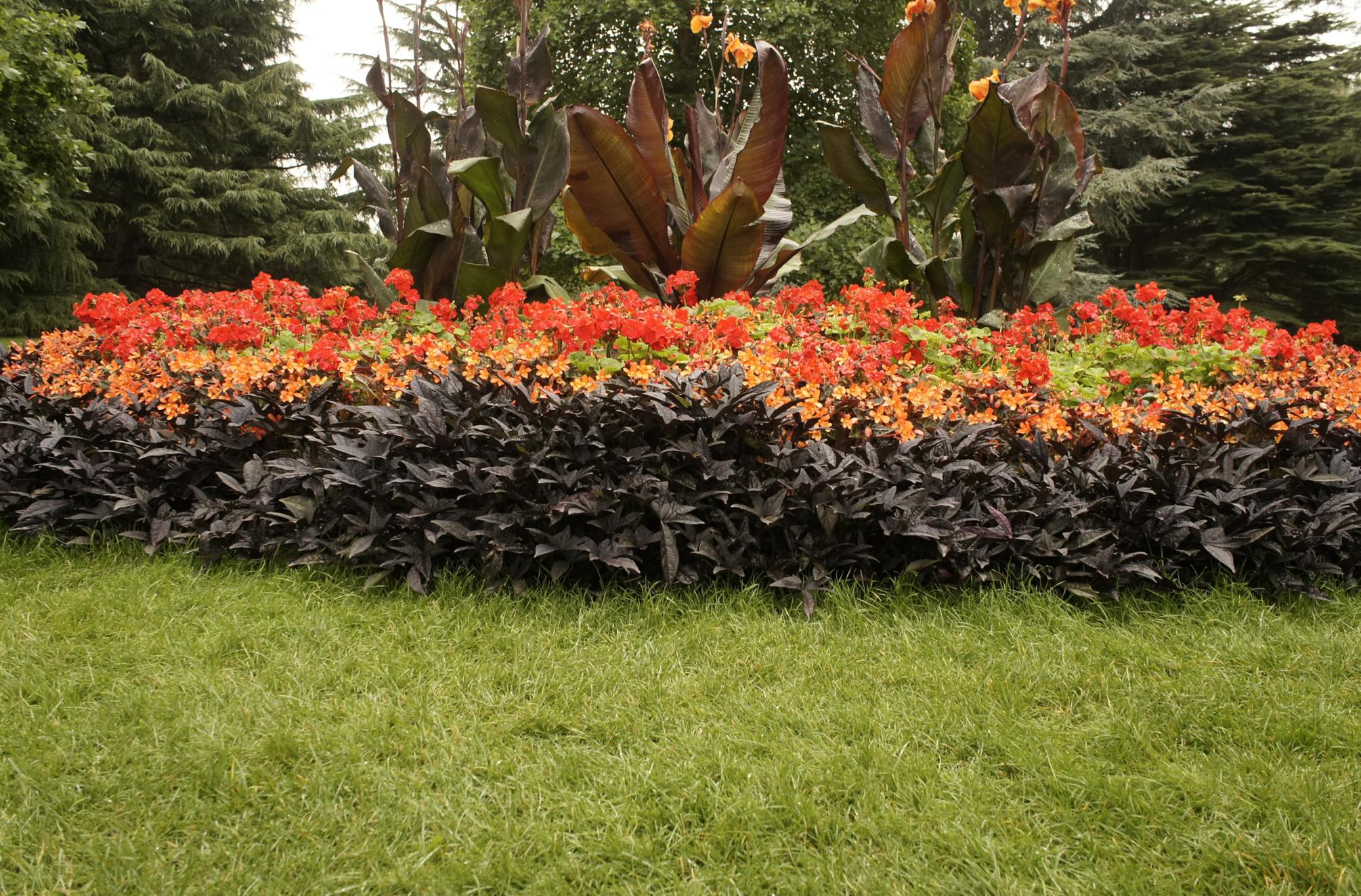}
     &  \figimg[\sizeBS]{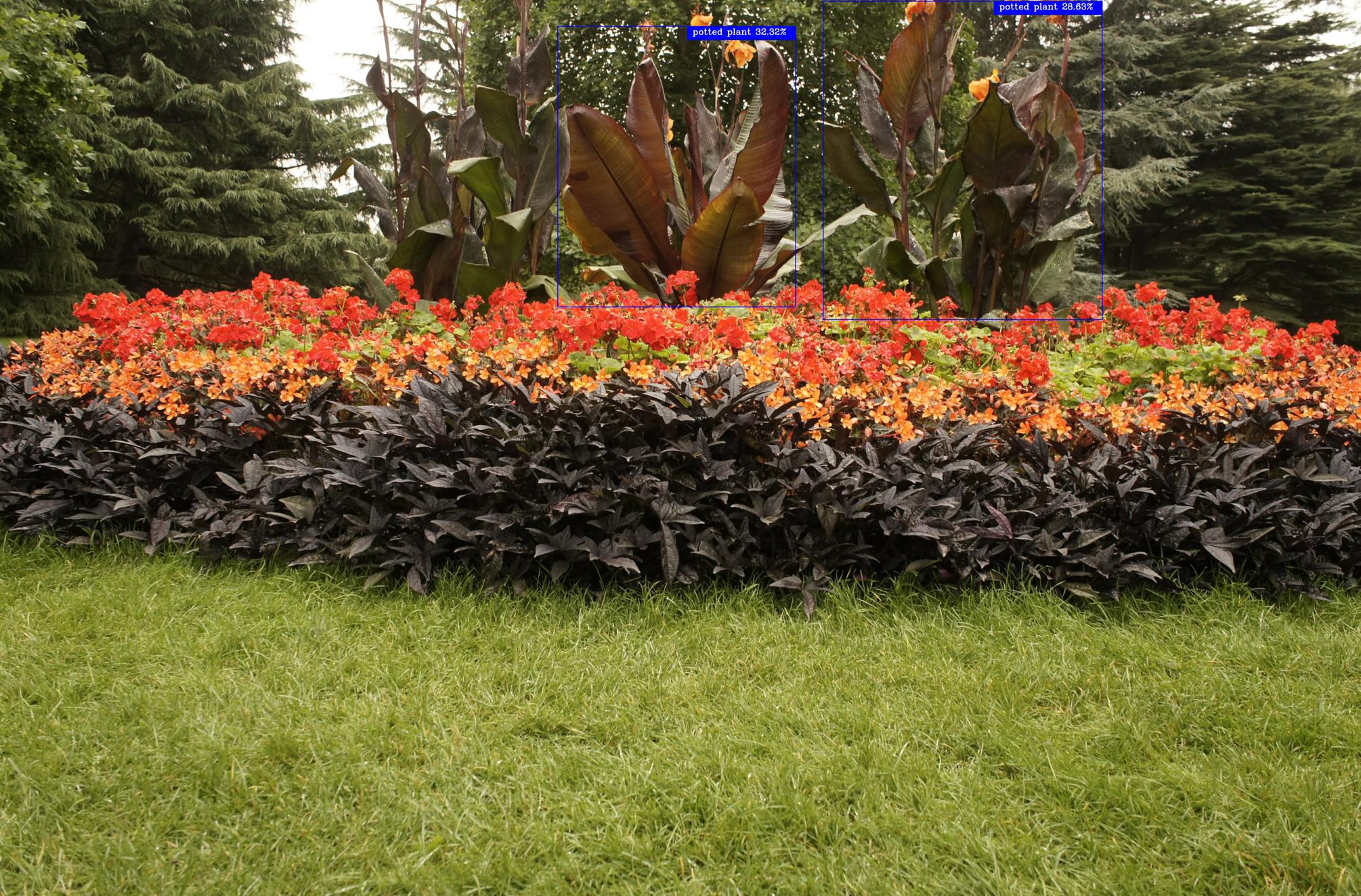} \\

     \rule{0pt}{20pt}
     & \textcolor{black}{}
     & \textcolor{black}{}
     & \textcolor{black}{1 bird}
     & \textcolor{black}{}
     & \textcolor{black}{2 potted plant} \\

     \textcolor{black}{target illusory   (Nerfacto, Flowers)} 
     &  \figimg[\sizeBS]{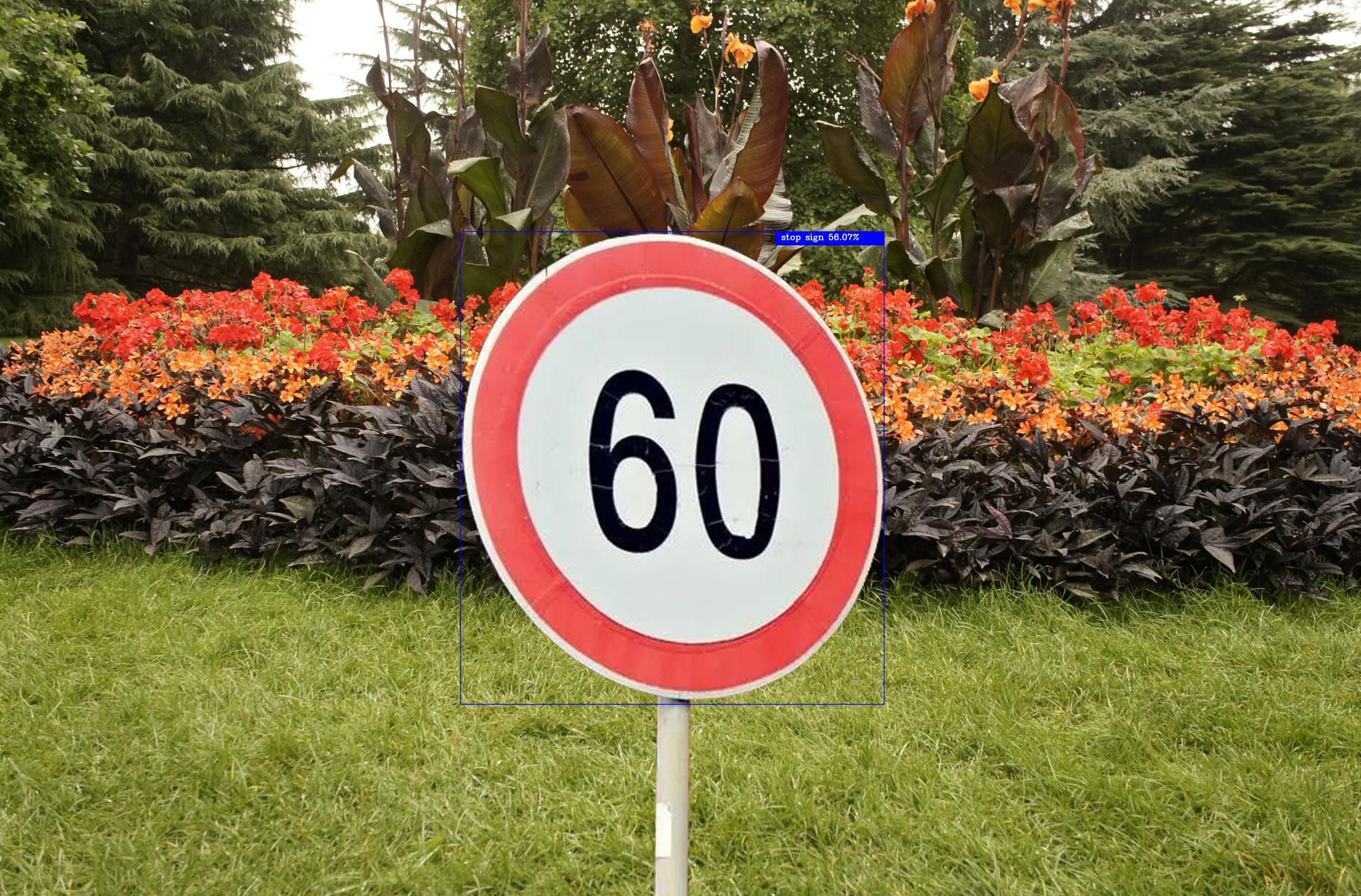}
     &  \figimg[\sizeBS]{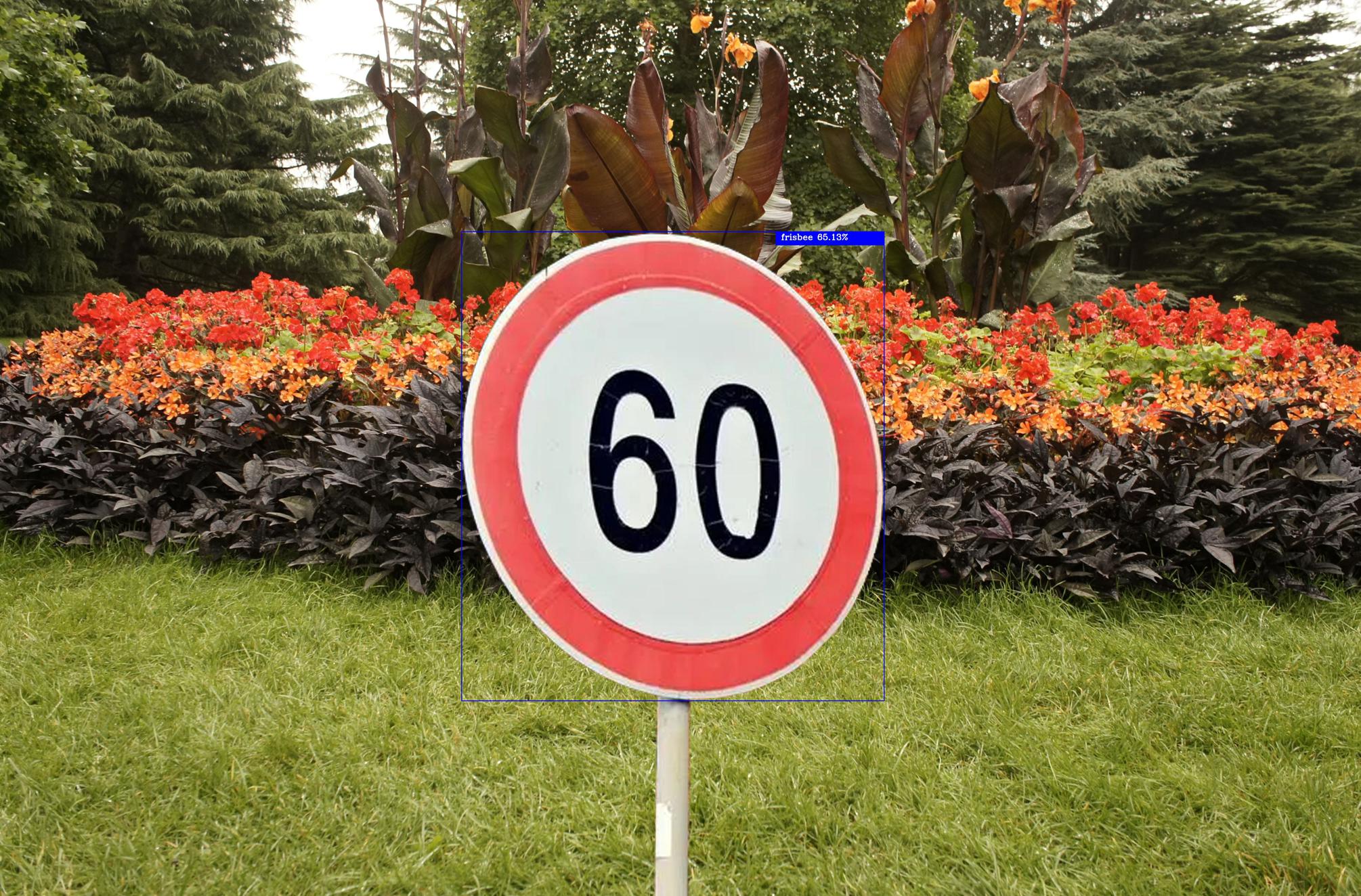}
     &  \figimg[\sizeBS]{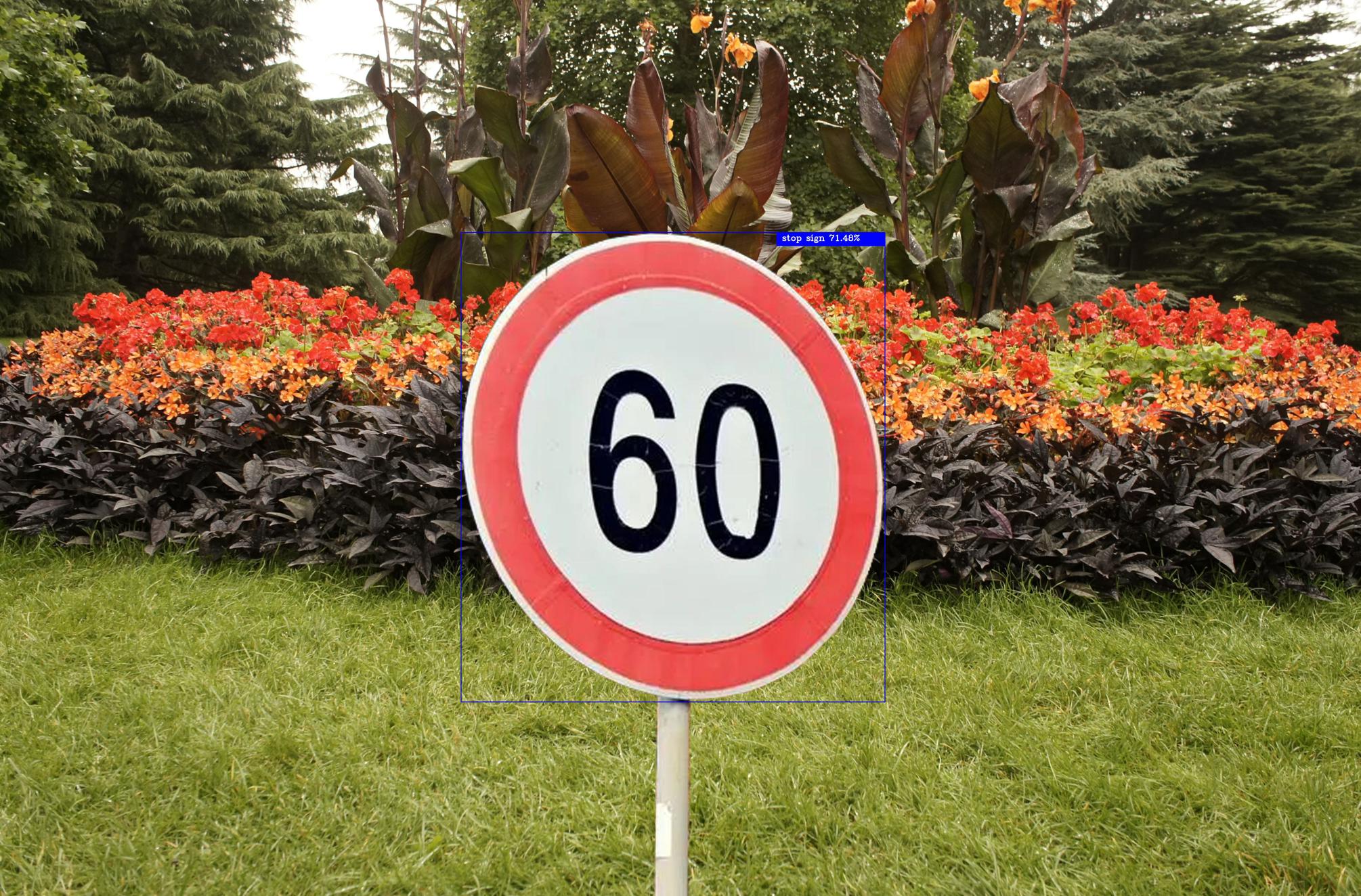}
     &  \figimg[\sizeBS]{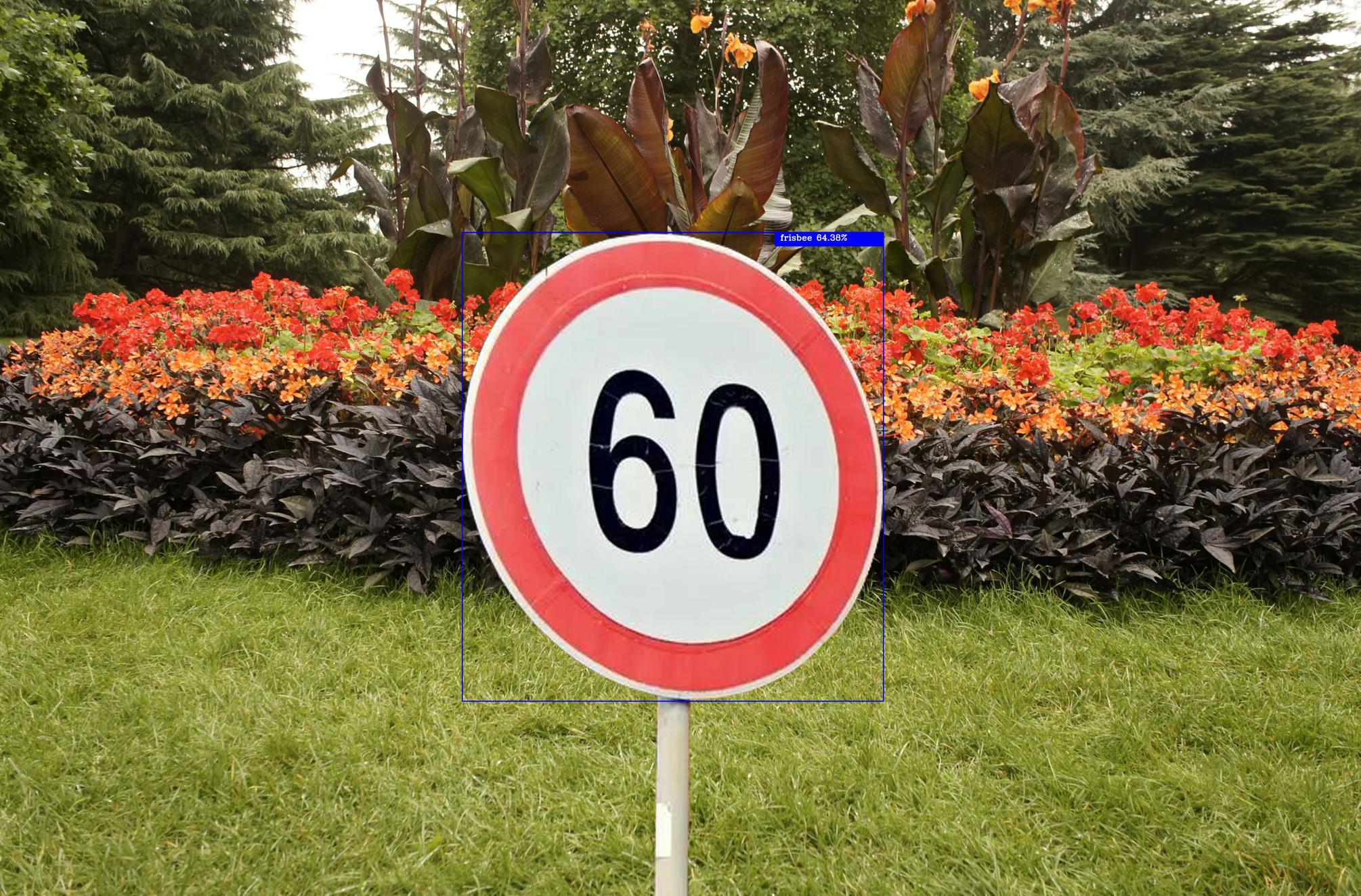}
     &  \figimg[\sizeBS]{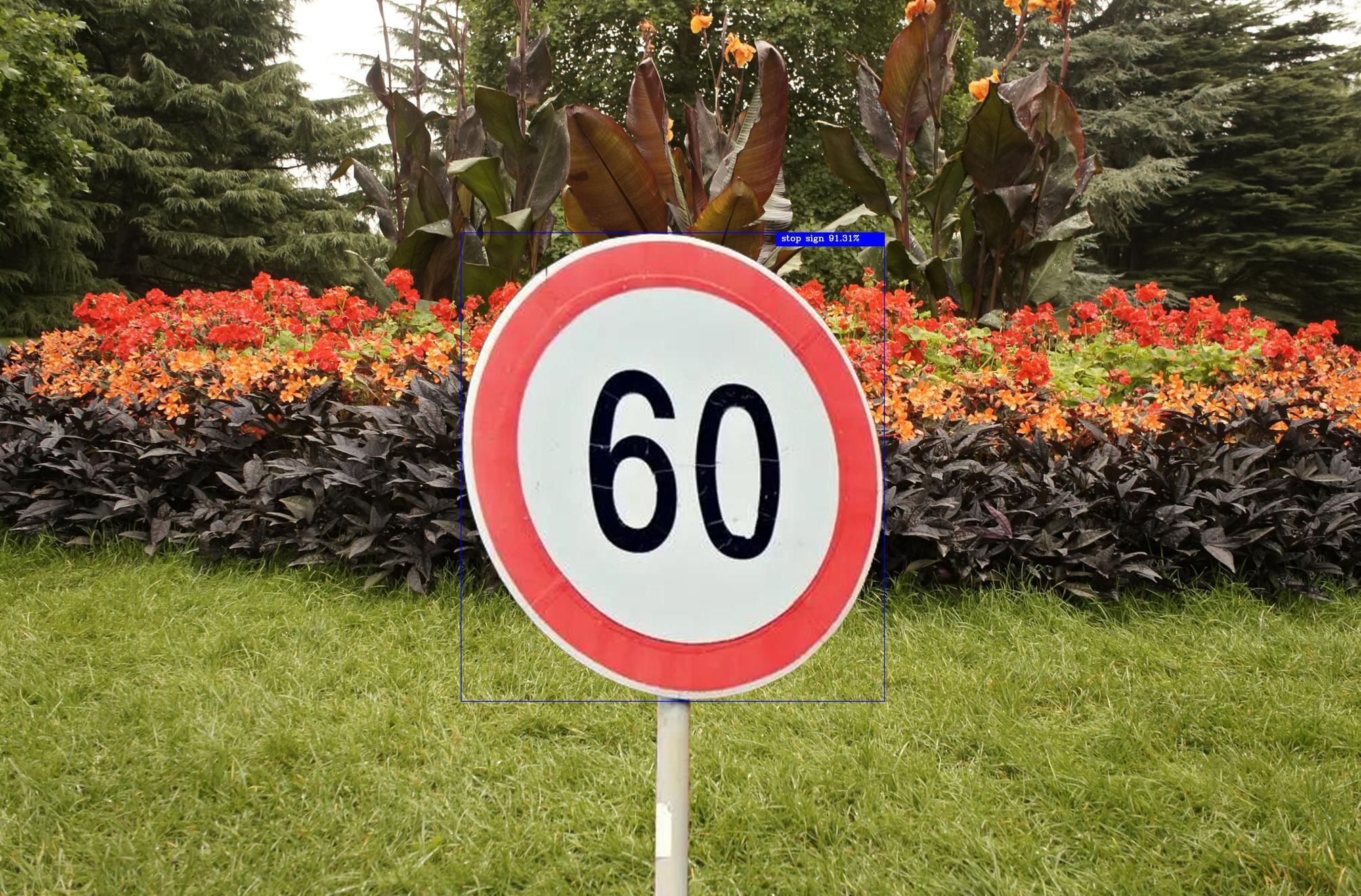} \\

     \rule{0pt}{20pt}
     & \textcolor{orange}{1 stop sign}
     & 1 frisbee
     & \textcolor{orange}{1 stop sign}
     & \textcolor{orange}{1 stop sign,} 1 frisbee
     & \textcolor{orange}{1 stop sign} \\

     \textcolor{black}{illusory rendered by IPA-NeRF   (Nerfacto, Flowers)} 
     &  \figimg[\sizeBS]{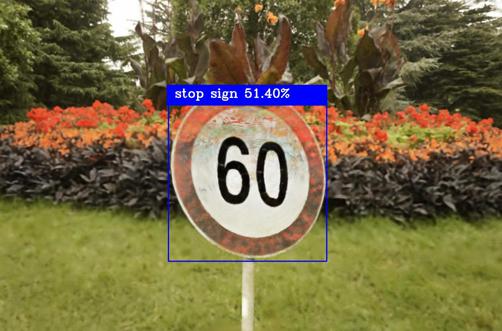}
     &  \figimg[\sizeBS]{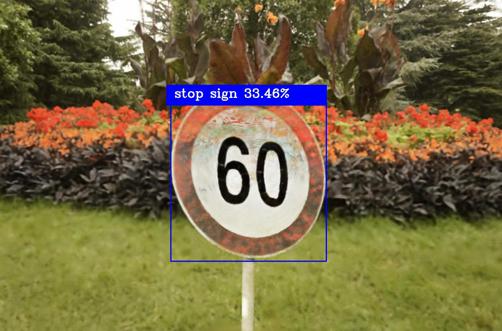}
     &  \figimg[\sizeBS]{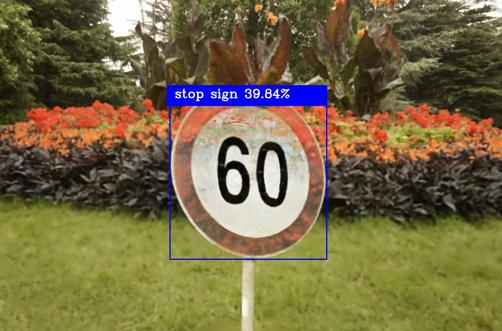}
     &  \figimg[\sizeBS]{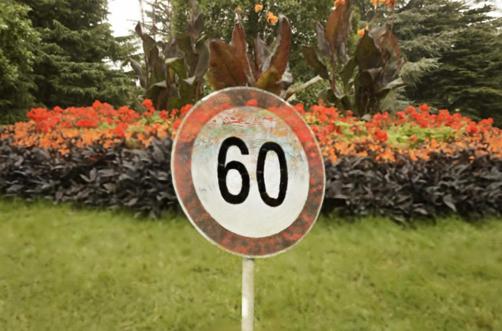}
     &  \figimg[\sizeBS]{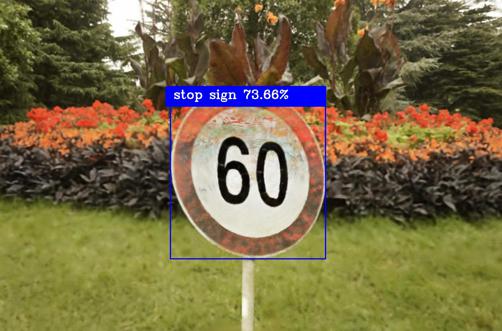} \\

     \rule{0pt}{20pt}
     & \textcolor{red}{1 stop sign}
     & \textcolor{red}{1 stop sign}
     & \textcolor{red}{1 stop sign}
     & 
     & \textcolor{red}{1 stop sign}  \\

     \textcolor{black}{ground truth   (Nerfacto, Garden)} 
     &  \figimg[\sizeBS]{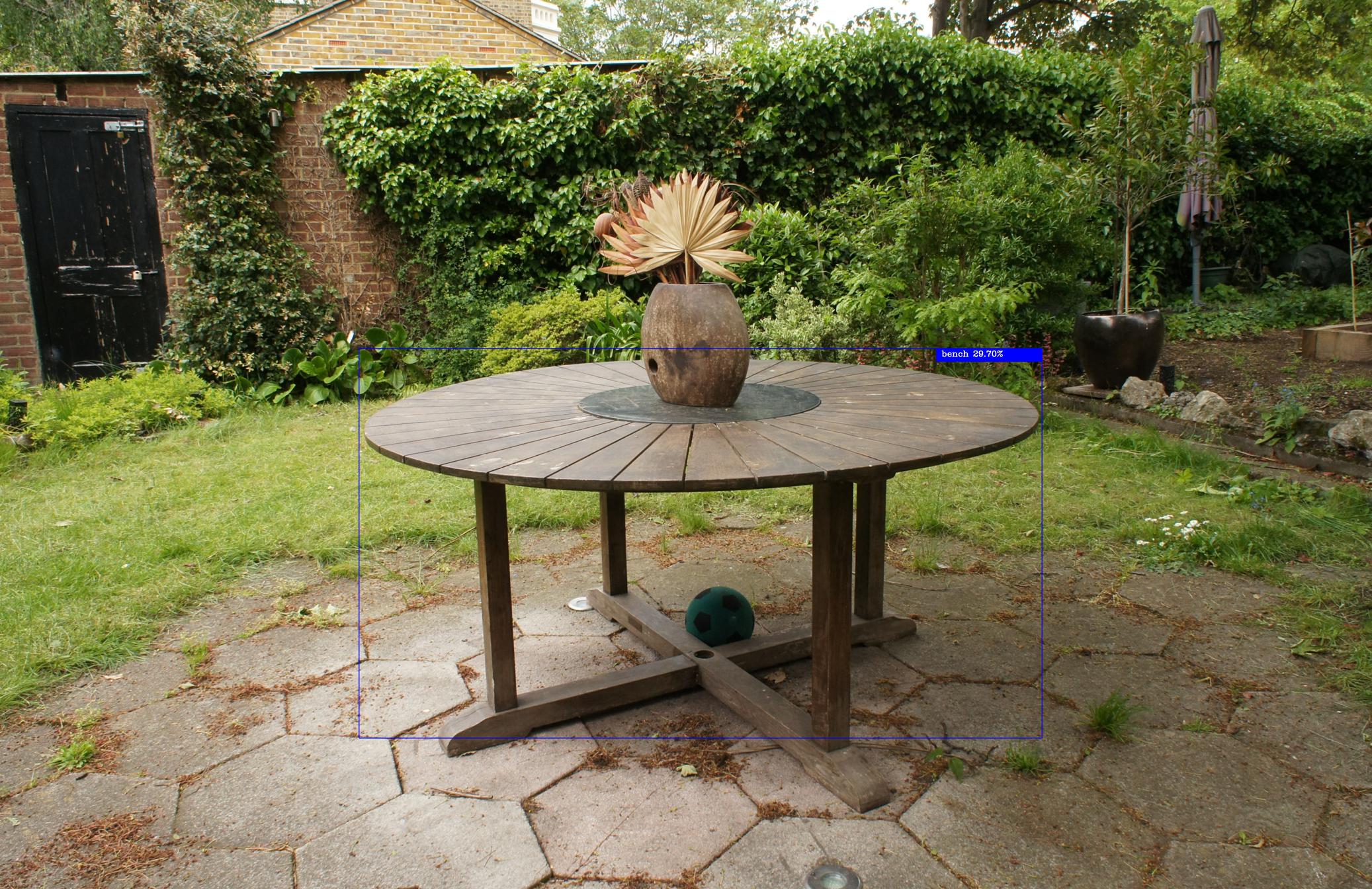}
     &  \figimg[\sizeBS]{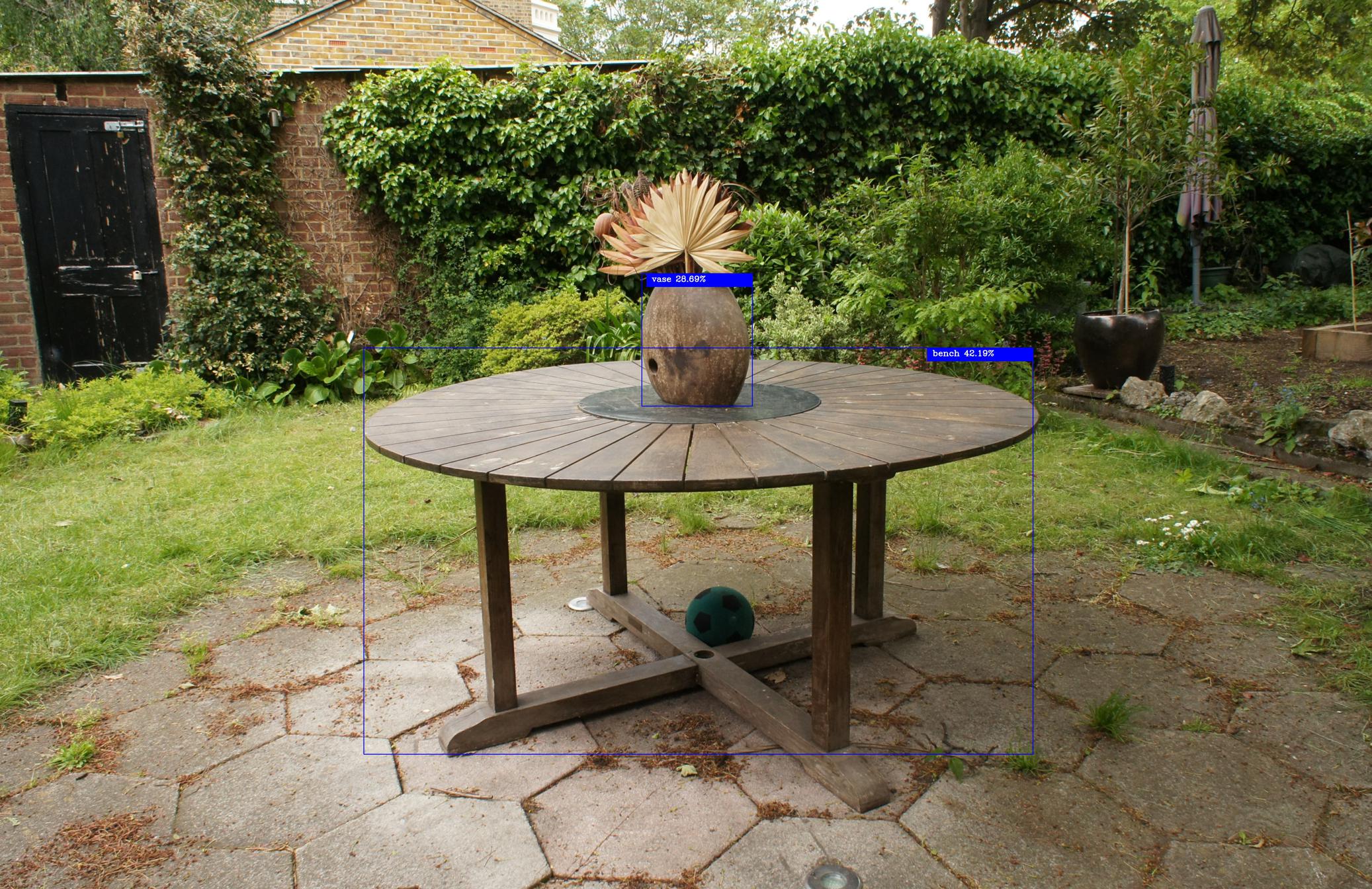}
     &  \figimg[\sizeBS]{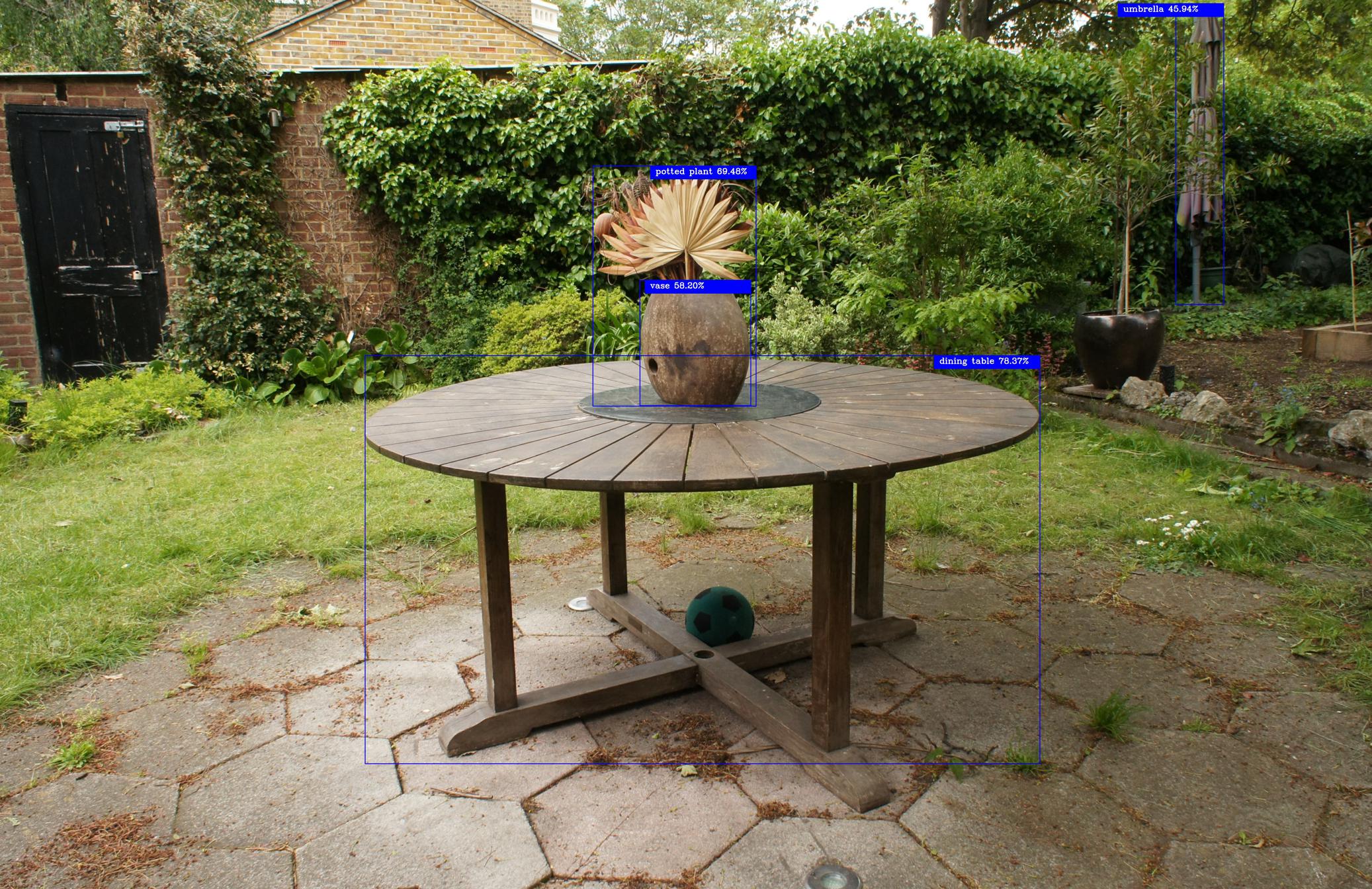}
     &  \figimg[\sizeBS]{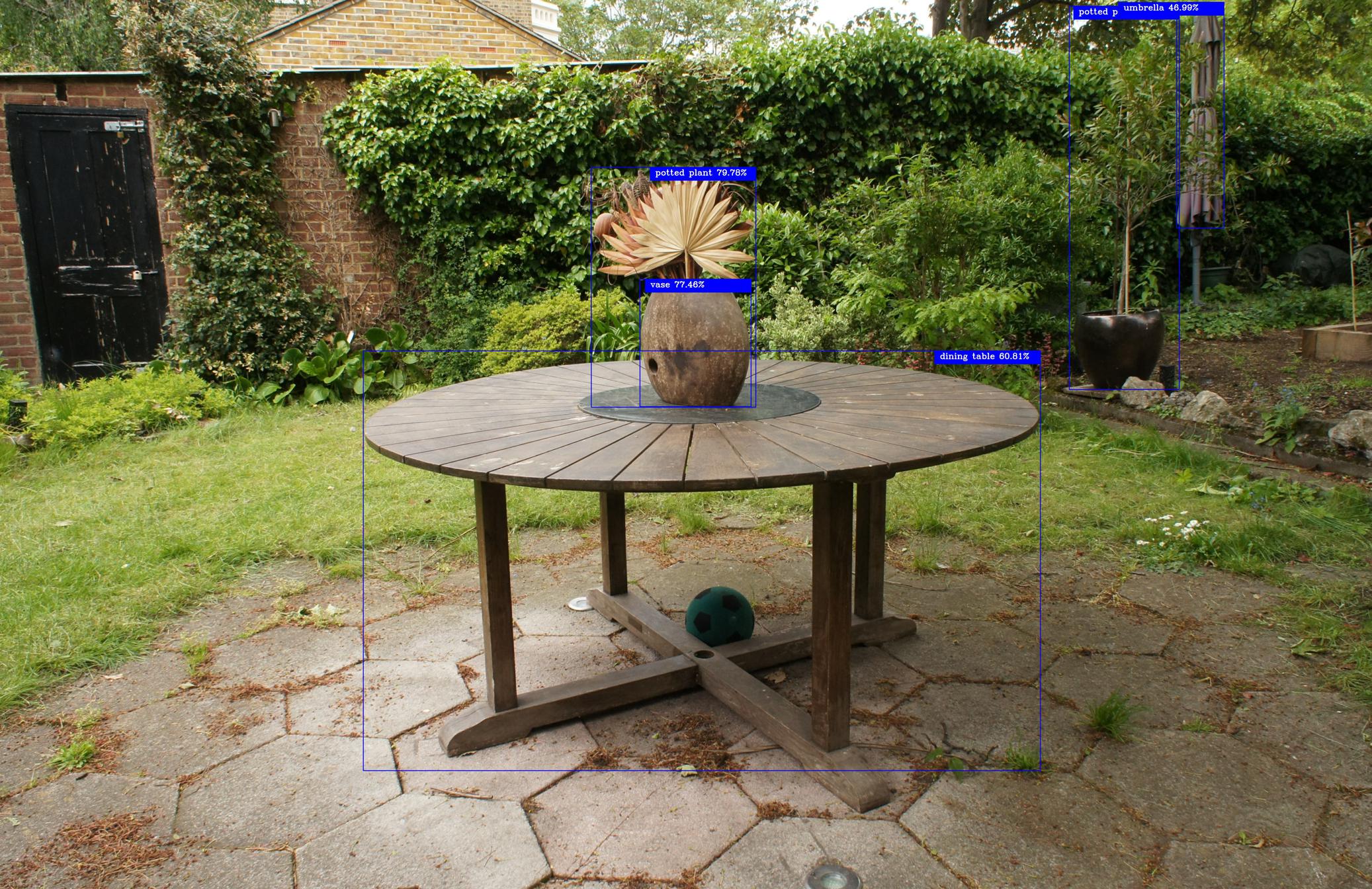}
     &  \figimg[\sizeBS]{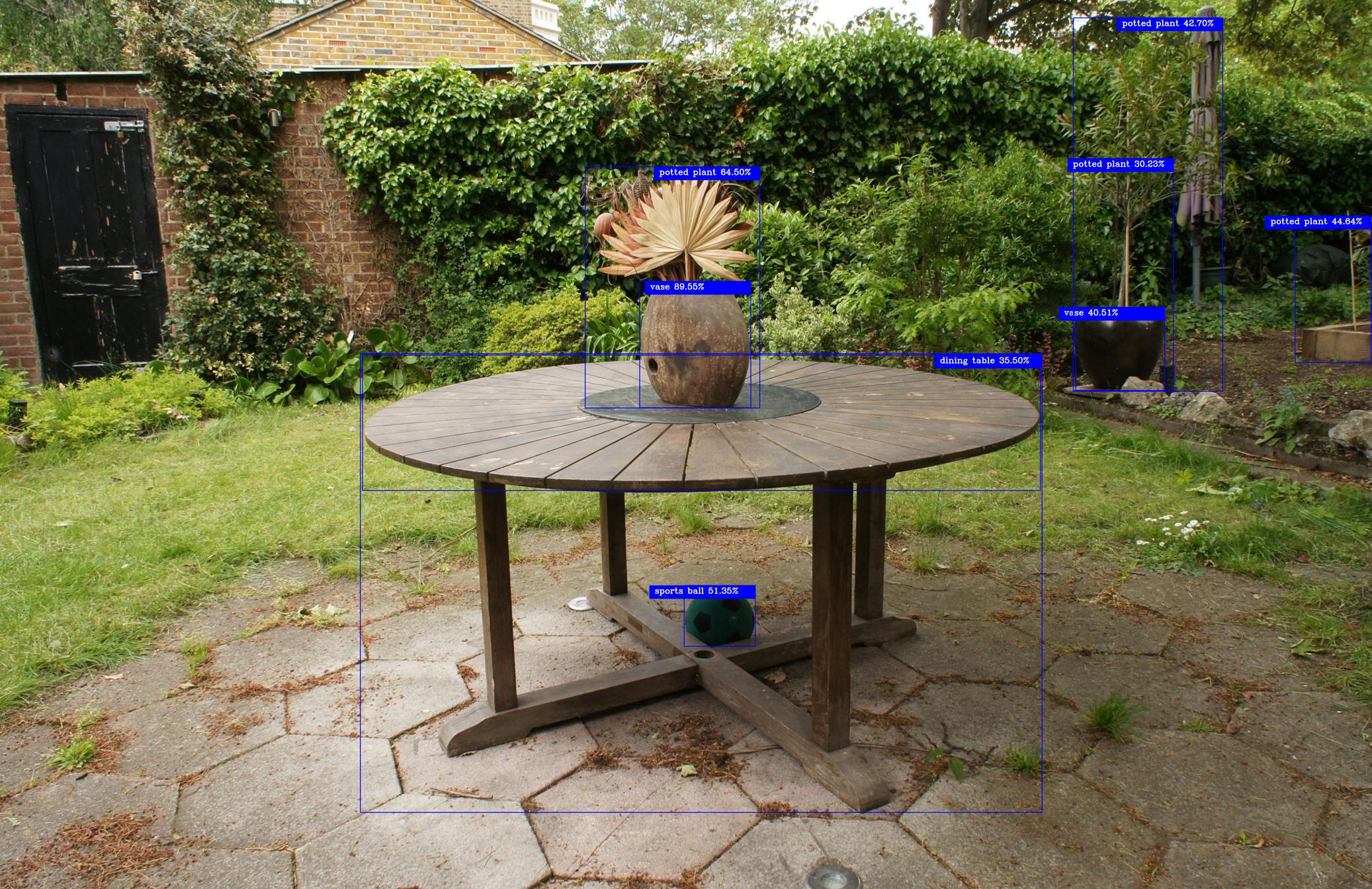} \\

     \rule{0pt}{20pt}
     & \textcolor{black}{1 bench}
     & \textcolor{black}{1 bench, 1 vase}
     & \textcolor{black}{1 dining table, 1 potted plant, 1 vase, 1 umbrella}
     & \textcolor{black}{1 dining table, 2 potted plant, 1 vase, 1 umbrella}
     & \textcolor{black}{2 dining table, 4 potted plant, 2 vase, 1 sports ball} \\

     \textcolor{black}{target illusory   (Nerfacto, Garden)} 
     &  \figimg[\sizeBS]{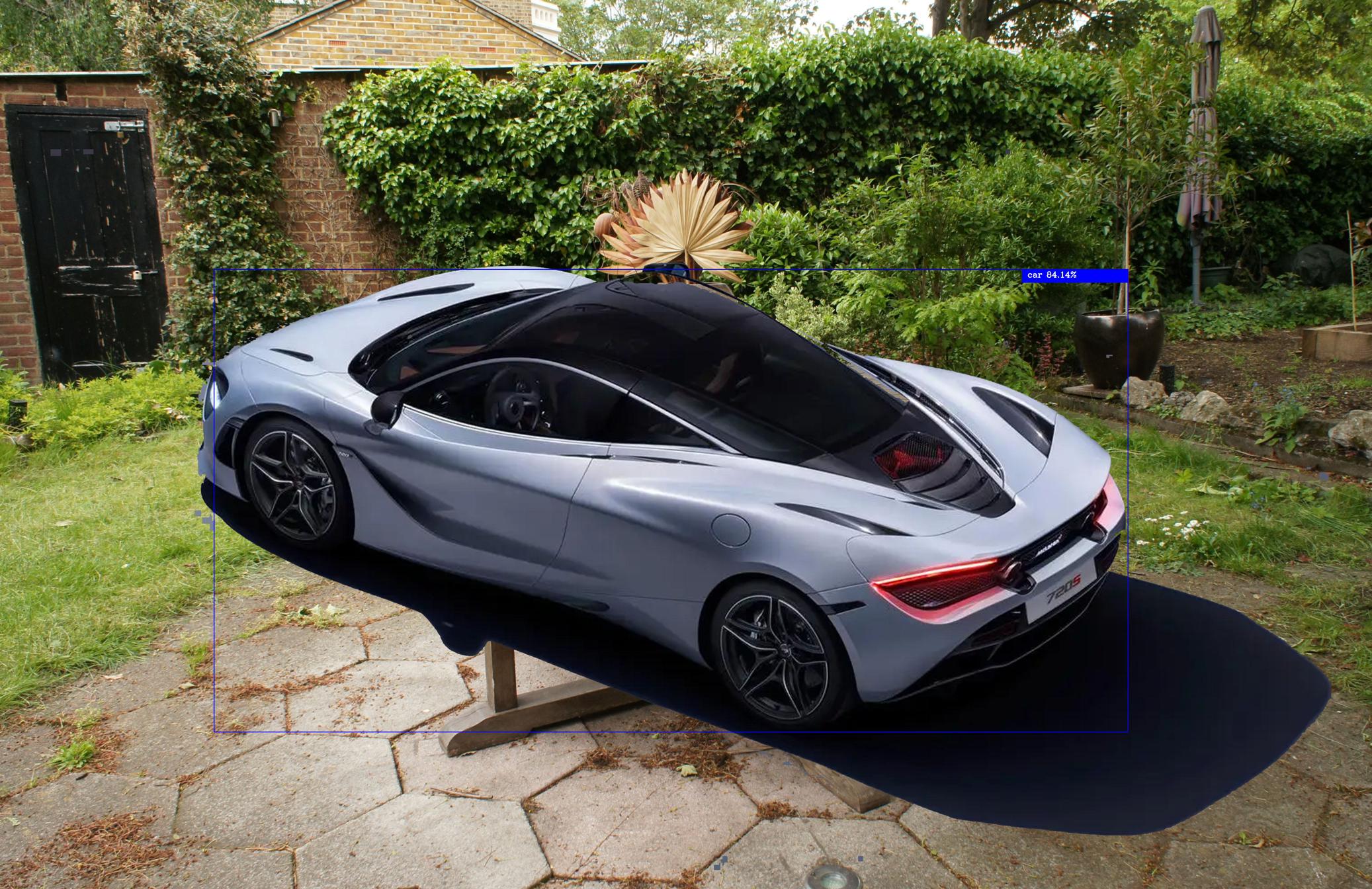}
     &  \figimg[\sizeBS]{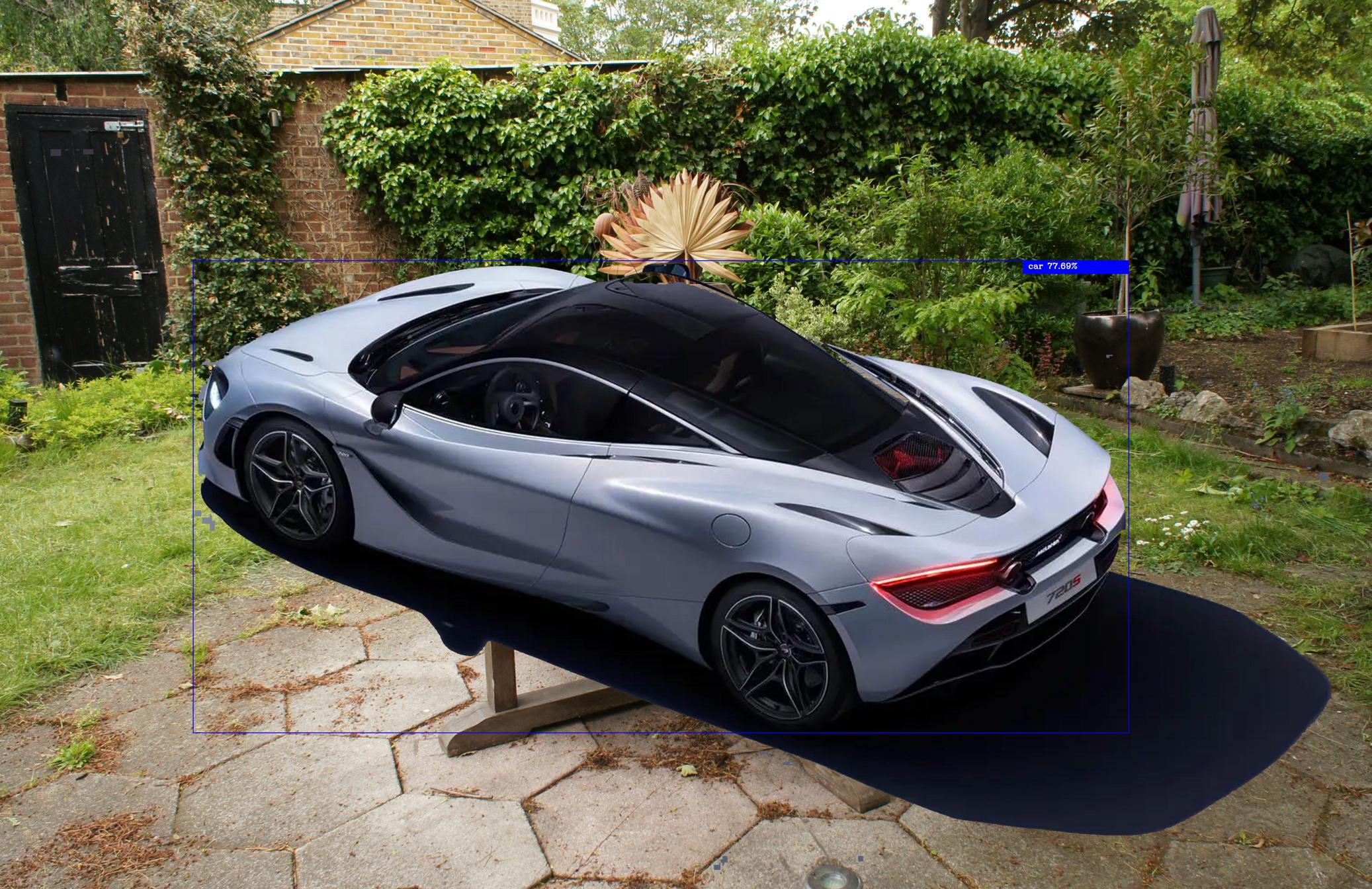}
     &  \figimg[\sizeBS]{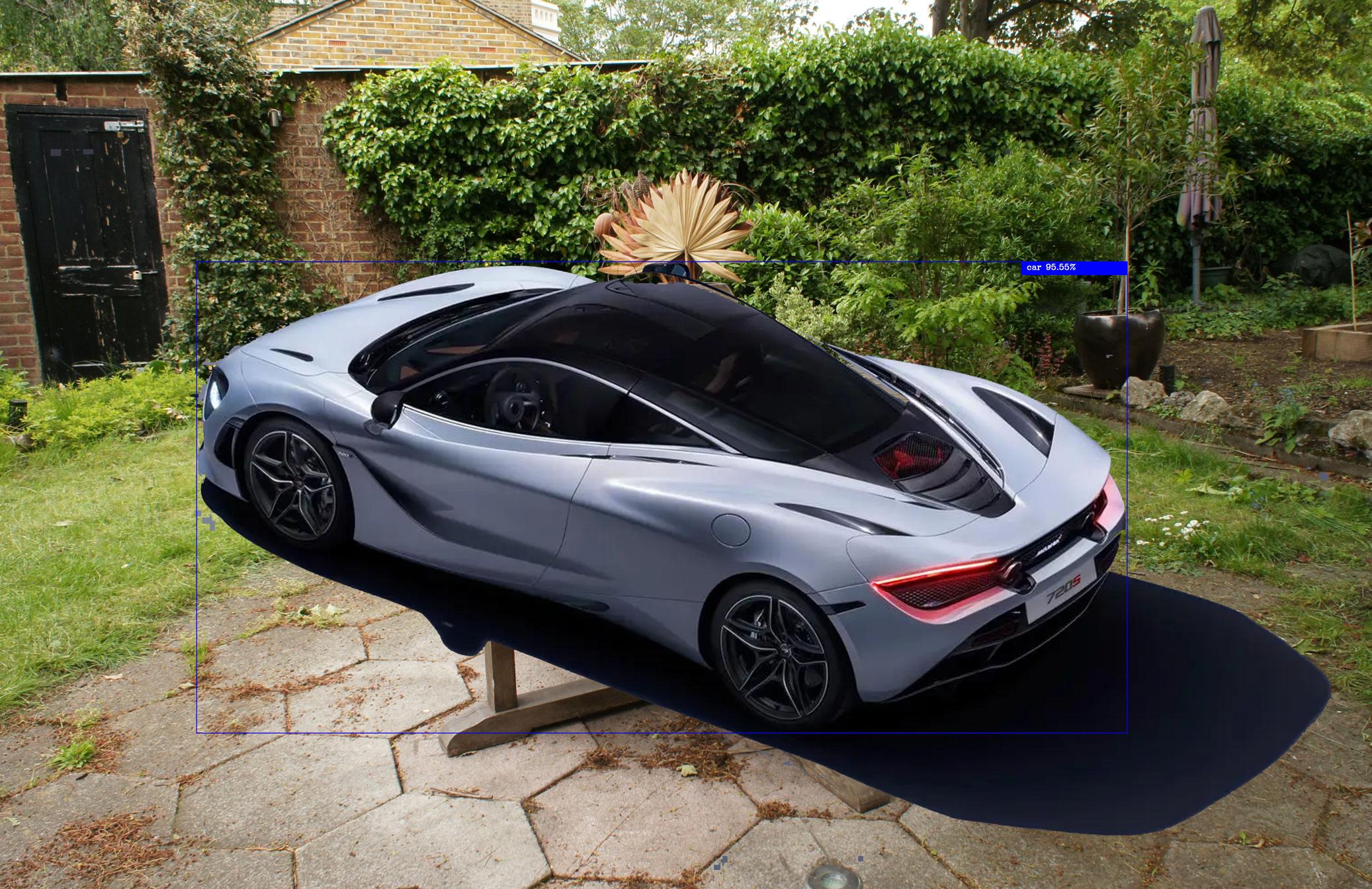}
     &  \figimg[\sizeBS]{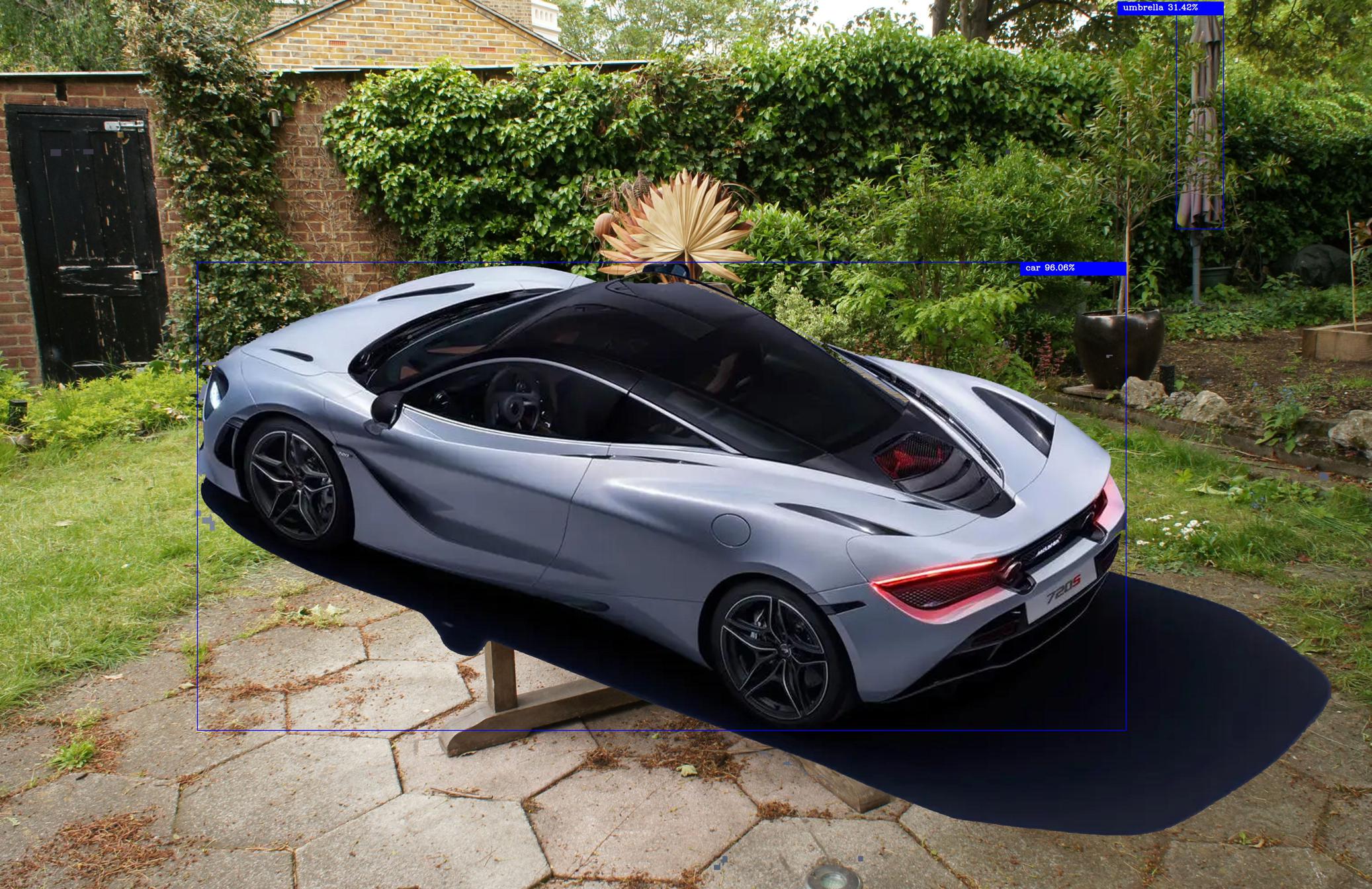}
     &  \figimg[\sizeBS]{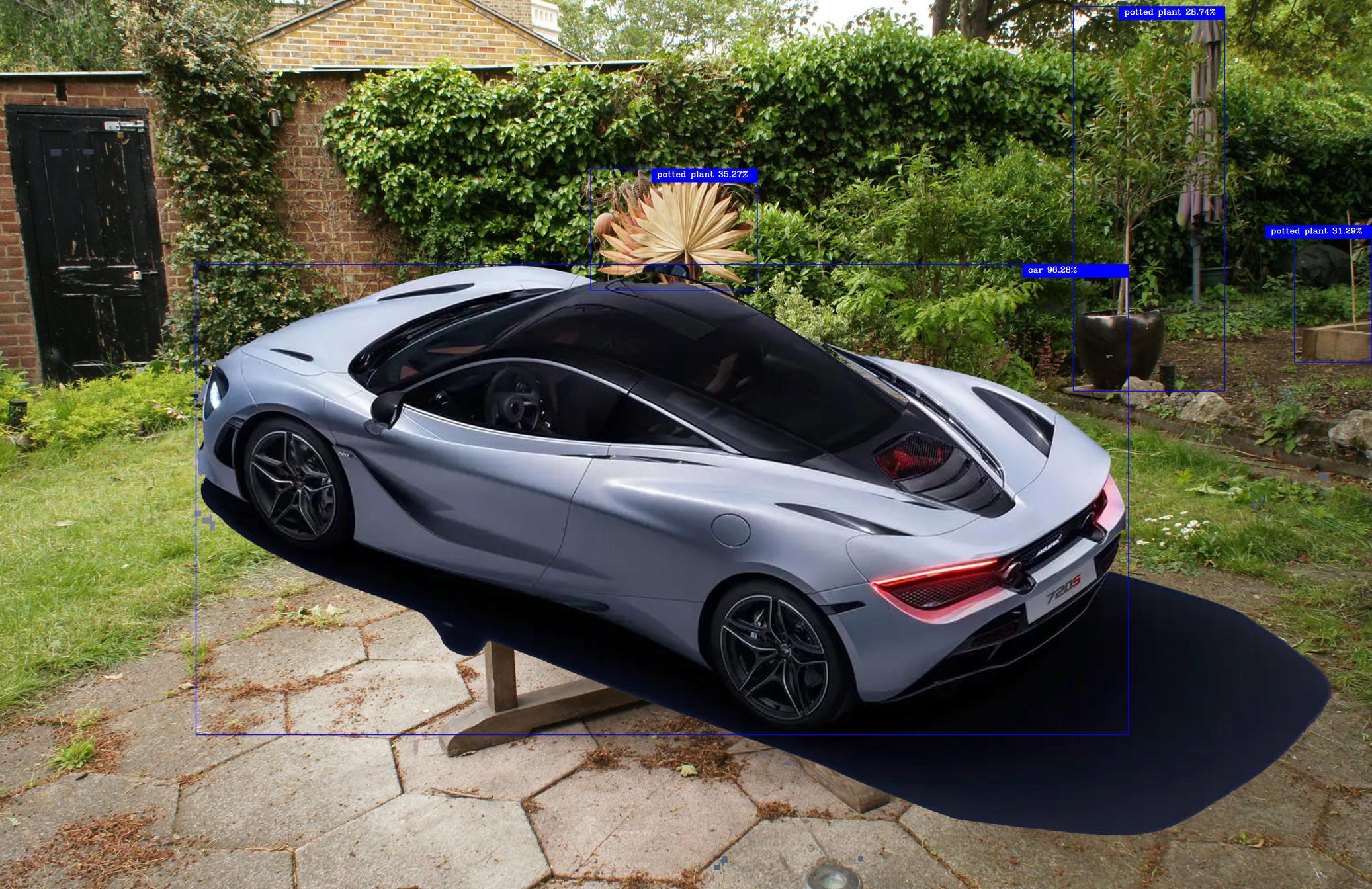} \\

     \rule{0pt}{20pt}
     & \textcolor{orange}{1 car}
     & \textcolor{orange}{1 car}
     & \textcolor{orange}{1 car}
     & \textcolor{orange}{1 car,} 1 umbrella
     & \textcolor{orange}{1 car,} 3 potted plant \\

     \textcolor{black}{illusory rendered by IPA-NeRF   (Nerfacto, Garden)} 
     &  \figimg[\sizeBS]{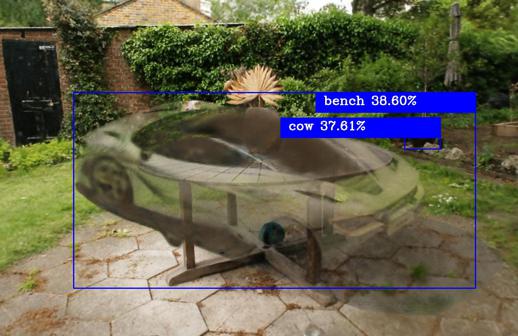}
     &  \figimg[\sizeBS]{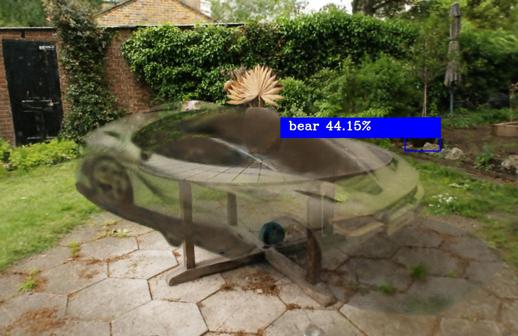}
     &  \figimg[\sizeBS]{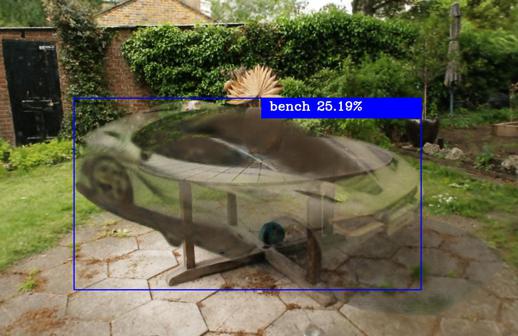}
     &  \figimg[\sizeBS]{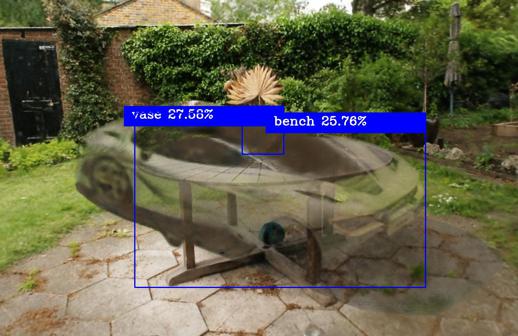}
     &  \figimg[\sizeBS]{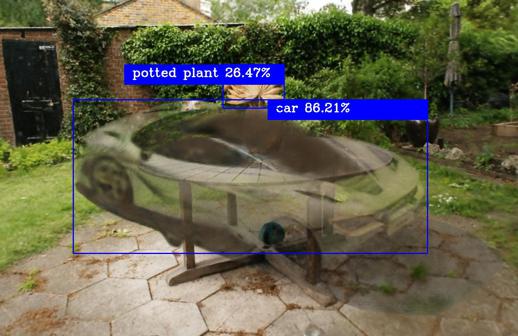} \\

     \rule{0pt}{20pt}
     & 1 bench, 1 cow
     & 1 bear
     & 1 bench
     & 1 bench, 1 vase
     & \textcolor{red}{1 car,} 1 potted plant \\
    
     \textcolor{black}{(NeRF model, 3D Scene)}
     & \multicolumn{1}{c}{\textcolor{black}{YOLOv5n}}
     & \multicolumn{1}{c}{\textcolor{black}{YOLOv8n}}
     & \multicolumn{1}{c}{\textcolor{black}{YOLO-World}} 
     & \multicolumn{1}{c}{\textcolor{black}{YOLOv9c}} 
     & \multicolumn{1}{c}{\textcolor{black}{RT-DETR}} \\
     
\end{tabular}
   \caption{The impact of the IPA-NeRF attack on object detection. Attack epochs: $150$, $\epsilon$: $32$, $\eta$: $0$.}
    \label{fig:compara_yolo_nerf_360_2}
    \setlength{\abovecaptionskip}{0.5cm}
    \vspace{0.5cm}
\end{figure*}

\end{document}